\documentclass[10pt,journal,compsoc]{IEEEtran}

\ifCLASSINFOpdf
\else
\fi

\usepackage{booktabs}
\usepackage{multirow}
\usepackage{arydshln}
\usepackage{makecell}
\usepackage{enumitem}
\usepackage{diagbox}
\usepackage{amssymb}
\usepackage{graphicx}
\usepackage{soul}
\usepackage{url}
\usepackage{amsmath}
\usepackage{cite}
\usepackage{amsthm}
\usepackage{hyperref}

\usepackage{algorithm}
\usepackage{algorithmic}
\usepackage{multirow}
\usepackage{marvosym}
\usepackage{threeparttable}
\usepackage{array,subfig}

\usepackage{mathrsfs}
\usepackage{color,xcolor}
\usepackage{colortbl}
\begin{document}
\title{Homeomorphism Prior for False Positive and Negative Problem in Medical Image Dense Contrastive Representation Learning}

\author{Yuting He,
        Boyu Wang,
        Rongjun Ge,
        Yang Chen,
        Guanyu Yang\IEEEauthorrefmark{1},
        Shuo Li
\thanks{\vspace*{-1\baselineskip} \newline {\emph{\IEEEauthorrefmark{1}Corresponding authors: G. Yang. (e-mail: yang.list@seu.edu.cn)}}}
\thanks{This research was supported by the National Natural Science Foundation of China (Grant No. 82441021), the Natural Science Foundation of Jiangsu Province (Grant No. BK20210291), the National Natural Science Foundation of China (Grant No. 62101249, T2225025), the Jiangsu Shuangchuang Talent Program (Grant No. JSSCBS20220202).}
\IEEEcompsocitemizethanks{
\IEEEcompsocthanksitem Y. He, Y. Chen and G. Yang\IEEEauthorrefmark{1} are with the Key Laboratory of New Generation Artificial Intelligence Technology and Its Interdisciplinary Applications (Southeast University), Ministry of Education, Nanjing, China, the Centre de Recherche en Information Biomédicale Sino-Français (CRIBs), and Jiangsu Provincial Joint International Research Laboratory of Medical Information Processing, Nanjing, China. (e-mail: yang.list@seu.edu.cn)
\IEEEcompsocthanksitem R. Ge is with the School of Instrument Science and Engineering, Southeast University, Nanjing, China (e-mail: rongjun\_ge@seu.edu.cn)
\IEEEcompsocthanksitem B. Wang is with the Department of Computer Science, Western University, London, ON N6A 3K7, Canada. (e-mail: bwang@csd.uwo.ca)
\IEEEcompsocthanksitem S. Li and Y. He are with the Department of Biomedical Engineering and the Department of Computer and Data Science, Case Western Reserve University, Cleveland, OH 44106 USA (e-mail: shuo.li11@case.edu).
}
\thanks{Manuscript received April 19, 2005; revised August 26, 2015.}}

\markboth{Journal of \LaTeX\ Class Files,~Vol.~14, No.~8, August~2015}%
{Shell \MakeLowercase{\textit{et al.}}: Bare Demo of IEEEtran.cls for Computer Society Journals}

\IEEEtitleabstractindextext{%

\begin{abstract}
Dense contrastive representation learning (DCRL) has greatly improved the learning efficiency for image dense prediction tasks, showing its great potential to reduce the large costs of medical image collection and dense annotation. However, the properties of medical images make unreliable correspondence discovery, bringing an open problem of \emph{large-scale false positive and negative} (FP\&N) \emph{pairs} in DCRL. In this paper, we propose \textbf{GE}o\textbf{M}etric v\textbf{I}sual de\textbf{N}se s\textbf{I}milarity (\textbf{GEMINI}) learning which embeds the homeomorphism prior to DCRL and enables a reliable correspondence discovery for effective dense contrast. We proposes a deformable homeomorphism learning (DHL) which models the homeomorphism of medical images and learns to estimate a deformable mapping to predict the pixels' correspondence under the condition of topological preservation. It effectively reduces the searching space of pairing and drives an implicit and soft learning of negative pairs via gradient. We also proposes a geometric semantic similarity (GSS) which extracts semantic information in features to measure the alignment degree for the correspondence learning. It will promote the learning efficiency and performance of deformation, constructing positive pairs reliably. We implement two practical variants on two typical representation learning tasks in our experiments. Our promising results on seven datasets which outperform the existing methods show our great superiority. We will release our code at a companion \href{https://github.com/YutingHe-list/GEMINI}{\color{magenta}\emph{website}}. 
\end{abstract}

}

\maketitle

\IEEEdisplaynontitleabstractindextext
\IEEEpeerreviewmaketitle

\section{Introduction}
\label{sec:intro}
\begin{figure}
  \centering
  \includegraphics[width=\linewidth]{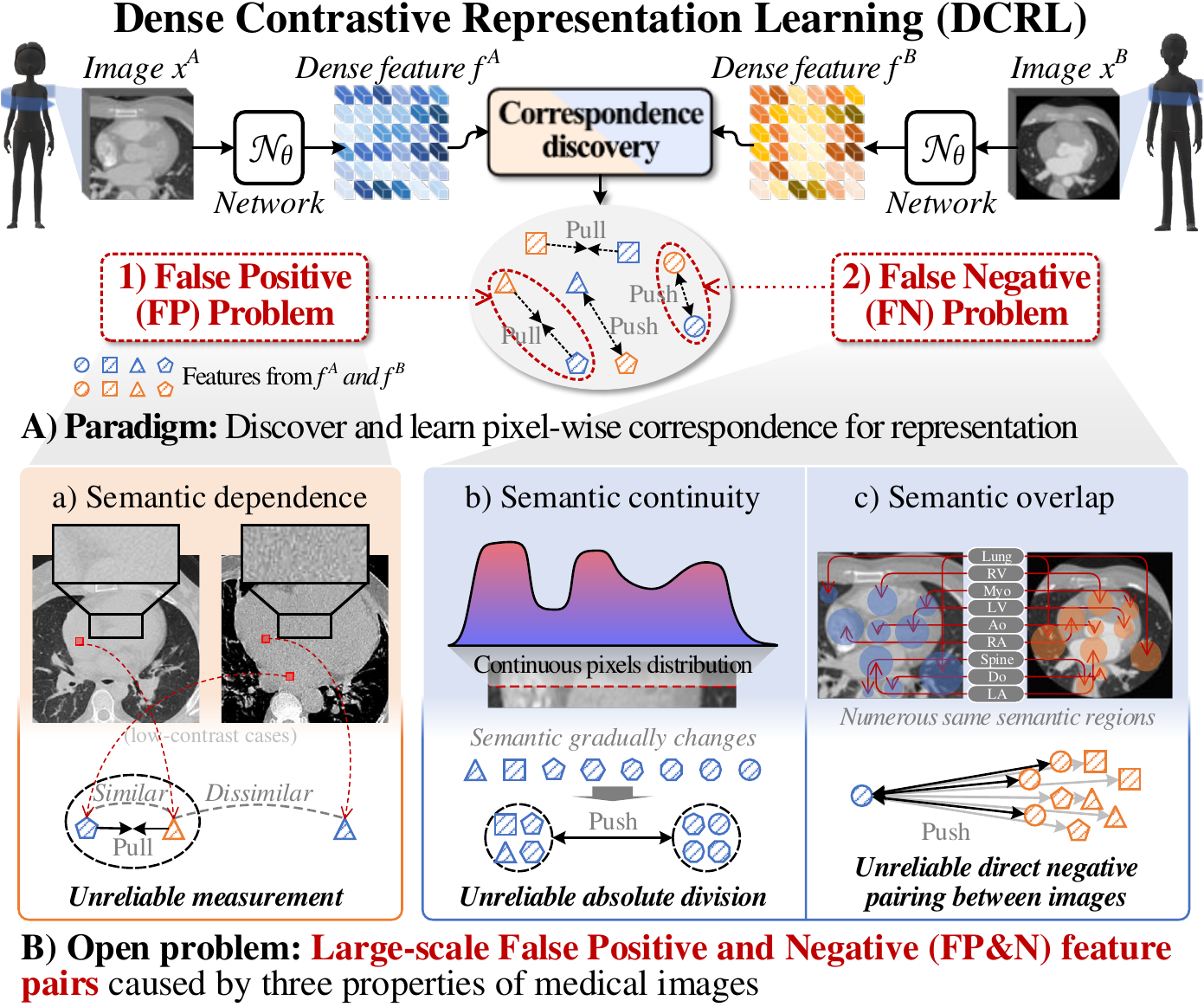}
  \caption{The DCRL with the large-scale FP\&N problem. A) The DCRL pulls and pushes the positive and negative feature pairs for consistent or distinct representation. However, B) medical images' properties cause unreliable correspondence discovery, resulting in the open problem of large-scale FP\&N features pairs in DCRL and extremely limiting the representation learning ability.}\label{intro:dcrl}
\end{figure}
\IEEEPARstart{D}{ense} contrastive representation learning (DCRL) \cite{li2021dense,o2020unsupervised,wang2022densecl,wang2022exploring,xie2021propagate,bengio2013representation,you2022simcvd} is crucial for medical image dense prediction (MIDP) tasks, e.g., segmentation \cite{he2022learning,he2021meta,you2022class}. With the increasing demand for deep learning in medical image applications \cite{piccialli2021survey}, the extremely high cost of medical image collection and dense annotation are becoming a large bottleneck \cite{cheplygina2019not}. The DCRL discovers the corresponding pixel\footnote{pixel for 2D and voxel for 3D images, we call "pixel" uniformly}-wise features \cite{milbich2021visual,zhang2022attributable,roth2020pads,you2022momentum} to drive the learning of consistent or distinct representation for them (Fig.\ref{intro:dcrl} A)) which will effectively capture the dense posterior distribution of the underlying explanatory factors for the input. Therefore, it will make models easier to extract useful information \cite{bengio2013representation} when learning MIDP tasks, thus pushing the label and data efficiency to soaring heights in the learning process and coping with the large challenge in the data collection and dense annotation \cite{litjens2017survey}.
\begin{figure}
  \centering
  \includegraphics[width=\linewidth]{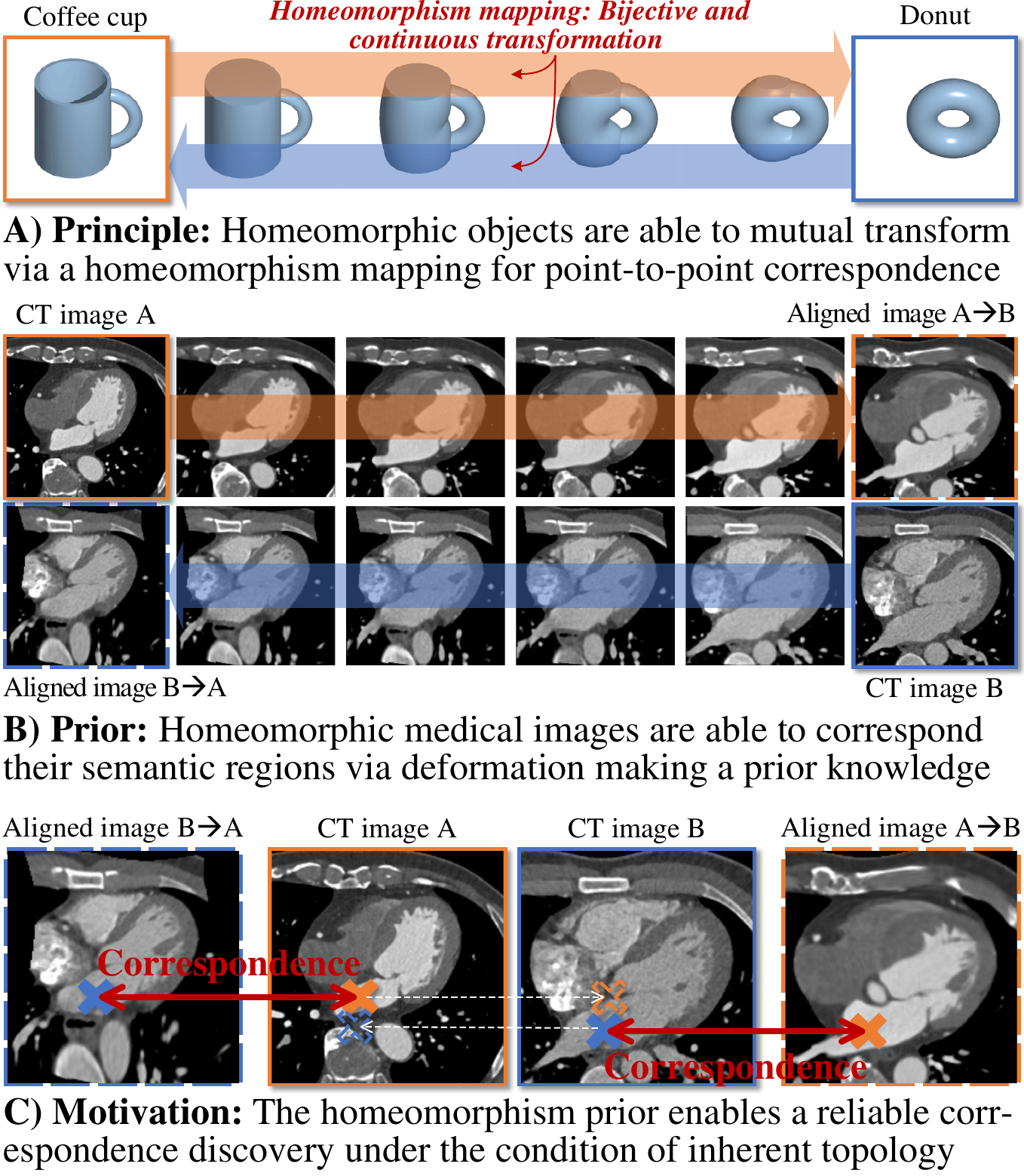}
  \caption{The homeomorphism prior enables the pixel-wise correspondence discovery under the condition of medical images' inherent topology, promoting its reliability. A) In topologie, the homeomorphic objects are able to align their topologies via a homeomorphism mapping for point-to-point correspondence with topological preservation. B) Due to the consistency of human body, the medical images are homeomorphic in image space. This provides prior knowledge to construct a deformable mapping for the pixels' correspondence under the condition of their inherent topology, which will effectively reduce the searching space of pairing. C) This gives a potential to enable a reliable pixel-wise correspondence discovery in the medical image DCRL.}\label{intro:1}
\end{figure}

Although some DCRL works \cite{li2021dense,o2020unsupervised,wang2022densecl,wang2022exploring,xie2021propagate} on natural images have been reported, medical images' properties will cause extremely unreliable correspondence discovery (Fig.\ref{intro:dcrl} B)), leading to an open problem of \emph{large-scale false positive and negative} (FP\&N) \emph{feature pairs} \cite{chuang2020debiased,chuang2022robust} in DCRL:

\emph{1) Large-scale false positive (FP) pairs caused by the semantic dependence:} (Fig.\ref{intro:dcrl} B) a)) Some existing works \cite{wang2022densecl,li2021dense} measure the similarity between the pixel-wise features to discover the positive pairs. However, due to the low- and noisy-contrast acquisitions \cite{zhou2019high} of medical images, numerous semantic regions in these images are insignificant (e.g., the soft tissues in CT images). These insignificant regions have large dependence with each other making it a challenge to distinguish them. Therefore, if directly measuring the similarity for all features, the features will be mispaired easily resulting in large-scale FP pairs. Although some other works \cite{o2020unsupervised,xie2021propagate,wang2022exploring} utilize the correspondence from the manual transformation of the same images weakening the FP problem, the diversity of positive pairs is limited (only paired from the same images) and they are still limited by large-scale FN problem.

\emph{2) Large-scale false negative (FN) pairs caused by semantic continuity and semantic overlap:} \textbf{a.} Different from the image-wise contrastive learning \cite{he2020momentum,chen2020simple,chen2020improved,grill2020bootstrap} whose unit is ``instance", the DCRL utilizes the ``pixel" as the unit which continuously arranges on images and constructs semantics via numerous pixels due to the continuity of image signal \cite{he2020momentum} (Fig.\ref{intro:dcrl} B) b)). This makes the semantics continuously distributed on images, so it is unreliable to absolutely divide the pixel-wise features as different semantics on images as negative pairs. \textbf{b.} Some natural DCRL methods \cite{wang2022densecl,li2021dense,wang2022exploring} utilize the features from different images as negative pairs, like momentum contrast mechanism \cite{He2020CVPR,wang2022densecl,li2021dense} which utilizes a memory bank to save previous features as negative samples (Fig.\ref{intro:dcrl} B) c)). However, due to the consistency of human body, medical images have similar global content which shares numerous same anatomies. This makes numerous overlapped semantics between images, so that it will construct numerous negative pairs with the same semantics resulting in FN pairs.

During correspondence discovery, these properties will enlarge the risk of the FP\&N feature pairs \cite{chuang2020debiased,chuang2022robust} resulting in large-scale FP\&N problem. It will train the network's representation to deviate from reality, making the pre-trained network even worse than the randomly initialized network. Therefore, we seek to answer the following question: \emph{How to cope with the FP\&N problem for reliable pixel-wise correspondence in the medical image DCRL?}

\emph{Motivation:} Inspired by topologie \cite{alexandroff2013topologie} (Fig.\ref{intro:1}), homeomorphism \cite{i1996new} between medical images \cite{heimann2009statistical,bazin2008homeomorphic,miller2001group}, e.g., CT, MR, X-ray, provides a prior knowledge for reliable pixel-wise correspondence. An often-repeated mathematical joke is that ``topologists cannot tell the difference between a coffee cup and a donut" \cite{hubbard1991differential} (Fig.\ref{intro:1} A)), because the coffee and donut are homeomorphic (have the same topology) and they are able to transform to each other via a topology-preserved mapping (homeomorphism, a bijective and continuous transformation). The consistency of human genes determines that the human bodies have similar anatomies \cite{netter2014atlas}, for example, human hearts have four chambers with a fixed spatial relationship, and the human brain has a fixed functional regions distribution \cite{bazin2008homeomorphic}. This makes the medical images scanned from the same body ranges have stable similar anatomies \cite{netter2014atlas} with consistent context topology \cite{He_2023_CVPR}, showing the homeomorphic topology (Fig.\ref{intro:1} B)). Therefore, based on the topologie principle and the intrinsic homeomorphic topology of medical images, it will be easy to align the semantic regions via a deformation (a homeomorphism mapping function, we name it deformable mapping in this paper). This makes an effective prior knowledge that enables a reliable pixel-wise correspondence discovery inter images in DCRL under the condition of medical images' inherent topology (Fig.\ref{intro:1} C)), reducing the searching space of constructing the correspondence. Therefore, we hypothesize that \emph{``Embedding this homeomorphism prior knowledge to the medical images DCRL will prompt the pixel-wise correspondence discovery to improve the dense representation.}

However, it is challenging to directly utilize this homeomorphism prior knowledge in DCRL due to: \textbf{1)} Lack of negative pairs: Although the homeomorphism prior enables a reliable pixel-wise correspondence for positive pairs, the non-corresponding positions are unable to be directly divided as negative pairs due to the semantic continuity, limiting the contraction of negative pairs. \textbf{2)} Weak positive pairs: The estimation of the deformation requires a measurement of the alignment degree between images. However, due to the limitation in the insignificant and varied appearance of medical images, the widely used visual similarity \cite{7987758,ba2018un,haskins2020deep,dalca2019unsupervised,He_2023_CVPR}, which utilizes the pixel intensity of the images to measure the alignment degree, will be interrupted. It will limit the deformation accuracy and cause numerous false positive pairs on misalignment regions, resulting in weak learning once very poor alignment occurs.

In this paper, we advance the geometric visual similarity learning (GVSL, our CVPR 2023 work \cite{He_2023_CVPR} discussed in Sec.\ref{sec:adv}) in DCRL, model the homeomorphism behind the GVSL, and propose the \textbf{GE}o\textbf{M}etric v\textbf{I}sual de\textbf{N}se s\textbf{I}milarity (\textbf{GEMINI}) Learning. Its objective is to model the homeomorphism prior to DCRL to cope with the large-scale FP\&N problem in DCRL, thus advancing the effectiveness of dense representation learning for medical images with reliable pixel-wise correspondence discovery. It has two aspects:

\textbf{1) Deformable Homeomorphism Learning (DHL)} for soft learning of feature pairs. Based on the homeomorphism prior, it promotes the GVSL, models that two images have homeomorphic topology and learns to estimate a deformable mapping to align them for pixel-wise correspondence. It consists of two share-weighted representation networks (backbone) and one deformation network (deformer). The deformer is trained on the represented dense features from the backbones to predict a displacement vector flow (DVF, a deformable mapping) which deforms one image to align the other image via moving the pixels. Instead of  directly dividing negative pairs \cite{li2021dense,o2020unsupervised,wang2022densecl,wang2022exploring,xie2021propagate}, the deformer learns to discover the corresponding feature pairs from numerous pixel-wise features in the receptive field. This gradient-driven approach encourages the backbones to extract distinct features for non-corresponding (negative) pairs and consistent features for corresponding (positive) pairs between the image with overlapped semantic regions. Therefore, it will implicitly learn the feature pairs and also avoid the hard division of pixel-wise features as negative pairs, making soft learning for the continuous image signal.

\textbf{2) Geometric Semantic Similarity (GSS)} for reliable learning of positive pairs. Our GSS fuses semantic similarity into the measurement of alignment degree to promote the learning of correspondence for accurate alignment, thus constructing and learning the positive pairs reliably. Due to the representability of the backbones, the extracted features will represent significant semantic information inner their corresponding receptive fields. The same and different semantic regions will have consistent and distinct features, which will bring a more efficient measurement than the original geometric visual similarity (GVS) \cite{He_2023_CVPR}. Therefore, our GSS calculates the distance of the features between images, contributing to the DCRL in two aspects: a) This measurement will consider the significant semantic information and avoid the interference of appearance's limitation, driving an effective learning of correspondence in DHL as a novel loss function. Therefore, it will improve the soft learning of feature pairs in DHL. b) It reliably discovers pixel-wise correspondence under the condition of medical images' inherent topology, reliability enhancing the effectiveness of consistency learning from these positive features for powerful representation.

Finally, our GEMINI trains the backbones to extract consistent and distinct features for the same and different semantic regions, achieving powerful dense representation ability. Based on our GEMINI learning, we implement two practical variant frameworks on two typical representation learning tasks (semi-supervised learning, representation pre-training) \cite{bengio2013representation} in our experiments, proving the powerful ability of this novel and effective learning paradigm. This paper is an extension of the CVPR 2023 conference vision (GVSL), we have detailed discussed the advancement of the preliminary work in Sec.\ref{sec:adv}. This paper makes four significant contributions:
\begin{enumerate}[leftmargin=*]
  \item For the first time, we propose the GEMINI learning which is the first framework for the open problem of large-scale FP\&N pairs in the medical image DCRL. It embeds the homeomorphism prior and constructs a reliable correspondence discovery under the condition of topological preservation, thus building an effective learning of pixel-wise features, and promoting the representation for MIDP tasks.
  \item Our proposed \emph{Deformable Homeomorphism Learning} (DHL) models the homeomorphism prior as the prediction of a deformable mapping which trains the network a distinct representation for non-corresponding regions via gradient, achieving soft learning of negative pairs.
  \item Our proposed \emph{Geometric Semantic Similarity} (GSS) utilizes the semantic information represented in features to efficiently measure the alignment degree and promote the correspondence learning, reliably learning positive pairs.
  \item Based on our GEMINI learning, we implement two practical variant frameworks on two typical representation learning tasks, and our complete experiments on these two tasks on seven datasets demonstrate our powerful representation ability and application potential.
\end{enumerate}

Overall, our GEMINI learning has three key advantages: \textbf{a) Great efficiency:} Our GEMINI captures the posterior distribution of the underlying explanatory factors for the observed input from unlabeled images \cite{bengio2013representation}, improving label and data efficiency to soaring heights in the MIDP learning process. \textbf{b) Higher reliability:} Compared with other DCRL methods \cite{li2021dense,o2020unsupervised,wang2022densecl,wang2022exploring,xie2021propagate}, our homeomorphism prior of medical images significantly promote the correspondence discovery, bringing reliable learning of positive feature pairs and soft learning of negative feature pairs for powerful dense representation learning. \textbf{c) Powerful flexibility:} As a general DCRL paradigm, our GEMINI only needs to make some simple adjustments, e.g., adding the learning loss of MIDP tasks, changing the dimensions of the backbone and deformer networks, etc., to achieve variant frameworks that adapt to different settings and dimensions for efficient learning. In this paper, we provide two variant frameworks on both 2D and 3D experiment settings only via very simple adjustments showing our GEMINI's powerful flexibility.

\section{Related Work}
\label{sec:related}
\textbf{1) Correspondence Problem:} Broadly speaking, the correspondence problem is one of the basic problems in cognitive science \cite{nehaniv2002correspondence,brass2005imitation}, machine learning \cite{scholkopf2005object}, and computer vision \cite{he2021few}. It studies the notion of correspondence between two autonomous agents in human cognition \cite{nehaniv2002correspondence}, thus further exploring the social learning, imitation, copying, or mimicry in human activation. These human cognition studies for correspondence have been influential in machine learning in the past several decades \cite{lake2015human,scholkopf2005object}. Numerous machine learning tasks, such as the protein 3D structure prediction from amino acid sequence \cite{jumper2021highly}, the machine translation \cite{wu2016google}, the position alignment between images \cite{he2021few}, etc., are able to be modeled as correspondence problems. In this paper, we limit our scope to visual contrastive representation learning and review the below topics that are relevant to the applications considered in the sequel.

\textbf{2) Dense Contrastive Representation Learning:} DCRL is a typical correspondence problem that learns consistent or distinct representation for pixel-wise features of positive or negative pairs via constructing pixel-wise correspondence \cite{milbich2021visual,zhang2022attributable,roth2020pads}. It will effectively capture the posterior distribution of the underlying explanatory factors and make models easier to extract useful information \cite{bengio2013representation} when learning MIDP tasks. Therefore, the label and data efficiency will be effectively improved to cope with the large challenge of the extremely high cost of medical image collection and pixel-wise annotation. It has three kinds to construct positive and negative pairs. \textbf{a.} The pixel similarity-based methods \cite{wang2022densecl,li2021dense,xie2021propagate,chen2021unsupervised,gao2022unsupervised} measure the Mahalanobis \cite{de2000mahalanobis} or Euclidean \cite{wang2005euclidean} distance between pixel-wise features for the correspondence. However, the low-contrast medical images limit the discrimination of features making the measurement unreliable and constructing FP\&N pairs. \textbf{b.} The location-based methods take the shared part of two cropped patches from one image\cite{o2020unsupervised} or the same position between two medical images \cite{chaitanya2020contrastive} as the positive pairs and the features from different images or different positions as the negative pairs, avoiding the limitation of mismeasurement. However, due to the consistency of image content, the same semantic pixel regions are widely existing in different images making a serious false negative problem, and interfering with the representation learning process. \textbf{c.} The attention-based method \cite{wang2022exploring} utilizes the attention maps to extract positive pairs and their negative pairs are still directly paired from different images. It relies on a large dataset to train an attention prediction model, once the dataset is not large enough, the inaccurate attention will bring numerous mis-correspondence. The unreliable negative pairs also bring large limitations.

\textbf{3) False Positive and Negative Pairs Problem:} FP\&N Problem \cite{chuang2020debiased,chuang2022robust} is one of the key open problems in contrastive representation learning \cite{Chen2021CVPR,grill2020bootstrap,chen2020simple,He2020CVPR,caron2018deep} and the existing works focus on FP\&N problem in image-wise. FP\&N problem is caused by the mis-correspondence of feature pairs which makes the networks learn distinct representations for the same-semantic pairs and consistent representations for the different-semantic pairs. The network will learn the inaccurate posterior distribution which is contrary to the underlying explanatory factors, extremely limiting the learning of practical tasks. Some methods \cite{grill2020bootstrap,xie2021propagate} remove the construction of negative pairs, and only learn the positive pairs constructed by different views of one image to avoid the FP\&N problem. However, it will bring the risk of dimensional collapse \cite{jing2021understanding} which makes the network unable to represent the information in images. Some other methods \cite{chuang2020debiased,chuang2022robust,huynh2022boosting} construct FP\&N-robust losses or FP\&N-discovery mechanisms to reduce the interference of inaccurate feature pairs, and have achieved promising results. However, these methods are sensitive to their additional hyper-parameters, and these hyper-parameters have to be carefully adjusted for their effectiveness. There is no success reported to cope with the FP\&N problem in the medical image DCRL whose special properties (continuous image signal, low contrast, varied appearance, consistent image content) bring more challenges in correspondence.

\textbf{4) Homeomorphism in Medical Images Analysis:} Homeomorphism is a powerful prior for medical image learning. As introduced in Sec.\ref{sec:intro}, this property comes from the consistency of human anatomy \cite{netter2014atlas}, and numerous medical image categories, e.g., CT, MR, X-ray, etc., inherit it. Therefore, using homeomorphism as prior knowledge, some medical image works have been studied. One of the classic applications is the registration \cite{he2021few,7987758,ba2018un,haskins2020deep,dalca2019unsupervised}. According to this prior, it aligns the anatomical structures in medical images to the same spatial coordinate system, so that the images' consistent anatomies will be aligned in image space. Based on the registration, some works \cite{bazin2008homeomorphic,6226425,8014481,wang2020lt,BalakrishnanVoxelMorph(u)} further study the atlas-based segmentation methods. They align the labeled images to unlabeled images, thus indirectly mapping the labels to unlabeled images for segmentation results. These methods have very small label requirements and large robustness due to homeomorphism. With the development of deep learning \cite{lecun2015deep,shen2017deep}, the homeomorphism prior is further promoting the medical image learning, e.g., few-shot segmentation \cite{he2022learning,he2020deep,ding2021modeling,zhao2019data,xu2019deepatlas}. They have effectively improved the learning efficiency and robustness in these tasks, owing to the large contribution from the homeomorphism prior. \textbf{However}, due to the limitations illustrated in Sec.\ref{sec:intro}, it is still challenging to embed this great prior into the DCRL tasks and there are no preliminary studies reported. 
\section{Methodology}
\label{sec:methodology}
\begin{figure}
  \centering
  \includegraphics[width=\linewidth]{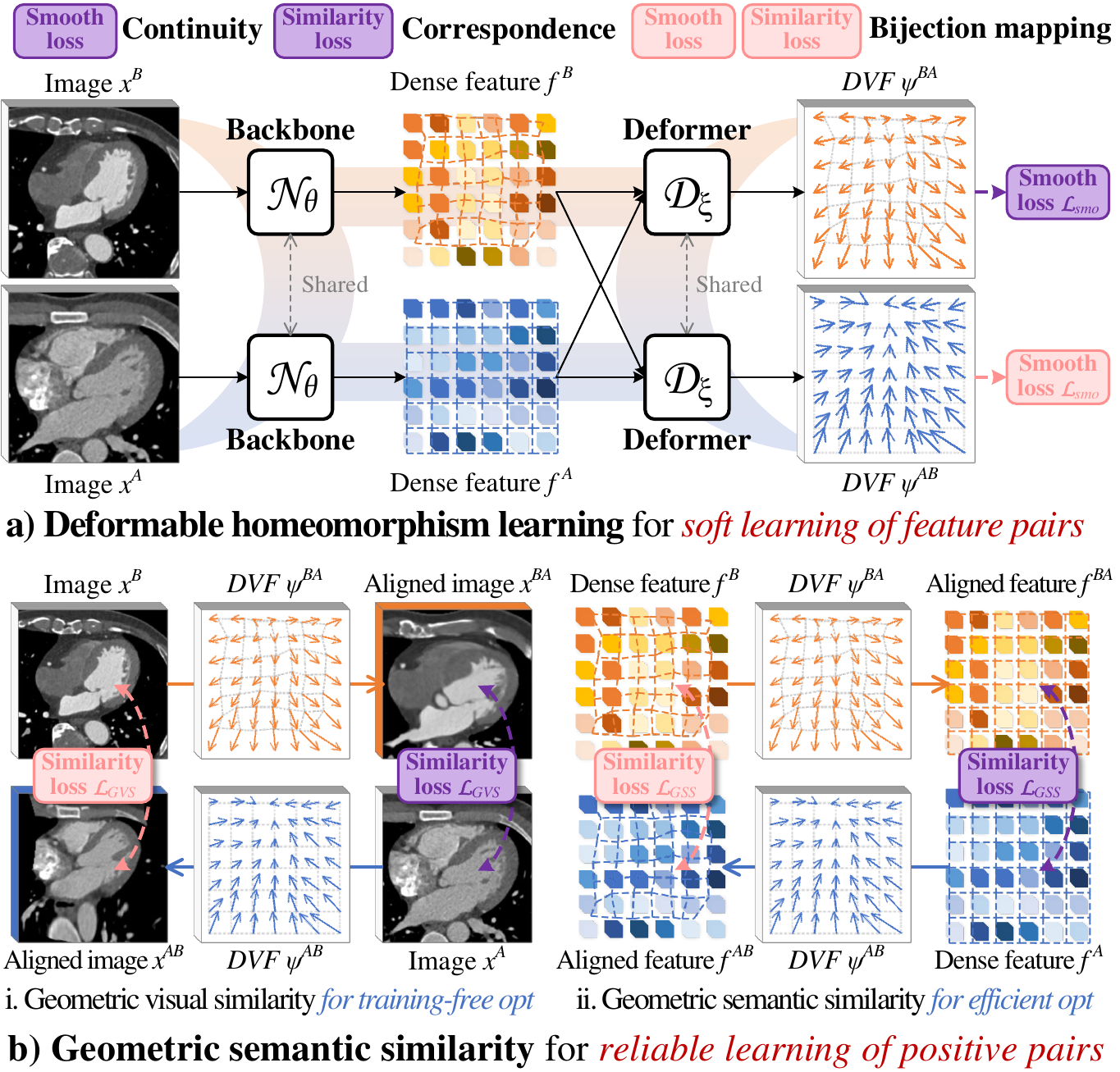}
  \caption{Our GEMINI embedded the homeomorphism prior in medical images achieves a reliable correspondence discovery in DCRL. It has two aspects: a) The DHL (Sec.\ref{sec:dhl}) learns a deformable mapping for soft learning of negative pairs. b) The GSS (Sec.\ref{sec:gss}) fuses semantic similarity into the measurement of correspondence degree to construct the positive pairs reliably. The ``opt" is the optimization. More details are described in Sec.D of our \emph{Supplementary Materials}.}\label{method:framework}
\end{figure}
The proposed GEMINI paradigm (Fig.\ref{method:framework}) constructs soft learning of negative pairs (DHL, Sec.\ref{sec:dhl}) and reliable learning of positive pairs (GSS, Sec.\ref{sec:gss}) based on the homeomorphism prior knowledge for the large-scale FP\&N problem in DCRL for powerful representation.

\subsection{Preliminary of Homeomorphism and Formulation}\label{sec:hp}
In topologie \cite{alexandroff2013topologie}, the ``two objects are homeomorphic" means they have topological equivalence. There is a bijective and continuous function (like Fig.\ref{intro:1} A)) between them for a topological-preserved transformation (homeomorphism), and each point of one object corresponds to a point of the other object. Let's define two objects in topological space, $X$ and $Y$, where the $X$ and $Y$ are the point sets of the objects (e.g., the semantic regions in images). There is a mapping function $F:X\rightarrow Y$ that transforms the $X$ to the $Y$. If the $F$ is a \emph{bijection}, is \emph{continuous}, and its \emph{inverse function} $F^{-1}$ is also \emph{continuous}, this mapping function $F$ is a homeomorphism, the two objects are homeomorphic $X\cong Y$. Based on the above theory, the \emph{bijection} and \emph{continuity} are two key elements in constructing a homeomorphic mapping for our reliable correspondence discovery.

\textbf{Formulation:} Due to the consistency of human anatomy and the intrinsic topology property of images \cite{kong1989digital}, numerous categories of medical images scanned from the same human body region are approximately homeomorphic in topologie. Let's denote two same category medical images $x^{A}$ and $x^{B}$ sampled from the dataset $\mathcal{S}$ are homeomorphic $x^{A}\cong x^{B}$. The individual variations of human body, such as height, figure, and scanning posture, make the semantic regions' position in image space different between medical images. Therefore, this enables us to construct a homeomorphism mapping function $\psi_{\mathbb{R}^{n}}$ that transforms the pixel positions in the image space $\mathbb{R}^{n}$ with $n$ dimensions for the alignment of images. It is formulated as:
\begin{equation}\label{equ:trans}
\begin{aligned}
&\psi_{\mathbb{R}^{n}}(x^{A})=x^{B}\\
&\begin{array}{ll}
s.t. & \forall I\in x^{A},\psi_{\mathbb{R}^{n}}(I)\in x^{B}\\
     & \forall J\in x^{B},\psi^{-1}_{\mathbb{R}^{n}}(J)\in x^{A},
\end{array}
\end{aligned}
\end{equation}
where the $I$ and $J$ are subsets of $x^{A}$ and $x^{B}$. The $\forall I\in x^{A},\psi_{\mathbb{R}^{n}}(I)\in x^{B}$ means that the mapping $\psi_{\mathbb{R}^{n}}$ is topological-preserved (continuous), and the $\forall J\in x^{B},\psi^{-1}_{\mathbb{R}^{n}}(J)\in x^{A}$ further means that the $\psi_{\mathbb{R}^{n}}$ is bijective and there is a continuous inverse mapping $\psi^{-1}_{\mathbb{R}^{n}}$.

\textbf{Simple Discussion:} The hypothesis of the homeomorphism prior in medical images is based on an ideal situation, i.e., the images have the exactly same content (one-by-one correspondence of the semantic regions). However, due to the potential difference in the scanning fields and the human body, medical images do not always have the same content. For example, the cardiac CT images in a large scanning window will have whole lungs, but if they are in a small scanning window, the images will only have a small part of the lungs. This makes only the shared parts homeomorphic and the whole images are not. Fortunately, in our method, we construct this homeomorphism mapping function by \emph{deformation} and use 0 value to fill the hole caused by deformation. This means that we append a point set of all zeros on the original image point set so that the points between the medical images that do not meet homeomorphism will correspond to these zero points, thus finally making the whole point set meet the homeomorphism. It is formulated as $\{x^{A},\phi^{0}\}\cong\{x^{B},\phi^{0}\}$, where the $\phi^{0}$ is the zero set has no gradient in training. Therefore, it enables us only need to focus on the homeomorphic part between medical images.

\subsection{Deformable Homeomorphism Learning (DHL)}
\label{sec:dhl}
As shown in Fig.\ref{method:framework} a), the DHL leans a deformable mapping that transforms the space of one medical image to the other image, thus driving the soft learning of feature pairs. It has two shared-weight networks which learn representation (backbone) $\mathcal{N}_{\theta}$ and a network that learns deformation (deformer) $\mathcal{D}_{\xi}$. Two medical images sampled from the dataset $\{x^{A},x^{B}\}\sim\mathcal{S}$ are putted into the backbone networks to extract their features $f^{A}=\mathcal{N}_{\theta}(x^{A}),f^{B}=\mathcal{N}_{\theta}(x^{B})$. These features are put into the deformer networks $\mathcal{D}_{\xi}$ in order ($[f^{A},f^{B}]$) and reverse order ($[f^{B},f^{A}]$) to estimate their deformable mapping functions, i.e., DVF $\psi_{\mathbb{R}^{n}}^{AB}$ from $x^{A}$ to $x^{B}$, and DVF $\psi_{\mathbb{R}^{n}}^{BA}$ from $x^{B}$ to $x^{A}$. It is formulated as
\begin{equation}\label{equ:DHL}
\psi^{AB}_{\mathbb{R}^{n}}=\mathcal{D}_{\xi}(f^{A},f^{B}),\psi^{BA}_{\mathbb{R}^{n}}=\mathcal{D}_{\xi}(f^{B},f^{A}).
\end{equation}

According to the homeomorphism prior (Equ.\ref{equ:trans}) which constructs a continuous and bijective correspondence in DVF, we train the network to learn these properties: \textbf{1)} \emph{For correspondence}, we propose a geometric semantic similarity $\mathcal{L}_{GSS}$ for efficient measurement of the correspondence degree and drive the deformer to predict reliable DVF together with the original geometric visual similarity $\mathcal{L}_{GVS}$ in GVSL \cite{He_2023_CVPR} (Sec.\ref{sec:gss}). \textbf{2)} \emph{For continuity}, we utilize a smooth loss $\mathcal{L}_{smo}$ to constrain the DVF $\psi_{\mathbb{R}^{n}}$ to perform a topological-preserved (smooth) transformation, i.e., the deformable mapping, so the semantics of the regions inner medical images will be preserved, improving the reliability of correspondence:
\begin{equation}\label{equ:smo}
   \mathcal{L}_{smo}(\psi_{\mathbb{R}^{n}})=\sum_{p\in\mathbb{R}^{n}}\|\triangledown\psi_{p}\|^{2},
\end{equation}
where the $p$ is the position of the pixels in image space $\mathbb{R}^{n}$, the $\triangledown\psi_{p}$ is the gradient of position $p$. Therefore, it will avoid over-transformation which breaks the topological structures of semantic regions, and keep the homeomorphic property of the mapping. Therefore, the loss $\mathcal{L}_{DM}$ to learn a deformable mapping (DM) is
\begin{align}\label{equ:DHLloss}
&\mathcal{L}_{DM}(\theta,\xi,\{x^{A},x^{B}\})=\underbrace{\lambda_{smo}\mathcal{L}_{smo}(\psi^{AB}_{\mathbb{R}^{n}})}_{\textbf{\text{Continuity}}}\\
&+\underbrace{\lambda_{GVS}\mathcal{L}_{GVS}(x^{A},x^{B},\psi^{AB}_{\mathbb{R}^{n}})+\mathcal{L}_{GSS}(f^{A},f^{B},\psi^{AB}_{\mathbb{R}^{n}})}_{\textbf{\text{Correspondence}}},\notag
\end{align}
where the $\lambda_{smo}$ is the weights for the smooth loss and $\lambda_{GVS}$ is the weight of the geometric visual similarity. \textbf{3)} \emph{For bijection}, we simultaneously learn forward deformable mapping and inverse deformable mapping, thus training the deformers to learn the symmetry between two medical images and constructing a bidirectional optimization objective in our DHL $\mathcal{L}_{DHL}$:
\begin{equation}\label{equ:DHLloss2}
\mathcal{L}_{DHL}=\underbrace{\mathcal{L}_{DM}(\theta,\xi,\{x^{A},x^{B}\})+\mathcal{L}_{DM}(\theta,\xi,\{x^{B},x^{A}\}))}_{\textbf{\text{Bijectivity:}}\ x^A\rightarrow x^B,\ x^B\rightarrow x^A}
\end{equation}
Therefore, during the learning of bijective deformable mapping, the deformers will try to learn a reliable correspondence between images in the DVFs ($\psi^{AB}_{\mathbb{R}^{n}},\psi^{BA}_{\mathbb{R}^{n}}$). The gradient of the loss $\mathcal{L}_{DHL}$ will optimize the backbones $\mathcal{N}_{\theta}$ for soft learning of feature pairs implicitly, bringing a reliable optimization for representation (analyzed in Sec.\ref{sec:int}).

\textbf{Simple discussion of the property:} Our DHL serves as the framework of GEMINI, enabling homeomorphism mapping between images. By leveraging the key elements of continuity and bijection, it trains the backbone to extract distinct features for non-corresponding pairs and consistent features for corresponding pairs through deformable learning gradients. This soft feature learning mitigates false negatives caused by direct negative pair distinctions. However, the backbone's weak initial representation, stemming from the 'two-player game' learning process (Sec.\ref{sec:int}), is addressed by embedding foundational tasks in GEMINI variants to warm up learning (analyzed in Sec.\ref{subsubsec:analysisiofrestor}).

\subsection{Geometric Semantic Similarities (GSS)}
\label{sec:gss}
Our GSS measures the correspondence (alignment) degree via the represented dense features to promote the learning efficiency of deformation to improve the alignment, thus constructing and learning the positive pairs reliably. As shown in Fig.\ref{method:framework} a), the original GVS \cite{He_2023_CVPR} utilizes the DVF $\psi^{AB}_{\mathbb{R}^{n}}$ to transform the image $x^{A}$ to align the image $x^{B}$, and calculate the distance of pixel intensity between the aligned image $x^{AB}$ and the image $x^{B}$ for their similarity. Due to the limitation of the appearance, the distance will be interfered which makes an unreliable measurement. Our GSS measure the similarity between the represented features (Fig.\ref{method:framework} b)) further considering the important semantic information. It utilizes the DVF $\psi^{AB}_{\mathbb{R}^{n}}$ to align the dense features $f^{A}$ to the dense features $f^{B}$, and calculate the distance of the features between the aligned features $f^{AB}$ and the feature $f^{B}$ for their similarity. In our GEMINI, the original GVS still takes part in the measurement due to its training-invariability which will not be interrupted by the learning process for a basic optimization objective. Different from the similarity measured throughout the whole space \cite{wang2022densecl}, our GSS or GVS measures the similarities only between the corresponding positions of the features or images in the image grid. Therefore, this will make the measurement under the condition of semantic regions' inherent topology which significantly reduces the searching space of the correspondence discovery, so we call the GSS and GVS ``geometric".

As discussed in Sec.\ref{sec:hp}, during the calculation of the geometric similarities, we only focus on the homeomorphic regions. Therefore, we utilize an adaptive mask mechanism to remove the zero point set and measure the alignment degree of the potentially shared regions between two images. Specifically, it generates a mask $\epsilon$ with the same size as the image $x^{A}$ and the value of $1$, and then transforms it to the space of $x^{B}$ via the DVF $\psi^{AB}_{\mathbb{R}^{n}}$ for $\epsilon^{AB}$. The void region caused by the transformation in the mask is filled with ``$0$" which is the appended zero point set. Therefore, the regions with a value of ``$1$" are the shared regions between images, and those with a value of ``$0$" are the non-corresponding regions. Only the value ``$1$" regions highlighted in the mask will be calculated for similarity. We utilize the normalized cross-correlation \cite{he2021few} for the GVS and the cosine similarity for our GSS with the mask $\epsilon$ into the measurement process:
\begin{align}\label{equ:GSSGVS}
  &\mathcal{L}^{\epsilon}_{GVS}=\sum_{p\in\{\epsilon^{AB}=1\}}\frac{(\sum_{p_{i}}(x^{AB}_{p_{i}}-\hat{x}^{AB}_{p})(x^{B}_{p_{i}}-\hat{x}^{B}_{p}))^{2}}{(\sum_{p_{i}}(x^{AB}_{p_{i}}-\hat{x}^{AB}_{p}))(\sum_{p_{i}}(x^{B}_{p_{i}}-\hat{x}^{B}_{p}))}, \\
  &\mathcal{L}^{\epsilon}_{GSS}=\sum_{p\in\{\epsilon^{AB}=1\}}\frac{f^{AB}_{p}\cdot f^{B}_{p}}{\|f^{AB}_{p}\|\|f^{B}_{p}\|},
\end{align}
where the $\epsilon^{AB}=\psi^{AB}_{\mathbb{R}^{n}}(\epsilon)$, the $x^{AB}=\psi^{AB}_{\mathbb{R}^{n}}(x^{A})$, and the $f^{AB}=\psi^{AB}_{\mathbb{R}^{n}}(f^{A})$.
\begin{figure}
  \centering
  \includegraphics[width=\linewidth]{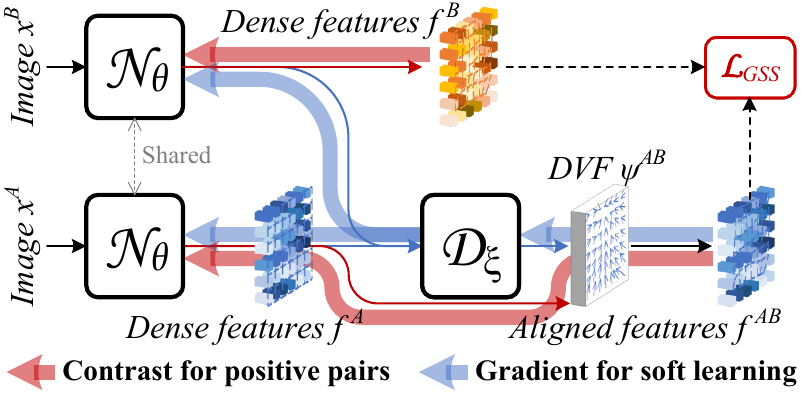}
  \caption{The gradients from the loss of our GSS simultaneously train the explicit contrast of positive pairs and drive the implicit and soft learning in our DHL.}
  \label{method:grad}
\end{figure}

Our GSS will both drive the learning of homeomorphism mapping as a loss in the DHL and train the reliable learning of positive pairs for their representation due to the homeomorphism-based correspondence. As shown in Fig.\ref{method:grad}, the gradient from the GSS will be divided into two parts, including the gradient for positive pairs (red path) and the gradient for the soft learning of feature (both positive and negative) pairs (blue path). The rad path directly learns the consistency of features in corresponding positions predicted by the deformer $\mathcal{D}_{\xi}$. Due to the homeomorphism prior, the correspondence discovery is under the condition of topological preservation which will be reliable for the contrastive learning of positive pairs. The blue path utilizes the gradient form the deformer $\mathcal{D}_{\xi}$ to optimize the backbone network $\mathcal{N}_{\theta}$ indirectly, i.e., the learning in our DHL, bringing implicit and soft learning of feature pairs (analyzed in Sec.\ref{sec:int}).

The whole optimization of our GEMINI can be modeled as two parts, including the optimization of parameters $\theta$ for representation and the optimization of the parameters $\xi$ for deformable mapping:
\begin{align}\label{equ:opt}
&\xi^{*}=\underset{\xi}{\arg\min}\ \mathcal{L}_{DHL}(\theta^{*},\xi,\mathcal{S})\\
s.t.\ & \theta^{*}=\underbrace{\underset{\theta}{\text{OPT}}(\theta,\frac{\partial\mathcal{L}_{DHL}(\theta,\xi,\mathcal{S})}{\partial \theta})}_{\text{Gradient for soft negative pairs}}+\underbrace{\lambda_{pos}\underset{\theta}{\arg\min}\ \mathcal{L}_{pos}(\theta,\mathcal{S})}_{\text{Objective for positive pairs}},\notag
\end{align}
where the $\mathcal{L}_{pos}$ represents the red path in Fig.\ref{method:grad} for positive pairs, $\lambda_{pos}$ is the weight of $\mathcal{L}_{pos}$, and the OPT$()$ is the optimization strategy to minimize the $\mathcal{L}_{DHL}$ for negative pairs (yellow path). If it is SGD \cite{bottou2010large} with one step, it will be $\theta^{*}\leftarrow\theta-\eta\frac{\partial\mathcal{L}_{DHL}(\theta,\xi,\mathcal{S})}{\partial \theta}$, where the $\eta$ is learning rate. The two optimization parts correspond to the two forward paths (Fig.\ref{method:grad}) in the inference process, so the whole optimization process is compatible with the existing gradient descent methods. We utilize the Adam \cite{kingma2014adam} for optimization.

\textbf{Simple discussion of the property:} Our GSS, a specific loss in GEMINI, facilitates correspondence learning—key to homeomorphism mapping—by leveraging the deformer to align features based on the inherent topology of medical images. This enables reliable positive pair learning, addressing the large-scale false-positive problem. As part of DHL, GSS also supports soft feature learning via gradients. To overcome limitations in discovering positive pairs in non-homeomorphic regions, our adaptive mask mechanism highlights shared regions.


\subsection{Intuitions on Behavior: Learning Reliable Positive and Implicit Negative Pairs for Dense Representation}\label{sec:int}
\begin{figure}
  \centering
  \includegraphics[width=\linewidth]{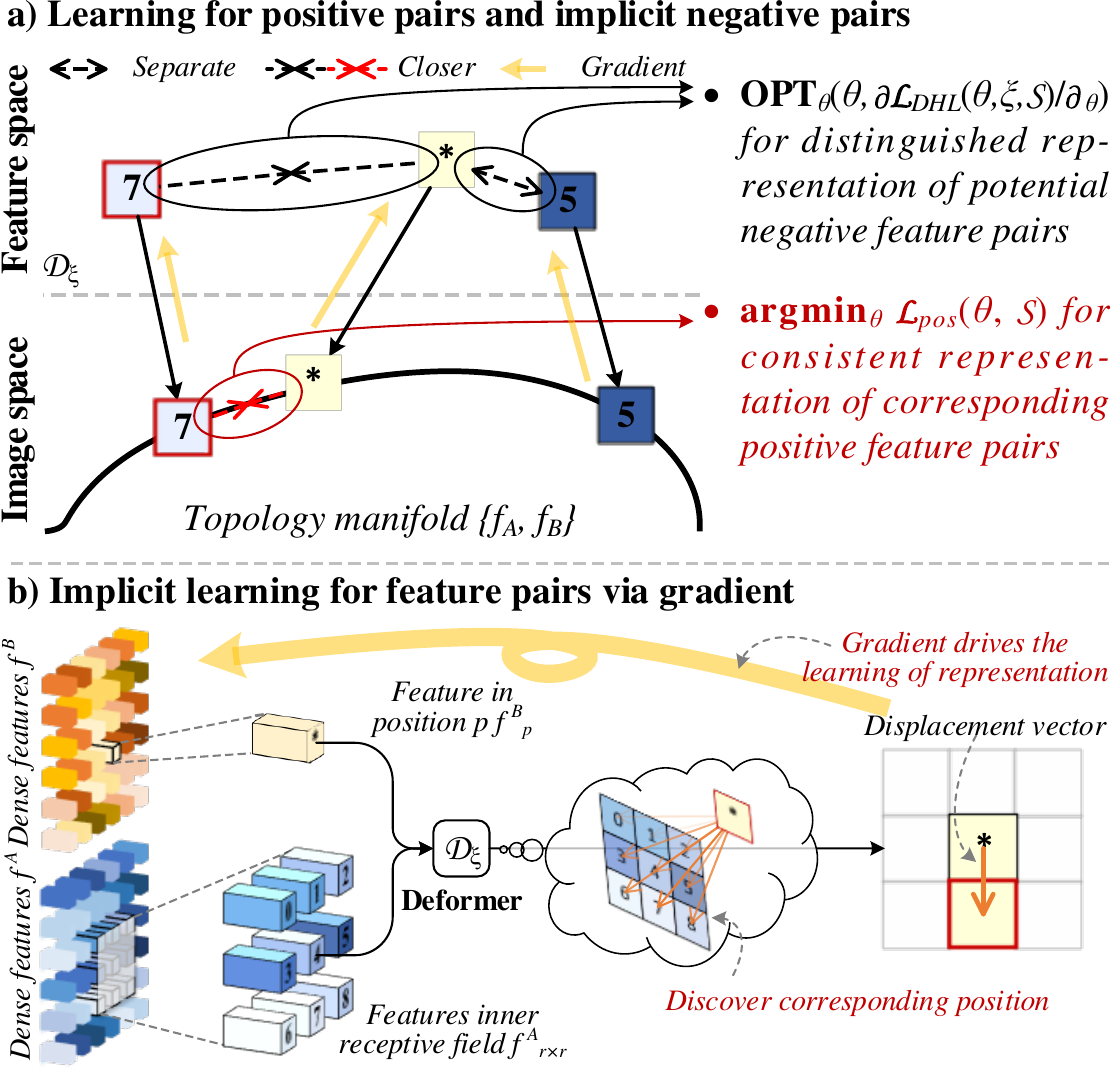}
  \caption{Intuitions on behavior. a) The two optimization objectives in Equ.\ref{equ:opt} for $\theta$ drive the reliable learning of positive and implicit learning of negative pairs. b) The feature pairs are learned softly via the gradient from the DHL.}\label{method:int}
\end{figure}
The two optimization objectives in Equ.\ref{equ:opt} for $\theta$ train the backbones to learn reliable positive feature pairs and implicit negative feature pairs, bringing an effective dense representation learning to the continuous image signal.

\textbf{For positive pairs}, it effectively reduces the searching space of pairing, bringing reliable learning for positive pairs. Our homeomorphism prior makes the correspondence discovery under the condition of the consistent topology of images formulated as $c_{i}=\mathcal{D}_{\xi}(f^{A},f^{B})_{i}={\psi^{AB}_{\mathbb{R}^{n}}}_{i}$. In the training of GEMINI, the $\psi^{AB}_{\mathbb{R}^{n}}$ have to meet the minimization of $\mathcal{L}_{smo}$, $\mathcal{L}_{GSS}$, and $\mathcal{L}_{GVS}$ which make the $\psi^{AB}_{\mathbb{R}^{n}}$ have topology-preservation ability and good alignment degree. Therefore, as shown in Fig.\ref{method:int} a), it makes our positive feature pairing consider the topology, i.e., pairing features on a topology manifold, efficiently reducing the searching space and improving the reliability. The objective for positive pairs $\arg\min_{\theta}\ \mathcal{L}_{pos}(\theta,\mathcal{S})$ learns their consistency on this manifold. The feature $f^{A}_{q}$ ($q=7$ in Fig.\ref{method:int}) from the dense features $f^{A}$ and the feature $f^{B}_{*}$ from the dense features $f^{B}$ will be consistent, i.e., $f^{A}_{q}=f^{B}_{*}$.

\textbf{For negative pairs}, it constructs implicit learning via the gradient $\text{OPT}_{\theta}(\theta,\frac{\partial\mathcal{L}_{DHL}(\theta,\xi,\mathcal{S})}{\partial \theta})$ from the DHL, avoiding the direct division of negative pairs and learning a soft contrast. Specifically, the deformer network $\mathcal{D}_{\xi}$ learns to discover the correspondence of the consistent features $\{f^{A},f^{B}\}$, thus driving the backbone represent distinct features for the features with inconsistent semantics. Fig.\ref{method:int} b) shows an example of this process. The features $f^{A}_{r^{2}}=\{f^{A}_{0},f^{A}_{1},f^{A}_{2},...,f^{A}_{r\times r}\}$ is the image $x^{A}$'s features in the receptive field of the Deformer network $\mathcal{D}_{\xi}$, where the $r$ is the width of the receptive field (it is 26 in our experiments). Due to the learning of positive pairs, the $f^{B}_{*}$ is constrained to $f^{A}_{q}$ where the $q$ is the corresponding position. Therefore, as demonstrated in Fig.\ref{method:int} a), in order to discover the corresponding position, it will train the features in non-corresponding position to distinct to the feature $f^{B}_{*}$ via a gradient.
\begin{equation}\label{equ:nega}
\begin{aligned}
&\mathcal{D}_{\xi}(f^{B}_{*},f^{A}_{r^2})=\psi^{BA}_{*},\\
&\begin{array}{lll}
s.t. & f^{B}_{*}=f^{A}_{q},&\psi^{BA}_{*}\ \text{is}\ *\rightarrow q
\end{array}
\end{aligned}
\end{equation}
Throughout the whole training process, the learning of representation in the backbone network $\mathcal{N}_{\theta}$ and the deformable mapping in the deformer network $\mathcal{D}_{\xi}$ is a two-player game \cite{saad2009coalitional}. The $\mathcal{D}_{\xi}$ learns to estimate the correspondence of semantic regions from the represented dense features $f^{A},f^{B}$ and measure their pixel displacement. The $\mathcal{N}_{\theta}$ learns to provide features of semantic regions to the $\mathcal{D}_{\xi}$ for their correspondence. To achieve more accurate correspondence, the deformer network $\mathcal{D}_{\xi}$ has to drive the backbone network $\mathcal{N}_{\theta}$ to output more distinct features in turn for different semantic regions via the gradient in backpropagation. Therefore, under this interaction, the $\mathcal{N}_{\theta}$ will provide more representative features for the $\mathcal{D}_{\xi}$ to improve the correspondence estimation, and the $\mathcal{D}_{\xi}$ will have a more powerful ability to learn the correspondence of pixel-wise features. This process needs a promising representation in backbone, but it is always unavailable at the beginning of training. Therefore, we train the GEMINI with a fundamental learning task, e.g., restoration \cite{zhou2019models,pathak2016context} in our implementation. 

\section{Experiment 1: Few-shot Semi-supervised Medical Image Segmentation (FS-Semi)}
\label{sec:task2}
We implement our GEMINI learning on few-shot semi-supervised (FS-Semi) medical image segmentation (GEMINI-Semi) providing a variant on the situation that labels are very few. Three public-available tasks are enrolled in our experiments for a very complete evaluation.
\subsection{Experiments configurations}
\label{sec:configurations2}
\subsubsection{Variant design} The variant of our GEMINI-Semi learns a segmentation head $Seg_{\kappa}$ on the extracted dense features $f^{A},f^{B}$. Therefore, except the optimization for deformable homeomorphism learning $\mathcal{L}_{DHL}$, the GEMINI-Semi also has an additional optimization for segmentation $\mathcal{L}_{Seg}$:
\begin{equation}\label{equ:variant2}
\underset{\xi,\theta,\kappa}{\arg\min}\ (\mathcal{L}_{DHL}(\theta,\xi,\mathcal{S}_{ul})+\mathcal{L}_{Seg}(\theta,\kappa,\mathcal{S}_{l})),
\end{equation}
where the $\mathcal{S}_{ul}$ and the $\mathcal{S}_{l}$ are the unlabeled dataset and the labeled dataset. In our experiment, we utilize the sum of Dice loss and cross-entropy loss \cite{ma2021loss} to train segmentation objective $\mathcal{L}_{Seg}$. The other compared DCRL methods (Sec.\ref{sec:comparison2}) also use the same setting as this variant which adds the $\mathcal{L}_{Seg}$ in the training to learn segmentation.
\begin{table}
  \centering
  \caption{Total seven publicly available datasets are involved in this paper for the experiments of our GEMINI's variants, achieving great reproducibility.}\label{dataset}
\resizebox{\linewidth}{!}{
  \begin{tabular}{lccccccccc}
  \toprule
  \textbf{Dataset}                       &\textbf{Type}    &\textbf{Num}  &\textbf{FS-Semi} &\textbf{SS-MIP}\\
  \midrule
  ASOCA \cite{gharleghi2022automated}    &3D cardiac CT    &60            &$\surd$          &\\
  CAT08 \cite{schaap2009standardized}    &3D cardiac CT    &32            &$\surd$          &\\
  WHS-CT \cite{zhuang2019evaluation}     &3D cardiac CT    &60            &$\surd$          &\\
  CANDI \cite{kennedy2012candishare}     &3D brain MRI     &103           &$\surd$          &$\surd$\\
  SCR \cite{van2006segmentation}         &2D chest X-ray   &247           &$\surd$          &$\surd$\\
  KiPA22 \cite{he2021meta}               &3D kidney CT     &130           &                 &$\surd$\\
  ChestX-ray8 \cite{wang2017chestx}      &2D chest X-ray   &112,120       &                 &$\surd$\\
  \bottomrule
  \end{tabular}}
\end{table}

\subsubsection{Datasets} We evaluate GEMINI on three public tasks in 2D and 3D dimensions, showcasing its powerful representation ability in semi-supervised tasks \cite{you2024mine,you2024rethinking} with minimal labels (Tab.\ref{dataset}). \textbf{Task 1: FS-Semi cardiac structure segmentation (3D)} targets seven cardiac structures on 3D CT images, combining WHS-CT \cite{zhuang2019evaluation} (20 labeled, 40 unlabeled), ASOCA \cite{gharleghi2022automated} (60 unlabeled), and CAT08 \cite{schaap2009standardized} (32 labeled from\footnote{\url{http://www.sdspeople.fudan.edu.cn/zhuangxiahai/0/mmwhs/}}). Images are cropped and resampled to $144\times144\times128$, with a five-shot evaluation (5, 100, and 47 images as labeled training, unlabeled training, and testing sets). \textbf{Task 2: FS-Semi brain tissue segmentation (3D)} involves 27 brain tissues on 3D T1 MR images from the CANDI dataset \cite{kennedy2012candishare} (103 labeled). Cropped volumes of $160\times160\times128$ undergo five-shot evaluation (5, 78, and 20 images as labeled training, unlabeled training, and testing sets). \textbf{Task 3: FS-Semi chest structure segmentation (2D)} focuses on three chest-related structures in 2D chest X-rays using the SCR dataset \cite{van2006segmentation} (247 labeled) whose images are from the JSRT database \cite{shiraishi2000development}, split into 5 labeled, 142 unlabeled, and 100 testing images for five-shot evaluation. All tasks use rotation [$-20^\circ$, $20^\circ$] and scaling [0.75, 1.25] for data augmentation.

\subsubsection{Comparison setting} \label{sec:comparison2}
We compare GEMINI-Semi with 19 widely-used methods and our GVSL \cite{He_2023_CVPR} (CVPR 2023) to demonstrate its superiority. \textbf{1)} We train a U-Net \cite{ronneberger2015u} to establish upper and lower bounds using 5 labeled images (Five) and all labeled training data (Full). \textbf{2) Semi-supervised methods} without homeomorphism prior (UA-MT \cite{yu2019uncertainty}, MASSL \cite{chen2019multi}, DPA-DBN \cite{he2020dense}, CPS \cite{chen2021semi}) highlight the significance of prior knowledge for semi-supervised learning with limited labels. \textbf{3) Atlas-based methods} with homeomorphism prior (VM \cite{ba2018un}, LC-VM \cite{BalakrishnanVoxelMorph(u)}, LT-Net \cite{wang2020lt}) illustrate the limitation caused by the inefficient correspondence learning. \textbf{4) Learning registration to learn segmentation methods} with homeomorphism prior (DeepAtlas \cite{xu2019deepatlas}, DataAug \cite{zhao2019data}, DeepRS \cite{he2020deep}, PC-Reg-RT \cite{he2021few}, BRBS \cite{he2022learning}) show gains from improved correspondence but are limited by pseudo-labels from unreliable GVS. \textbf{5) Dense contrastive representation learning methods} without homeomorphism prior (VADeR \cite{o2020unsupervised}, GLCL \cite{chaitanya2020contrastive}, DSC-PM \cite{li2021dense}, PixPro \cite{xie2021propagate}, DenseCL \cite{wang2022densecl}, SetSim \cite{wang2022exploring}) reveal FP\&N problem in DCRL. For fairness, all methods use 2D/3D U-Net \cite{ronneberger2015u} with group normalization \cite{wu2018group} as the backbone.

\subsubsection{Implementation and evaluation metrics} In this task, our GEMINI-Semi is implemented by PyTorch \cite{paszke2019pytorch} on NVIDIA GeForce RTX 3090 GPUs with 24 GB memory. We take Adam whose learning rate is $1\times10^{-4}$ to optimize our framework for fast convergence. For task 1 and task 2, we sample two unlabeled images and one labeled image randomly in each iteration to save the memory for large 3D images, and for task 3, we sample 10 unlabeled images and 5 labeled images randomly in each iteration for 2D images. Following \cite{he2022learning}, we perform an affine transformation on these images via AntsPy\footnote{\url{https://github.com/ANTsX/ANTsPy}} to normalize the spatial system. We utilize the DSC [\%] to evaluate the area-based overlap index and the average Hausdorf distances (AVD) to evaluate the coincidence of the surface \cite{taha2015metrics}.

\subsection{Results and Analysis}
\label{sec:results2}
\begin{table*}
\centering
\caption{The quantitative evaluation demonstrates our powerful representation ability in FS-Semi tasks. Our GEMINI-Semi achieves the best performance on CT, MR, and X-ray images compared with 19 popular methods and the GVSL. The ``unable" means that the extremely poor results make the AVD unable to be calculated. The ``-" means that the setting is unable to be implemented. The ``HP" means these methods have or do not have homeomorphism prior. ``T1", ``T2", ``T3" are the task 1, task 2, task 3. The red and blue values are the highest and the second-highest values in the columns.}
\resizebox{\textwidth}{!}{
\begin{tabular}{clccccccccccccccc}
  \toprule
  \multirow{2}{*}{\textbf{Type}}
  &\multirow{2}{*}{\textbf{Method}}
  &\multirow{2}{*}{\textbf{HP}}
  &\multicolumn{2}{c}{\textbf{T1: 3D cardiac structures}}
  &\multicolumn{2}{c}{\textbf{T2: 3D brain tissues}}
  &\multicolumn{2}{c}{\textbf{T3: 2D chest structures}}
  &\textbf{AVG}\\ \cmidrule(r){4-5}\cmidrule(r){6-7}\cmidrule(r){8-9}\cmidrule(r){10-10}
  &
  &
  &DSC$_{\pm std}\uparrow$
  &AVD$_{\pm std}\downarrow$
  &DSC$_{\pm std}\uparrow$
  &AVD$_{\pm std}\downarrow$
  &DSC$_{\pm std}\uparrow$
  &AVD$_{\pm std}\downarrow$
  &DSC$_{\pm std}\uparrow$
  \\
  \midrule
  Full
  &U-Net \cite{ronneberger2015u}
  &$\times$
  &-
  &-
  &88.7$_{\pm1.2}$
  &0.31$_{\pm0.04}$
  &96.1$_{\pm1.4}$
  &2.28$_{\pm1.00}$
  &-
  \\
  Five
  &U-Net \cite{ronneberger2015u}
  &$\times$
  &84.3$_{\pm9.6}$
  &2.43$_{\pm2.14}$
  &69.5$_{\pm8.8}$
  &1.59$_{\pm0.84}$
  &83.4$_{\pm6.9}$
  &10.34$_{\pm4.80}$
  &79.1$_{\pm8.4}$
  \\
  \cdashline{1-10}[0.8pt/2pt]
  Semi
  &UA-MT \cite{yu2019uncertainty}
  &$\times$
  &66.4$_{\pm16.2}$
  &4.69$_{\pm2.27}$
  &75.5$_{\pm3.4}$
  &1.31$_{\pm0.95}$
  &83.9$_{\pm6.2}$
  &9.52$_{\pm4.03}$
  &75.3$_{\pm8.6}$
  \\
  &CPS \cite{chen2021semi}
  &$\times$
  &87.4$_{\pm5.4}$
  &1.40$_{\pm0.76}$
  &37.1$_{\pm1.8}$
  &unable
  &63.2$_{\pm1.4}$
  &19.57$_{\pm5.67}$
  &62.6$_{\pm2.9}$
  \\
  &MASSL \cite{chen2019multi}
  &$\times$
  &77.4$_{\pm8.7}$
  &9.07$_{\pm3.11}$
  &80.5$_{\pm3.1}$
  &0.92$_{\pm0.43}$
  &81.9$_{\pm7.0}$
  &10.99$_{\pm4.58}$
  &79.9$_{\pm6.3}$
  \\
  &DPA-DBN \cite{he2020dense}
  &$\times$
  &68.0$_{\pm14.5}$
  &5.75$_{\pm3.89}$
  &68.7$_{\pm8.2}$
  &3.90$_{\pm2.39}$
  &67.4$_{\pm8.7}$
  &24.05$_{\pm6.75}$
  &68.0$_{\pm10.5}$
  \\
  Atlas
  &VM \cite{ba2018un}
  &$\surd$
  &81.0$_{\pm6.1}$
  &2.13$_{\pm0.78}$
  &83.1$_{\pm1.8}$
  &0.56$_{\pm0.08}$
  &59.9$_{\pm5.0}$
  &15.36$_{\pm4.34}$
  &74.7$_{\pm4.3}$
  \\
  &LC-VM \cite{BalakrishnanVoxelMorph(u)}
  &$\surd$
  &81.7$_{\pm6.0}$
  &2.04$_{\pm0.77}$
  &83.0$_{\pm1.8}$
  &0.56$_{\pm0.07}$
  &60.2$_{\pm7.4}$
  &14.72$_{\pm4.89}$
  &74.9$_{\pm5.1}$
  \\
  &LT-Net \cite{wang2020lt}
  &$\surd$
  &77.8$_{\pm7.8}$
  &2.25$_{\pm0.95}$
  &82.6$_{\pm1.2}$
  &0.57$_{\pm0.05}$
  &60.4$_{\pm7.4}$
  &14.62$_{\pm4.84}$
  &73.6$_{\pm5.5}$
  \\
  LRLS
  &DeepAtlas \cite{xu2019deepatlas}
  &$\surd$
  &87.9$_{\pm4.3}$
  &1.30$_{\pm0.57}$
  &79.3$_{\pm2.6}$
  &0.74$_{\pm0.12}$
  &64.8$_{\pm9.6}$
  &12.87$_{\pm3.56}$
  &77.3$_{\pm5.5}$
  \\
  &DataAug \cite{zhao2019data}
  &$\surd$
  &82.2$_{\pm5.2}$
  &2.04$_{\pm0.73}$
  &83.9$_{\pm1.2}$
  &0.55$_{\pm0.06}$
  &22.2$_{\pm2.8}$
  &unable
  &62.8$_{\pm3.1}$
  \\
  &DeepRS \cite{he2020deep}
  &$\surd$
  &87.0$_{\pm5.0}$
  &1.60$_{\pm0.90}$
  &73.0$_{\pm5.9}$
  &0.93$_{\pm0.25}$
  &86.0$_{\pm5.6}$
  &8.55$_{\pm3.98}$
  &82.0$_{\pm5.5}$
  \\
  &PC-Reg-RT \cite{he2021few}
  &$\surd$
  &88.5$_{\pm4.9}$
  &1.23$_{\pm0.72}$
  &73.1$_{\pm3.1}$
  &1.09$_{\pm0.17}$
  &59.1$_{\pm3.6}$
  &20.71$_{\pm5.21}$
  &73.6$_{\pm3.9}$
  \\
  &BRBS \cite{he2022learning}
  &$\surd$
  &\color{blue}91.1$_{\pm3.9}$
  &\color{red}\textbf{0.93$_{\pm0.57}$}
  &\color{blue}87.2$_{\pm1.0}$
  &0.43$_{\pm0.05}$
  &71.5$_{\pm6.4}$
  &10.85$_{\pm2.99}$
  &83.3$_{\pm3.8}$
  \\
  DCRL
  &VADeR \cite{o2020unsupervised}
  &$\times$
  &85.4$_{\pm4.7}$
  &1.69$_{\pm0.77}$
  &81.2$_{\pm3.2}$
  &0.59$_{\pm0.13}$
  &79.9$_{\pm5.8}$
  &8.95$_{\pm3.37}$
  &82.2$_{\pm4.6}$
  \\
  &DenseCL \cite{wang2022densecl}
  &$\times$
  &87.3$_{\pm4.3}$
  &1.52$_{\pm0.79}$
  &83.9$_{\pm1.9}$
  &0.48$_{\pm0.09}$
  &77.1$_{\pm8.8}$
  &12.11$_{\pm6.51}$
  &82.8$_{\pm5.0}$
  \\
  &SetSim \cite{wang2022exploring}
  &$\times$
  &87.0$_{\pm4.5}$
  &1.60$_{\pm0.84}$
  &81.2$_{\pm3.0}$
  &0.58$_{\pm0.13}$
  &79.0$_{\pm7.3}$
  &11.72$_{\pm5.03}$
  &82.4$_{\pm4.9}$
  \\
  &DSC-PM \cite{li2021dense}
  &$\times$
  &87.0$_{\pm4.6}$
  &1.60$_{\pm0.86}$
  &82.6$_{\pm2.1}$
  &0.53$_{\pm0.09}$
  &85.7$_{\pm6.2}$
  &7.33$_{\pm3.32}$
  &85.1$_{\pm4.3}$
  \\
  &PixPro \cite{xie2021propagate}
  &$\times$
  &89.5$_{\pm3.9}$
  &1.31$_{\pm0.75}$
  &86.3$_{\pm1.2}$
  &\color{blue}0.38$_{\pm0.04}$
  &83.3$_{\pm8.7}$
  &8.73$_{\pm4.55}$
  &\color{blue}86.4$_{\pm4.6}$
  \\
  &GLCL\cite{chaitanya2020contrastive}
  &$\times$
  &84.5$_{\pm7.0}$
  &1.82$_{\pm1.09}$
  &83.0$_{\pm2.7}$
  &0.52$_{\pm0.11}$
  &85.5$_{\pm8.9}$
  &8.65$_{\pm5.18}$
  &84.3$_{\pm6.2}$
  \\
  \cdashline{1-10}[0.8pt/2pt]
  \textbf{DCRL}
  &\textbf{GVSL-Semi (CVPR)} \cite{He_2023_CVPR}
  &$\surd$
  &90.0$_{\pm3.7}$
  &1.21$_{\pm0.81}$
  &82.3$_{\pm5.9}$
  &0.55$_{\pm0.27}$
  &\color{blue}86.3$_{\pm5.5}$
  &\color{blue}7.18$_{\pm4.01}$
  &86.2$_{\pm5.0}$
  \\
  \textbf{(Ours)}
  &\textbf{GEMINI-Semi}
  &$\surd$
  &\color{red}\textbf{91.2$_{\pm3.6}$}
  &\color{blue}0.97$_{\pm0.56}$
  &\color{red}\textbf{87.3$_{\pm1.0}$}
  &\color{red}\textbf{0.35$_{\pm0.03}$}
  &\color{red}\textbf{87.7$_{\pm5.2}$}
  &\color{red}\textbf{7.14$_{\pm3.63}$}
  &\color{red}\textbf{88.7$_{\pm3.3}$}
  \\
  \bottomrule
\end{tabular}
}
\label{tab:metrics2}
\end{table*}
\begin{figure}
  \centering
  \includegraphics[width=\linewidth]{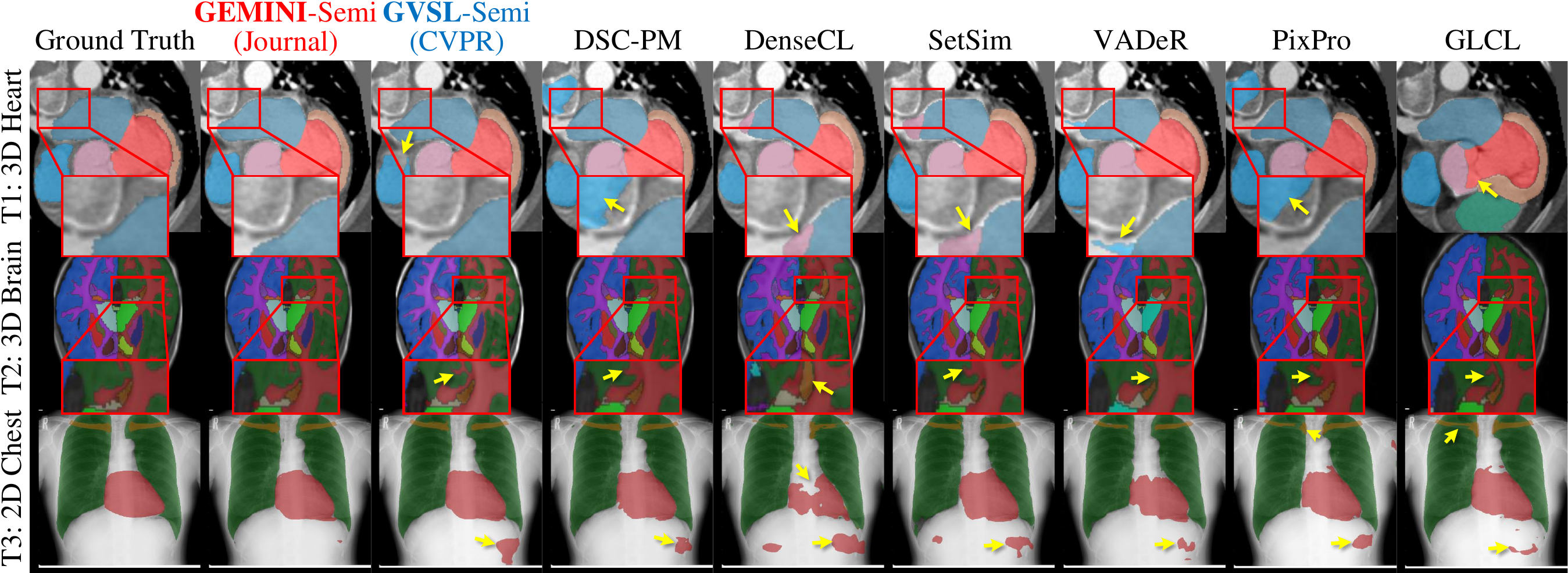}
  \caption{Our GEMINI-Semi has significant visual superiority on three FS-Semi medical image segmentation tasks.}\label{Fig:results2}
\end{figure}
\subsubsection{Quantitative evaluation shows metric superiority}
As shown in Tab.\ref{tab:metrics2}, 19 compared methods demonstrate that the DCRL will greatly improve the representability, and the homeomorphism prior (``HP") further improves the reliability of the representation learning. There are three interesting observations in Tab.\ref{tab:metrics2}: \textbf{1)} The semi-supervised methods are limited by the extremely few labels. They utilize the pseudo-label generation (UA-MT, CPS) or multi-task learning (MASSL, DPA-DBN) to improve the representation, but the extremely few labels have no enough ability to give them reliable optimization directions to overcome the noise in pseudo labels or multiple tasks. As a result, the UA-MT, MASSL, and DPA-DBN have worse performance than U-Net on task 1, and the CPS is worse on task 2 and 3. \textbf{2)} With the ``HP", the Atlas and LRLS methods achieve robust performance in task 1 and task 2, but are limited in task 3. The ``HP" brings an alignment between labeled and unlabeled images for numerous reliable pseudo labels. Therefore, they have achieved significant improvement on task 1 and task 2 compared with the semi-supervised methods. However, the X-ray images in task 3 have low contrast and their appearances are varied caused by the 2D projection of 3D human body, this makes inefficient GVS brings large misalignment between images, thus interfering with the learning and reducing the performance. \textbf{3)} The DCRL methods have robust performance in all three tasks compared with the LRLS methods, although the VADeR, DenseCL, SetSim, DSC-PM, PixPro and GLCL have no homeomorphism prior. Because their feature-level learning reduce the direct interference caused by misalignment in LRLS's pseudo labels and the supervision from the few labels bring basic representability which will promote their correspondence discovery. However, the FP\&N problem is still a problem in the learning and their performance on task 3 is poor without ``HP" like the semi-supervised methods.

Compared with the LRLS, other DCRL methods, and our previous GVSL-Semi, our GEMINI-Semi achieves the best performance on three tasks with four observations: \textbf{1)} Compared with the LRLS methods which have ``HP", our method has better performance on all tasks. Although the BRBS has similar performance as our GEMINI-Semi on task 1 and task 2, our method achieves 16.2\% DSC and 3.71 AVD higher and lower than it on task 3. This is because our GEMINI-Semi utilizes our GSS for alignment measurement and shares the representation between the segmentation and deformation learning, bringing more efficient and robust learning for alignment. It has a great ability to construct positive feature pairs even with varied appearances. The gradient from our DHL also trains the soft negative feature pairs to drive the learning of distinct representations for potentially different semantics in shared backbones, bringing a regularization for potential mispaired positive pairs. \textbf{2)} Compared with the other DCRL methods which have no ``HP", our GEMINI-Semi shows great improvements in all three tasks. It achieves more than 1.7\%, 1.0\%, and 2.0\% DSC improvements on task 1, 2, and 3 compared with the best DCRL models without ``HP" (PixPro in task 1 and 2, DSC-PM in task 3). Because the ``HP" in our GEMINI-Semi constructs a more reliable correspondence discovery process which reduces the production risk of the FP\&N pairs, bringing comprehensive improvement for the DCRL. \textbf{3)} Compared to our CVPR vision (GVSL-Semi), we find even though the GVSL utilizes the visual similarity like the BRBS, it also achieves great performance in task 3, demonstrating the superiority of the DCRL paradigm. The GVSL-semi avoids the interference of pseudo labels like BRBS reducing the noisy information from the extremely mis-alignment, so that it takes the advantage of DCRL and our homeomorphism prior and achieves good performance in all three tasks. Our GEMINI-Semi promotes the GVSL and utilizes the GSS for a more powerful dense representation learning, thus achieving the highest 88.7\% average DSC in this experiment. \textbf{4)} Compared with the fully supervised setting (``Full") in task 2 (83 labeled images), our GEMINI-Semi achieves a similar performance only with 5 labeled images demonstrating our great potential in reducing of annotation costs. In the task 3, our framework is lower than the upper bound (96.1\%) only with five annotations, but it still achieves significant improvement (4.3\%) compared with the model directly trained on five labeled images.

\subsubsection{Qualitative evaluation shows visual superiority}
As shown in Fig.\ref{Fig:results2}, we show typical cases on the three tasks in this experiment and our framework has higher accuracy on thin regions and fewer outliers. In the task 1, the segmentation result of our method has better integrity, and the different semantic structures have good adjacency without outliers. However, the other four DCRL methods have discontinuous mis-segmentation which destroys the heart topology. This is because the pairing strategies in the DCRL methods are unable to make the pairs under the condition of topology consistency, so the large-scale mispaired features interrupt the learning and make numerous outliers. The same as the task 3, there are also serious outlier problems in the four typical DCRL methods and the GVSL, and our GEMINI-Semi has fine segmentation. In the task 2, our GEMINI and GVSL show finer segmentation on the thin brain structures which is sensitive and will be interrupted by the noise in the semi-supervised learning process. In some prominent and gully regions of task 2 (enlarged part), the compared four DCRL methods have numerous distortions due to their unreliable correspondence discovery, showing their fragility.

\section{Experiment 2: Self-supervised Medical Image Pre-training (SS-MIP)}
\label{sec:task1}
We further implement our GEMINI learning on self-supervised medical image pre-training (SS-MIP) task (GEMINI-MIP), providing powerful tools to transfer potential tasks and giving complete experiments. Four public-available datasets are enrolled in our experiments for very effective evaluations.

\begin{table*}
\centering
\caption{The fine-tuning evaluations demonstrate our great transferring ability on SS-MIP tasks. Our GEMINI-MIP achieves the best performance compared with 18 methods on 3 downstream tasks. ``T1", ``T2", and ``T3" are the task 1, task 2, and task 3. ``AVG" is the average value of the row. The cells with gray backgrounds are the inner-scene (same image category) transferring and the others are the inter-scene (different image category) transferring.}
\begin{tabular}{clcccccccccc}
\toprule
\multirow{2}{*}{\textbf{Type}}
&\multirow{2}{*}{\textbf{Pre-training}}
&\multicolumn{2}{c}{\textbf{T1: SCR$_{25}$} \emph{Inner-scene}}
&\multicolumn{2}{c}{\textbf{T2: KiPA22} \emph{Inter-scene}}
&\multicolumn{2}{c}{\textbf{T3: CANDI} \emph{Inter-scene}}
&\textbf{AVG}
\\
\cmidrule(r){3-4}\cmidrule(r){5-6}\cmidrule(r){7-8}\cmidrule(r){9-9}
&
&DSC$_{\pm std}\uparrow$
&AVD$_{\pm std}\downarrow$
&DSC$_{\pm std}\uparrow$
&AVD$_{\pm std}\downarrow$
&DSC$_{\pm std}\uparrow$
&AVD$_{\pm std}\downarrow$
&DSC$_{\pm std}\uparrow$
\\
\midrule
-
&Scratch (2D U-Net)
&81.8$_{\pm8.2}$
&9.00$_{\pm6.37}$
&74.1$_{\pm12.3}$
&3.59$_{\pm1.97}$
&65.0$_{\pm4.4}$
&1.27$_{\pm0.21}$
&73.6$_{\pm8.3}$
\\
\cdashline{1-9}[0.8pt/2pt]
Sup
&ImageNet \cite{deng2009imagenet}
&\color{blue}92.0$_{\pm3.1}$
&\color{red}\textbf{4.09$_{\pm1.64}$}
&72.6$_{\pm15.5}$
&4.78$_{\pm4.86}$
&71.1$_{\pm19.8}$
&1.35$_{\pm2.07}$
&78.6$_{\pm12.8}$
\\
GRL
&Denosing \cite{vincent2010stacked}
&\cellcolor[gray]{0.9}83.9$_{\pm7.8}$
&\cellcolor[gray]{0.9}11.17$_{\pm7.81}$
&60.3$_{\pm17.7}$
&7.55$_{\pm5.18}$
&67.7$_{\pm2.1}$
&1.21$_{\pm0.08}$
&70.6$_{\pm9.2}$
\\
&In-painting \cite{pathak2016context}
&\cellcolor[gray]{0.9}85.1$_{\pm6.6}$
&\cellcolor[gray]{0.9}16.59$_{\pm10.74}$
&64.4$_{\pm16.4}$
&5.79$_{\pm4.13}$
&66.2$_{\pm2.3}$
&1.26$_{\pm0.08}$
&71.9$_{\pm8.4}$
\\
&Models Genesis \cite{zhou2019models}
&\cellcolor[gray]{0.9}86.1$_{\pm4.6}$
&\cellcolor[gray]{0.9}6.22$_{\pm2.27}$
&66.6$_{\pm16.3}$
&5.86$_{\pm3.14}$
&88.1$_{\pm3.1}$
&0.32$_{\pm0.10}$
&80.3$_{\pm8.0}$
\\
&Rotation \cite{komodakis2018unsupervised}
&\cellcolor[gray]{0.9}80.5$_{\pm7.7}$
&\cellcolor[gray]{0.9}20.62$_{\pm12.55}$
&69.7$_{\pm15.3}$
&6.45$_{\pm4.33}$
&78.3$_{\pm2.6}$
&0.75$_{\pm0.09}$
&76.2$_{\pm8.5}$
\\
CRL
&SimSiam \cite{Chen2021CVPR}
&\cellcolor[gray]{0.9}87.2$_{\pm5.1}$
&\cellcolor[gray]{0.9}11.87$_{\pm8.10}$
&72.6$_{\pm13.3}$
&4.10$_{\pm3.25}$
&76.7$_{\pm2.1}$
&0.82$_{\pm0.06}$
&78.8$_{\pm6.8}$
\\
&BYOL \cite{grill2020bootstrap}
&\cellcolor[gray]{0.9}89.4$_{\pm4.9}$
&\cellcolor[gray]{0.9}8.48$_{\pm4.37}$
&74.1$_{\pm12.6}$
&3.87$_{\pm2.93}$
&70.5$_{\pm2.1}$
&1.08$_{\pm0.07}$
&78.0$_{\pm6.5}$
\\
&SimCLR \cite{chen2020simple}
&\cellcolor[gray]{0.9}89.0$_{\pm4.0}$
&\cellcolor[gray]{0.9}11.28$_{\pm6.53}$
&74.4$_{\pm11.3}$
&3.68$_{\pm2.65}$
&79.0$_{\pm2.6}$
&1.02$_{\pm0.40}$
&80.8$_{\pm6.0}$
\\
&MoCov2 \cite{chen2020improved}
&\cellcolor[gray]{0.9}84.3$_{\pm6.5}$
&\cellcolor[gray]{0.9}11.06$_{\pm5.19}$
&69.6$_{\pm14.4}$
&6.28$_{\pm4.70}$
&82.9$_{\pm3.6}$
&0.52$_{\pm0.12}$
&78.9$_{\pm8.2}$
\\
&DeepCluster \cite{caron2018deep}
&\cellcolor[gray]{0.9}84.0$_{\pm8.1}$
&\cellcolor[gray]{0.9}19.71$_{\pm13.15}$
&72.7$_{\pm15.1}$
&4.91$_{\pm3.50}$
&60.0$_{\pm2.2}$
&1.52$_{\pm0.07}$
&72.2$_{\pm8.5}$
\\
DCRL
&VADeR \cite{o2020unsupervised}
&\cellcolor[gray]{0.9}85.2$_{\pm5.1}$
&\cellcolor[gray]{0.9}7.15$_{\pm3.05}$
&62.8$_{\pm15.6}$
&7.23$_{\pm4.73}$
&86.1$_{\pm3.4}$
&0.40$_{\pm0.12}$
&78.0$_{\pm8.0}$
\\
&DenseCL \cite{wang2022densecl}
&\cellcolor[gray]{0.9}85.0$_{\pm6.3}$
&\cellcolor[gray]{0.9}11.74$_{\pm7.02}$
&70.8$_{\pm14.8}$
&5.48$_{\pm3.95}$
&76.8$_{\pm2.9}$
&1.22$_{\pm0.68}$
&77.5$_{\pm8.0}$
\\
&SetSim \cite{wang2022exploring}
&\cellcolor[gray]{0.9}85.2$_{\pm5.1}$
&\cellcolor[gray]{0.9}9.63$_{\pm7.64}$
&70.8$_{\pm14.4}$
&4.92$_{\pm3.26}$
&74.9$_{\pm2.5}$
&0.89$_{\pm0.08}$
&77.0$_{\pm7.3}$
\\
&DSC-PM \cite{li2021dense}
&\cellcolor[gray]{0.9}90.5$_{\pm3.5}$
&\cellcolor[gray]{0.9}5.44$_{\pm2.87}$
&77.2$_{\pm12.2}$
&3.87$_{\pm3.34}$
&83.3$_{\pm2.4}$
&0.74$_{\pm0.64}$
&83.7$_{\pm6.0}$
\\
&PixPro \cite{xie2021propagate}
&\cellcolor[gray]{0.9}91.5$_{\pm3.3}$
&\cellcolor[gray]{0.9}9.83$_{\pm5.34}$
&73.6$_{\pm12.9}$
&4.00$_{\pm3.33}$
&63.9$_{\pm2.0}$
&1.35$_{\pm0.06}$
&76.3$_{\pm6.1}$
\\
&GLCL \cite{chaitanya2020contrastive}
&\cellcolor[gray]{0.9}87.3$_{\pm5.8}$
&\cellcolor[gray]{0.9}9.35$_{\pm4.68}$
&76.5$_{\pm11.9}$
&4.33$_{\pm2.94}$
&82.8$_{\pm2.6}$
&0.56$_{\pm0.09}$
&82.2$_{\pm6.8}$
\\
\cdashline{1-9}[0.8pt/2pt]
\textbf{DCRL}
&\textbf{GVSL-MIP (CVPR)}\cite{He_2023_CVPR}
&\cellcolor[gray]{0.9}89.7$_{\pm3.7}$
&\cellcolor[gray]{0.9}10.52$_{\pm7.23}$
&\color{blue}78.9$_{\pm11.2}$
&\color{red}\textbf{2.95$_{\pm1.55}$}
&\color{blue}89.7$_{\pm2.6}$
&\color{blue}0.27$_{\pm0.08}$
&\color{blue}86.1$_{\pm5.8}$
\\
\textbf{(Ours)}
&\textbf{GEMINI-MIP}
&\color{red}\cellcolor[gray]{0.9}\textbf{92.1$_{\pm2.8}$}
&\cellcolor[gray]{0.9}\color{blue}5.38$_{\pm2.65}$
&\color{red}\textbf{79.1$_{\pm11.1}$}
&\color{blue}3.22$_{\pm2.24}$
&\color{red}\textbf{89.8$_{\pm2.6}$}
&\color{blue}0.27$_{\pm0.08}$
&\color{red}\textbf{87.0$_{\pm5.5}$}
\\
\bottomrule
\end{tabular}
\label{tab:metrics1}
\end{table*}

\subsection{Experiments configurations}
\label{sec:configurations1}
\subsubsection{Variant design}
Our GEMINI-MIP task is the pretext task to pre-train the representation of the backbone network $\mathcal{N}_{\theta}$ and then the pre-trained network is transferred to the downstream tasks $\mathcal{L}_{DS}$. It also learns a self-restoration head $Res_{\tau}$ (fundamental task) on the dense features $f^{A},f^{B}$ due to the initial weak representation in the pretext task for a warm-up of our GSS. The self-restoration is based on the prior of edges and shapes in images and trains the network to capture the features from the broken distribution. Therefore, the variant framework for SS-MIP has an additional optimization for restoration $\mathcal{L}_{Res}$:
\begin{align}\label{equ:variant1}
&\textbf{\text{Pretext:}}\ \underset{\xi,\theta,\tau}{\arg\min}\ (\mathcal{L}_{DHL}(\theta,\xi,\mathcal{S}_{ul})+\mathcal{L}_{Res}(\theta,\tau,\mathcal{S}^{*}_{ul}),\notag\\
&\textbf{\text{Downstream:}}\ \underset{\kappa}{\arg\min}\ \mathcal{L}_{DS}(\theta,\kappa,\mathcal{S}_{l}),
\end{align}
where the $\mathcal{S}_{ul}$ is the unlabeled dataset, and the $\mathcal{S}^{*}_{ul}$ is the unlabeled dataset with the appearance transformation $\mathcal{T}(\mathcal{S}_{ul})=\mathcal{S}^{*}_{ul}$ for self-restoration, the $\mathcal{S}_{l}$ is the labeled dataset in the downstream task, $\mathcal{L}_{DS}$ is the loss for the downstream task, and the $\kappa$ is the parameters in the learning head of the downstream task. In our experiment, we utilize the mean square error as the loss for the self-restoration following \cite{zhou2019models}, $\mathcal{L}_{Res}(x,\mathcal{T}(x))=|x-Res_{\tau}(\mathcal{N}_{\theta}(\mathcal{T}(x)))|^{2}$, to train self-restoration objective. We utilize the random in-painting, local-shuffling, and non-linear transformation in the $\mathcal{T}$ to transform the unlabeled images.

\subsubsection{Datasets} We evaluate the representation learning ability of our GEMINI-MIP on four datasets with three downstream tasks to demonstrate the properties and advantages of our method in different aspects.

\textbf{Pretext datasets:} We utilize the \emph{ChestX-ray8} \cite{wang2017chestx} which has 112,120 frontal-view chest X-ray images with $1024\times1024$ resolution and 0 to 255 grayscale values. 44,810 of them are scanned from the anterior to posterior (AP) view and 67,310 of them are scanned from the posterior to anterior (PA) view. 63,340 of them are male and 48,780 of them are female. In our experiment, we resize the images into $512\times512$ and normalize them to [0, 1]. For a better homeomorphic property, we randomly pair these chest X-ray images with the same perspective (PA/AP) and gender (male/female).

\textbf{Downstream datasets:} Three publicly available datasets (SCR \cite{van2006segmentation}, KiPA22 \cite{he2021meta}, CANDI \cite{kennedy2012candishare}) are used to demonstrate the superiorities of our framework. \textbf{Task 1: SCR} dataset \cite{van2006segmentation} segments 3 chest-related structures on 247 X-ray images. We set 47 of them as the validation set, 100 of them as the training set, and 100 of them as the testing set. We utilize 25\% of the training set in this experiment (SCR$_{25}$) to build a limited data situation and more data amount evaluations are performed in our analysis (Sec.\ref{sec:discussion}). \textbf{Task 2: KiPA22} dataset \cite{he2021meta} segments four renal cancer-related structures on 130 3D CT images. We set 30 of them as the validation set, 70 of them as the training set, and 30 of them as the testing set. \textbf{Task 3: CANDI} dataset \cite{kennedy2012candishare} segments 28 brain tissues on 103 3D T1 MR images. We set 20 of them as the validation set, 40 of them as the training set, and 43 of them as the testing set. For the 3D datasets used in the 2D task, we train the networks on the 2D slices of the images, predict segmentation results at each 2D slice and evaluate the results for 3D volumes.

\subsubsection{Comparison setting} We benchmark GEMINI against 17 state-of-the-art or widely-used methods across four categories and compare with our previous GVSL to highlight advancements. \textbf{1)} A 2D supervised network pre-trained on ImageNet \cite{deng2009imagenet} evaluates the representation ability of supervised learning. \textbf{2)} Generative representation learning (GRL) methods (Denoising \cite{vincent2010stacked}, In-painting \cite{pathak2016context}, Models Genesis \cite{zhou2019models}, Rotation \cite{komodakis2018unsupervised}) provide baseline performance for classic approaches. \textbf{3)} Contrastive representation learning (CRL) methods (SimSiam \cite{Chen2021CVPR}, BYOL \cite{grill2020bootstrap}, SimCLR \cite{chen2020simple}, MoCov2 \cite{chen2020improved}, DeepCluster \cite{caron2018deep}) reveal limitations of global contrastive representation on MIDP tasks. \textbf{4)} Dense contrastive representation learning (DCRL) methods (VADeR \cite{o2020unsupervised}, GLCL \cite{chaitanya2020contrastive}, DSC-PM \cite{li2021dense}, PixPro \cite{xie2021propagate}, DenseCL \cite{wang2022densecl}, SetSim \cite{wang2022exploring}) highlight GEMINI's superior performance through reliable positive and negative pair learning. For downstream tasks, pre-trained 2D CNN feature extractors (backbone $\mathcal{N}_{\theta}$) are used in a 2D U-Net \cite{ronneberger2015u} (introduced in Sec.\ref{sec:task2}), trained with a combined Dice and cross-entropy loss ($\mathcal{L}_{DS}$).

\subsubsection{Implementation and evaluation metrics} All tasks are implemented by PyTorch \cite{paszke2019pytorch} on NVIDIA GeForce RTX 3090 GPUs with 24 GB memory, optimized by Adam \cite{kingma2014adam} whose learning rate is $10^{-4}$. The pretext task is trained with $2\times10^{5}$ iterations. We utilize the fine-tuning evaluation on the downstream tasks. The downstream tasks are trained with $4\times10^{4}$ iterations and validated every 200 iterations to save the best models on their validation sets. For a fair comparison, all methods in our experiment take the same basic training setting. Ten 2D X-ray images are randomly sampled in each iteration for stable pre-training, and five 2D images are randomly sampled in each iteration of downstream transferring. We use the same metrics as our experiment 1 (Sec.\ref{sec:task2}) for the evaluation of the performance.

\subsection{Results and Analysis}
\label{sec:results1}
\begin{figure}
  \centering
  \includegraphics[width=\linewidth]{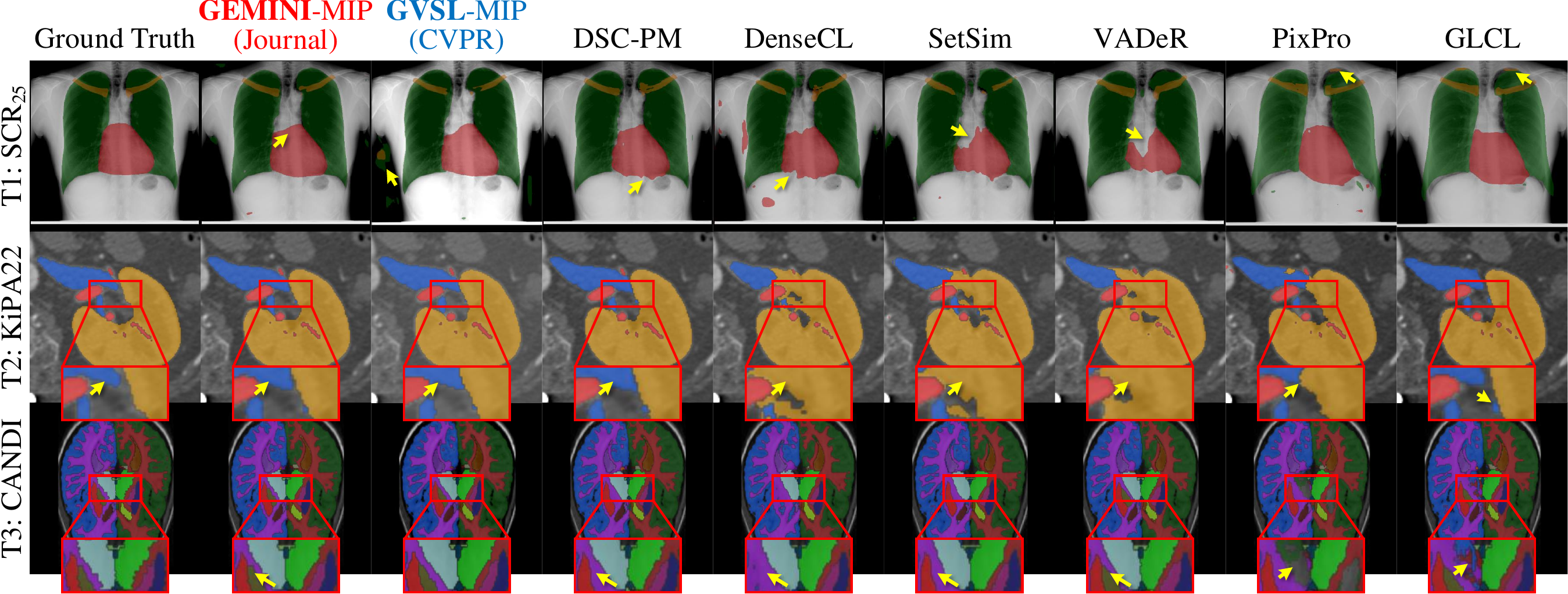}
  \caption{Our GEMINI-MIP also has very significant visual superiority in the three downstream tasks.}\label{Fig:results1}
\end{figure}

\subsubsection{Quantitative evaluation shows metric superiority}
As shown in Tab.\ref{tab:metrics1}, the fine-tuning evaluation demonstrates the great transferring ability of our GEMINI-MIP due to the reliable positive and negative pairs discovery promoted by our homeomorphism prior. We can find two interesting observations in Tab.\ref{tab:metrics1}: \textbf{1)} The pre-training will bring better performance than random initialization (``Scratch") to most of the networks. This is because the learned representation from the pretext task with large-scale data will stimulate the network to learn diverse low-level patterns, although the FN or FP problem will interrupt the representation learning of high-level semantics, the diverse low-level patterns' knowledge will promote the transferring. \textbf{2)} Due to the large interference of FP or FN problem, the CRL and DCRL methods all have weaker performance than the network pre-trained by ImageNet in task 1. This is because the FP or FN problem interrupts the representation learning of high-level semantics and makes their representations deviate from reality. Therefore, even though these self-supervised pre-trained networks have improved the learning of downstream tasks, their upper limit is extremely limited.

Compared with the other DCRL methods and the conference vision (GVSL), our GEMINI-MIP achieves the best performance with three observations: \textbf{1)} Our GEMINI-MIP has great MIP ability with the highest DSC (92.1\%, 79.1\%, 89.8\%) in all tasks. Although the DSC-PM also achieves great performance in these three tasks (90.5\%, 77.2\%, 83.3\% DSC), its average DSC is still lower than our GVSL-MIP and GEMINI-MIP owing to the interference of FP\&N. Our homeomorphism prior brings reliable correspondence discovery and significantly weakens the FP\&N problem, thus greatly improving the representation. \textbf{2)} Our GEMINI brings significant improvements in both inner-scene and inter-scene transferring tasks. It achieves a very competitive 92.1\% DSC on task 1, and the best score (79.1\%, 89.8\% DSC) in the other inter-scene transferring tasks. The PixPro has reasonable performance on inner-scene tasks (91.5\% DSC), but it only has 73.6\% and 63.8\% DSC on the inter-scene tasks which is much lower than our framework. Because the reliable positive and negative pairs in our framework enable the network to pre-learn both low-level patterns and high-level semantics, this makes the pre-learned knowledge match the reality showing greater transferring ability. \textbf{3)} The GVSL-MIP has achieved similar performance as our GEMINI-MIP in inter-scene transferring because of its geometric visual similarity which will learn soft negative pairs. Our GEMINI-MIP further takes the geometric semantic similarity and has achieved 2.4\% DSC improvement in inner-scene transferring.

\subsubsection{Qualitative evaluation shows visual superiority}
As shown in Fig.\ref{Fig:results1}, the visualization of the segmentation results demonstrates our superiority in the SS-MIP tasks. Due to our reliable correspondence discovery, the pre-training process makes the network represent consistent and distinct features for the same and different semantic regions, having very effective initialization. Therefore, our GEMINI-MIP has a very fine visualization than the compared DCRL methods and conference vision, GVSL-MIP. In task 3, the VADeR, DenseCL, and SetSim lose some very small brain structures which are sensitive and easy to be interfered with by the misguidance from the FP\&N problem. The VADeR also have very poor performance on mixed kidney region in the tasks due to the FN problem which enlarges the network's challenge to distinguish these complex regions.

\section{Discussion and Analysis}
\label{sec:discussion}
\subsection{Framework analysis}
\subsubsection{Ablation study shows improvements of components}
\begin{figure}
  \centering
  \includegraphics[width=\linewidth]{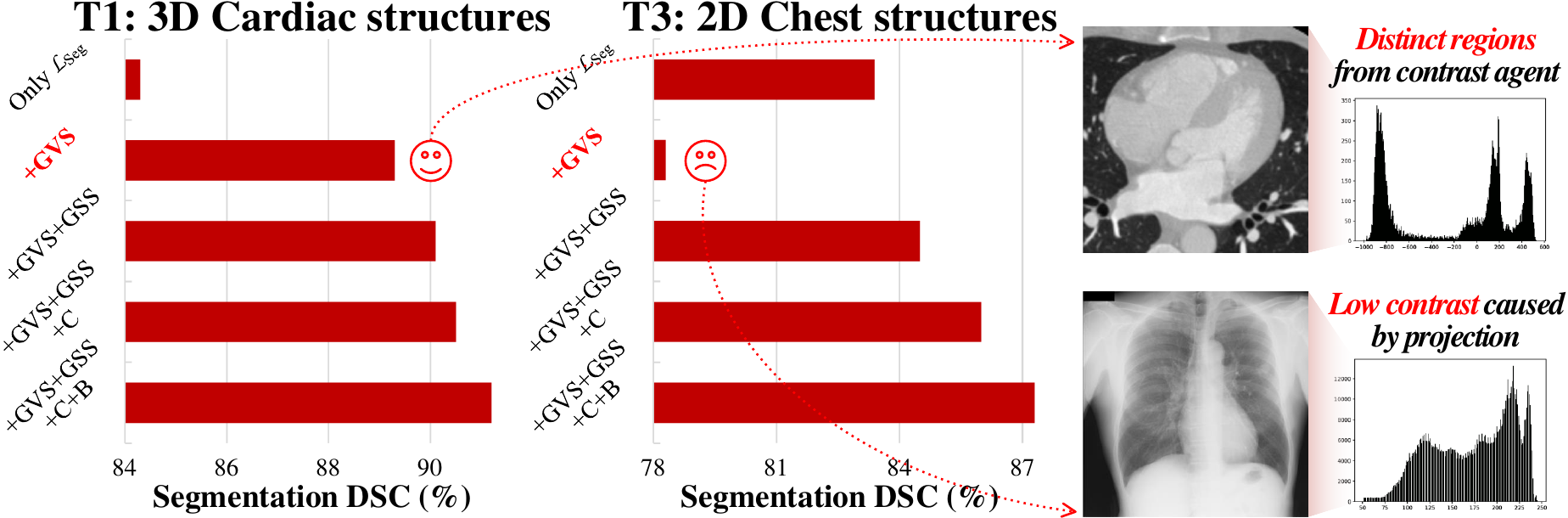}
  \caption{The ablation studies on the ``T1: 3D Cardiac structures" and ``T3: 2D Chest structures" demonstrate the great contributions of the components in our framework. The ``C" and ``B" are the learning for continuity and bijection.}
  \label{Fig:ablation}
\end{figure}
The ablation studies on the ``T1: 3D Cardiac structures" and ``T3: 2D Chest structures" demonstrate the great contributions of our innovations. It has two observations: \textbf{1)} The innovations in our framework all contribute to the performance. When we add the GVS and GSS into the correspondence learning, the two tasks all achieve very significant improvement compared with the direct segmentation learning on few labeled images. The smoothness loss $\mathcal{L}_{smo}$ for the continuity further improves the deformation accuracy and the smoothness (topology-preservation ability), so that the two tasks all achieve further improvement. Finally, when adding the loss for bijection, these tasks all obtain the highest segmentation DSC. This illustrates that learning under the condition of medical images' topology will improve the representation of the backbone network $\mathcal{N}_{\theta}$, bringing better performance on target tasks. \textbf{2)} Compared between ``T1" and ``T3", their learning only with GVS have very different results due to the appearance of their images. The learning with GVS in T1 improves the segmentation performance, but it in T3 extremely weakens the performance. Because the cardiac CT images in T1 are enhanced by contrast agents, they have distinct regions and will provide guidance to learn the correspondence. However, the chest X-ray images in T3 are projected from 3D human body and have very low contrast, so they are unable to measure the alignment degree and interfere with the correspondence learning. When adding our GSS, the performances in these two tasks are all improved. Because the measurement of the features will avoid interference from the appearance limitation, thus achieving better optimization guidance for correspondence.
\subsubsection{Analysis of hyper-parameters}
\begin{figure}
  \centering
  \includegraphics[width=\linewidth]{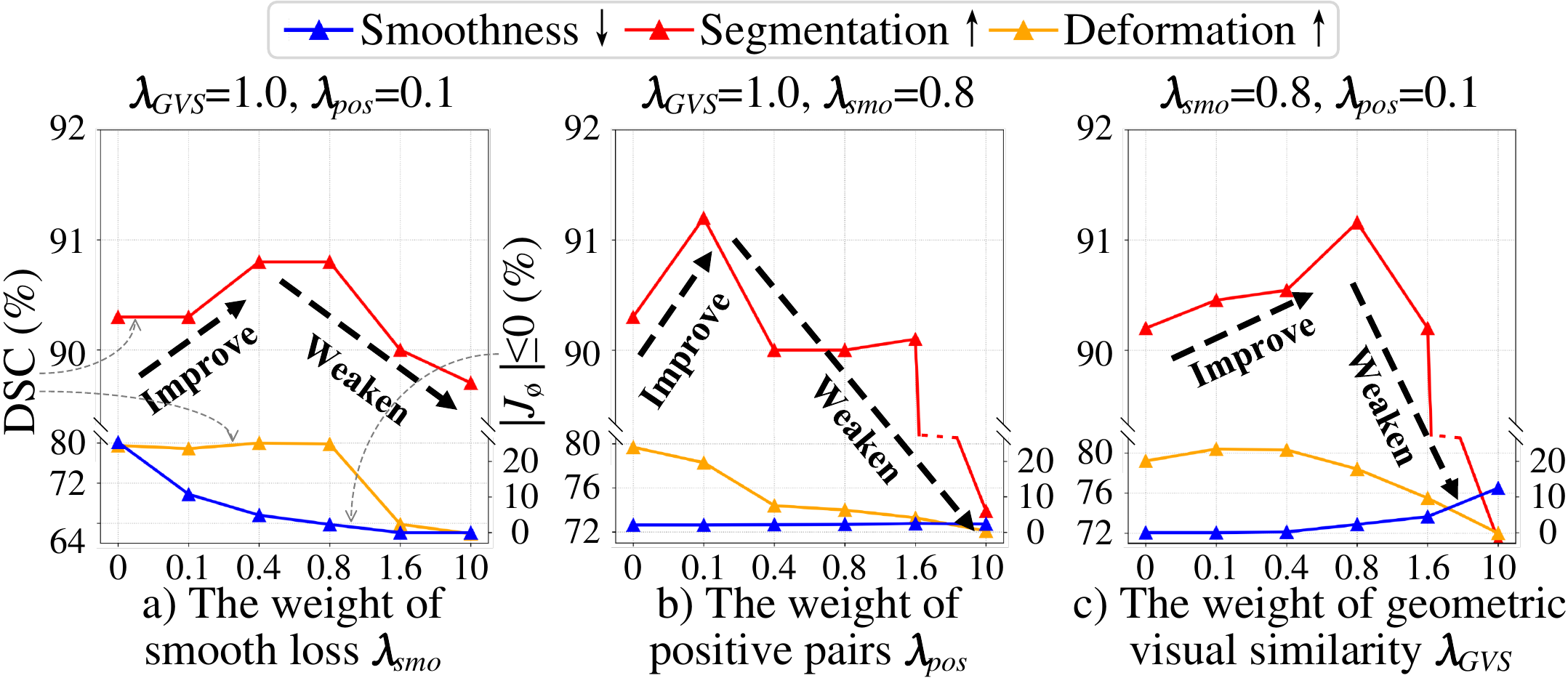}
  \caption{The ablation of the hyper-parameters on the ``T1: 3D Cardiac structures" show the effects from the weight of the smoothness loss $\lambda_{smo}$, the positive pairs $\lambda_{pos}$, and the GVS loss $\lambda_{GVS}$. The $|J_{\phi}|\leq0$ (\%) is  Jacobian matrix  \cite{he2021few} which evaluates smoothness of the deformation.}\label{Fig:hyper}
\end{figure}
We analyze the three hyper-parameters in our framework, i.e., the weight of the smoothness loss $\lambda_{smo}$, the positive pairs  $\lambda_{pos}$ and the GVS loss $\lambda_{GVS}$. With the enlarging of these three hyper-parameters, the segmentation performances of our framework are all improved and then weakened. Because: \textbf{a)} The smoothness loss improves the topology preservation ability and reduces the deformation degree. When the $\lambda_{smo}$ is small, the deformation accuracy (orange line) and topology preservation degree (blue line) are improved, promoting the reliability of correspondence. However, when the $\lambda_{smo}$ is large, excessive smoothing reduces the deformation accuracy, weakening the segmentation. \textbf{b)} The learning of positive pairs clusters the features in corresponding positions, making better representation for the same semantic features. However, when the $\lambda_{pos}$ is too large, the gradient from this loss will be much larger than the optimization for the negative pair ``OPT$()$", which makes the model tend to represent all features as consistent reducing their discrimination. The positive pairs will be further analyzed in Sec.\ref{sec:positive}. \textbf{c)} The training-free property of GVS improves the deformation accuracy and stabilizes the correspondence training when the $\lambda_{GVS}$ is small. However, the problem of appearance variation in the images is enlarged when the $\lambda_{GVS}$ is large, so the unreliable similarity will give an inaccurate optimization target, weakening the deformation accuracy and smoothness, and reducing the final segmentation performance. 

\subsubsection{Analysis of deformation in learning process}
\begin{figure}
  \centering
  \includegraphics[width=\linewidth]{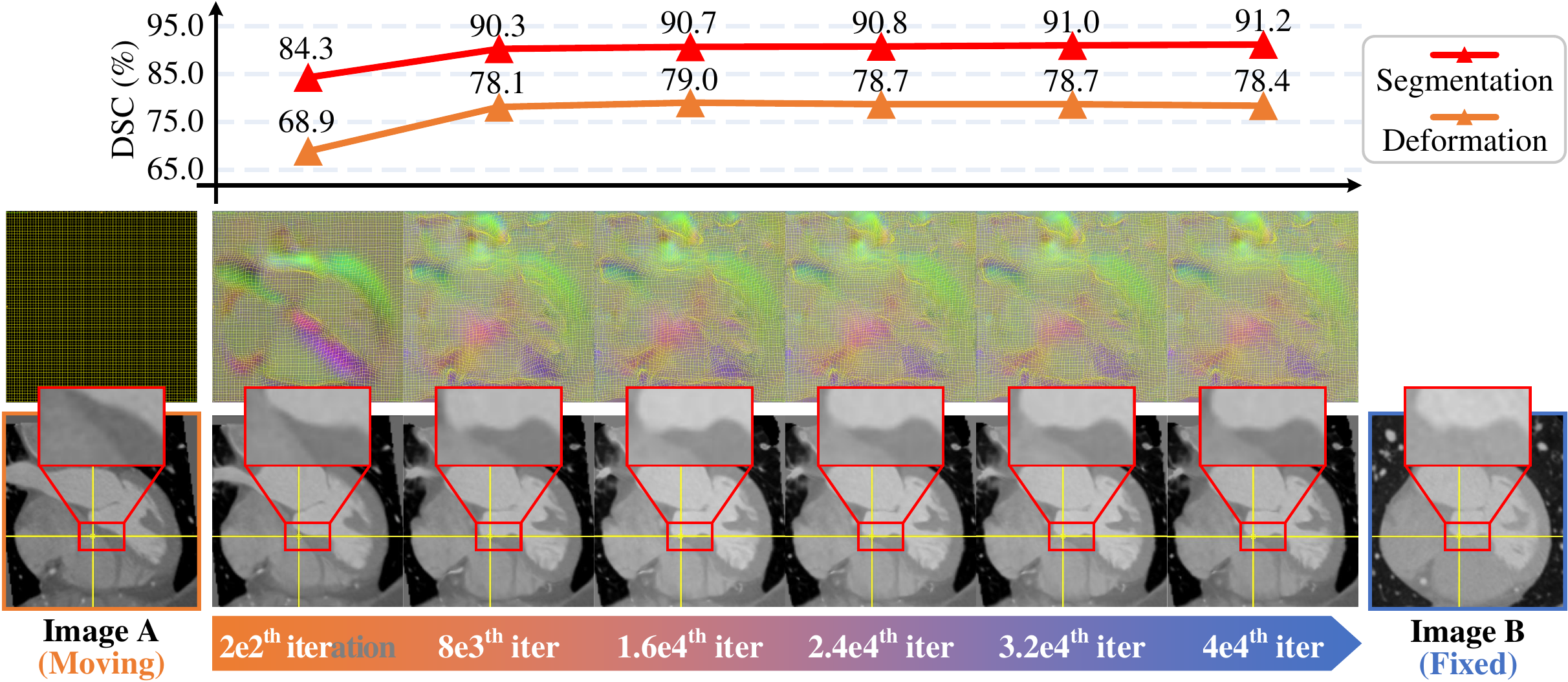}
  \caption{The visualization of the deformation in the learning process on the ``T1: 3D Cardiac structures". The first row is the line chart of the segmentation and deformation performance. The second row is the grids which demonstrates the deformation degree. The third row is the deformed image A during the learning process.}\label{Fig:registration}
\end{figure}
As shown in Fig.\ref{Fig:registration}, the deformation in our GEMINI-Semi demonstrates the great correspondence of the semantic regions between images. During the training, our model will quickly learn the correspondence of semantic regions between images in the beginning, so the deformation DSC is 78.1\% and the segmentation DSC is 90.3\% in the $8\times10^{3}$th iteration (1/5 of all iteration amount). These scores in this iteration have been very close to the final scores (78.4\% and 91.2\%). Visually, the deformed image in the $8\times10^{3}$th iteration also has a very high alignment degree to the target image B. In the later learning iterations, the segmentation DSC is improving slightly and up to 91.2\%, although there is no improvement in the deformation DSC. This is because the learn reliable correspondence is still promoting the representation learning via gradient, and the great alignment provides the reliable learning of positive pairs.
\subsubsection{Analysis of positive pairs $\mathcal{L}_{pos}$}
\label{sec:positive}
\begin{table}
\centering
\caption{The learning for positive pairs $\mathcal{L}_{pos}$ in different setting on the ``T1: 3D Cardiac structures". The ``No $\mathcal{L}_{pos}$" means the training without positive pairs. The ``Half $\mathcal{L}_{pos}$" means when training to half of the total iteration amount, the learning of positive pairs is added. The ``Full $\mathcal{L}_{pos}$" means the whole training process with positive pairs.} 
\setlength{\tabcolsep}{4.7mm}{
\begin{tabular}{lcccccccccccccccc}
\toprule
\textbf{Type}&\textbf{No $\mathcal{L}_{pos}$}&\textbf{Half $\mathcal{L}_{pos}$}&\textbf{Full $\mathcal{L}_{pos}$}\\
\midrule
DSC$_{\pm std}\uparrow$ &90.3$_{\pm3.6}$ &90.5$_{\pm3.5}$ &\textbf{91.2$_{\pm3.6}$}\\
\bottomrule
\end{tabular}}
\label{tab:pos}
\end{table}
As shown in Tab.\ref{tab:pos}, the learning of our positive pairs $\mathcal{L}_{pos}$ in different settings demonstrates its reliability. Due to the potential misalignment between images at the beginning of the training, the positive pairs will be constructed between misaligned regions, making some potential false positive pairs. Therefore, this experiment performs three situations to evaluate this potential problem. Without the learning of positive pairs (``No $\mathcal{L}_{pos}$"), our GEMINI-Semi has 90.3\% segmentation DSC. When adding the learning of positive pairs at half of the total iteration amount (``Half $\mathcal{L}_{pos}$"), it has 90.5\% DSC which has 0.2\% improvement. When directly adding the positive pairs in the whole process (``Full $\mathcal{L}_{pos}$"), it brings 0.9\% segmentation DSC improvement owing to the constraint for the feature consistency. As conclusion, the potential problem of false positive pairs in our task has less influence on learning, because the alignment accuracy is improved fast as demonstrated in Fig.\ref{Fig:registration}, and the potential interference is reduced in most of the training process. Therefore, we adopted the ``Full $\mathcal{L}_{pos}$" in our framework.
\subsubsection{Analysis of pre-training data amount in SS-MIP}
\begin{figure}
  \centering
  \includegraphics[width=\linewidth]{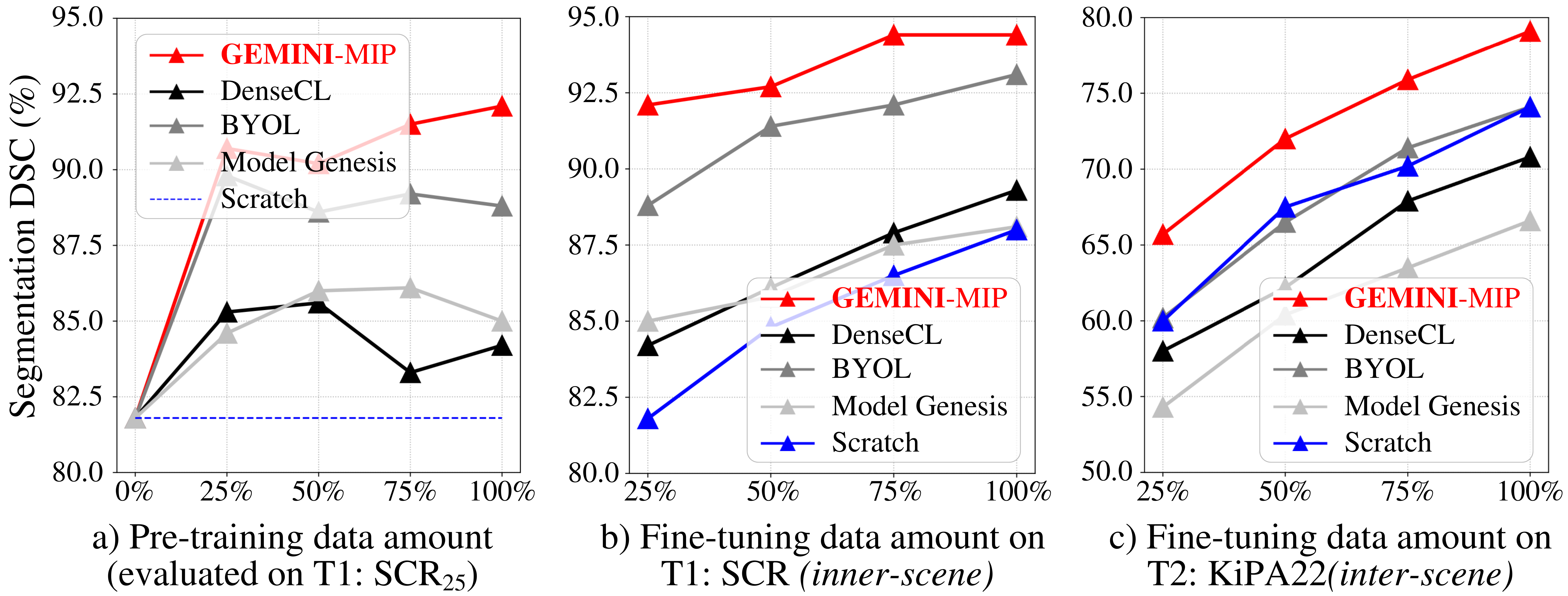}
  \caption{The pre-training and fine-tuning data amount analysis on SS-MIP.}\label{Fig:line}
\end{figure}
As shown in Fig.\ref{Fig:line} a), our GEMINI-MIP has a higher representation efficiency with fewer pre-training data. For the DenseCL, when enlarging the pre-training data amount to 75\%, it even has weaker performance than the 25\%. This is because when the enlarging of pre-training data amount, the contribution of the low-level texture knowledge is reducing and the high-level semantic knowledge is improving. The large-scale FP\&N problem makes the DenseCL learn the high-level semantics which deviate from reality, forbidding the improvement from larger datasets. The improvements of the BYOL and Model Genesis all slow down, owing to the limited representation ability in generation-based learning and the learning without negative pairs. Our GEMINI-MIP has achieved the best performance only with 25\% data which is higher than the BYOL with 100\% data, owing to our reliable learning of positive and negative pairs.

\subsubsection{Analysis of fine-tuning data amount in SS-MIP}
In the \emph{inner-scene} situation (Fig.\ref{Fig:line} b)), our GEMINI-MIP also has a better-transferring ability with fewer data, demonstrating our great data efficiency and cost-saving ability. In the T1: SCR, we evaluate the transferring performance with the enlarging of the downstream data amount in the SCR dataset (25\%, 50\%, 75\%, 100\%). Our GEMINI-MIP achieves the highest performance in all data amount settings. Especially with only 25\% data, it achieves higher performance than the DenseCL, Model Genesis, and Scratch models with 100\% data, and similar performance as the BYOL with 75\% data. Due to the FP\&N problem, although the DenseCL has higher performance than the ``Scratch", its improvement is very slight. Without the FP\&N problem, the BYOL shows a more competitive performance improvement, but it is still lower than ours owing to the lack of negative pairs in the BYOL learning, limiting the discrimination of features.

In the \emph{inter-scene} situation (Fig.\ref{Fig:line} c)), our GEMINI-MIP still has better transferring ability than other methods. We further evaluate the transferring performance with the enlarging of the downstream data amount in the KiPA22 dataset (25\%, 50\%, 75\%, 100\%) for an inter-scene evaluation. Although our GEMINI-MIP is unable to keep its performance when the data amount reduces which is different from it in the inner-scene situation, it still has the highest performance compared with the other four methods in each data amount. This is because our reliable learning of positive and negative feature pairs makes the networks pre-learn low-level patterns effectively which are shared in different tasks, so its representability of low-level features will transformed to different scenes, improving the transferring performance. The DenseCL, BYOL, and Model Genesis models only have similar or lower performance than the ``Scratch", because their pre-learned representations deviate from the reality due to their FP\&N problems, and this property is still transformed to the other scenes making large limitations.

\subsubsection{Analysis of the promotion for learning efficiency}
\begin{figure}
  \centering
  \includegraphics[width=\linewidth]{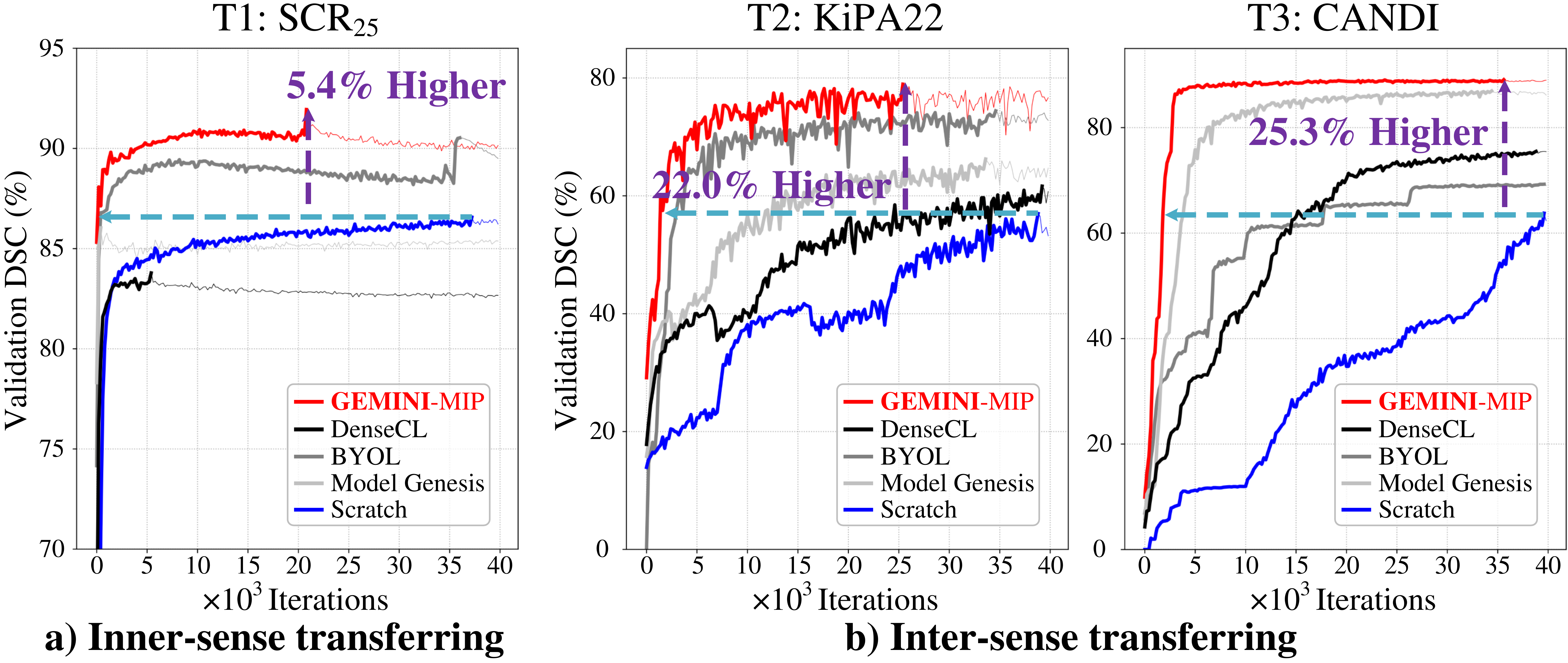}
  \caption{Our GEMINI-MIP has powerful learning efficiency both in the inner-scene and inter-scene transferring tasks.}\label{Fig:learning}
\end{figure}
As shown in Fig.\ref{Fig:learning}, in the transferring of downstream tasks, our GEMINI-MIP has a very powerful learning ability both in inner-scene and inter-scene situations. Compared with ``Scratch", our method has better performance both in accuracy (validation DSC) and learning speed. It only utilizes less than $5\times10^{3}$ iterations and can achieve the performance that exceeds ``Scratch"'s best model who has to train more than $20\times10^3$ iterations. Compared to the best models, our model also has more than 2.4\% DSC higher than the ``Scratch". This is because our model provides a reliable pre-trained initialization when learning the downstream tasks, promoting optimization efficiency. In the inner-scene transferring (T1: SCR$_{25}$), the DenseCL has a worse learning ability than the ``Scratch", because the large interference of the large scale FP\&N problem which makes the initial representation deviate from reality. All four pre-trained models (GEMINI-MIP, DenseCL, BYOL, Model Genesis) have better learning ability than the ``Scratch" in T3: CANDI (numerous small brain regions), because the pre-trained initialization will promote the perception for small structures which are challenging to learn by 2D network.
\subsubsection{Analysis of self-restoration in GEMINI-MIP}
\label{subsubsec:analysisiofrestor}
\begin{figure}
  \centering
  \includegraphics[width=0.8\linewidth]{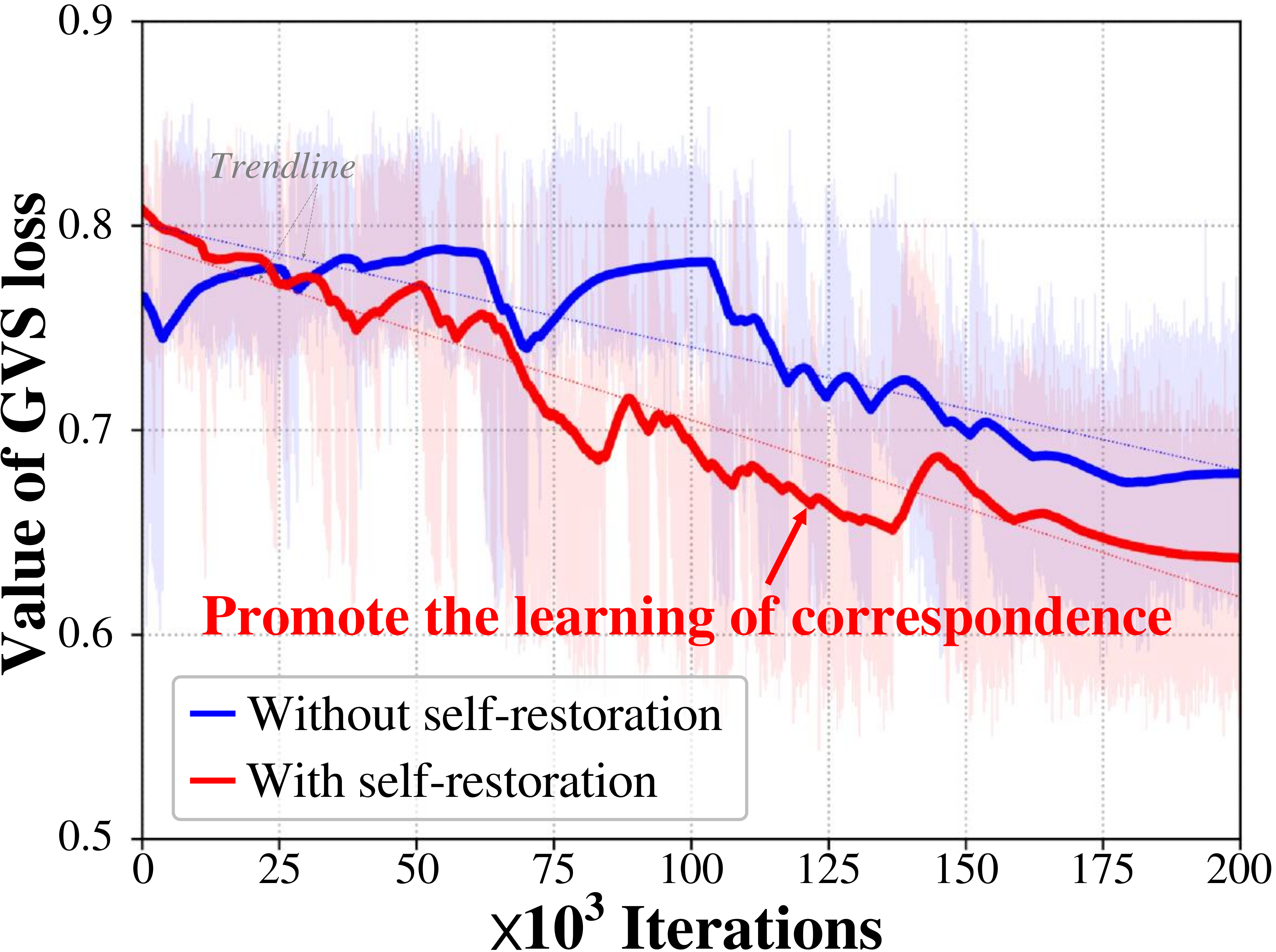}
  \caption{The necessity of the fundament in our GEMINI-MIP. When learning without the fundamental learning task, the GVS loss converges slowly due to the initial weak representation limiting the GSS and GVS for correspondence. When adding the fundament (self-restoration), warmup from the basic representation of semantic regions drives correspondence learning efficiently.}\label{Fig:opt}
\end{figure}
The self-restoration learns a basic representation for the features, thus providing an efficient similarity measurement in our GSS and driving more efficient correspondence learning. As shown in Fig.\ref{Fig:opt}, we evaluate the GVS loss in the training process to demonstrate the improvement of correspondence degree. When only learning our GEMINI-MIP without the fundamental task, the network’s initial weak representation makes inefficient learning of the correspondence, and until the late stage of the training, the GVS began to decline. When adding the fundamental pretext task, driven by the basic representation of semantic regions from the self-restoration, the GVS loss is converging continuously to learn the correspondence of semantic regions.

\subsection{Observations and discussions}
\subsubsection{FS-Semi v.s. SS-MIP}
With our homeomorphism prior, our GEMINI has powerful representation learning ability with little (FS-Semi) or no (SS-MIP) supervision. In our experiment (Sec.\ref{sec:task2}) of FS-Semi, the unlabeled images and the very few labeled images are learned together, so that the networks in the DCRL methods have a supervised optimization target which makes a basic distinction for the features of different segmentation regions. Therefore, this will improve the reliability of the correspondence discovery in the DCRL methods, and achieves more than 70\% DSC on all tasks in FS-Semi. But the reliability in the existing DCRL methods is still weak without our homeomorphism prior, and they all only have lower performance than our GEMINI-Semi. In our experiment (Sec.\ref{sec:task1}) of SS-MIP, the unlabeled images and the labeled images are trained separately in the pretext tasks and downstream tasks, so the networks in the DCRL methods have no supervision during their learning unlabeled images. Their correspondence discovery is extremely limited owing to the properties of medical images, constructing a lot of FP\&N pairs, and learning very poor representation. Therefore, in the downstream tasks, some of them even have worse performance than the ``Scratch". Based on our homeomorphism prior, our GEMINI-MIP has much more reliability to discover the correspondence even without supervision, achieving reliable positive pairs and implicit negative pairs for pre-training and powerful performance for transferring.
\subsubsection{GSS v.s. GVS}
\begin{table}
\centering
\caption{The comparison between our GEMINI-Semi with only GSS and with only GVS on the ``T1: 3D Cardiac structures" demonstrates the advantages of our geometric semantic similarity.}

\setlength{\tabcolsep}{2.5mm}{
\begin{tabular}{lcccccccccccccccc}
\toprule
\multirow{2}{*}{\textbf{Type}}&\textbf{Segmentation}&\multicolumn{2}{c}{\textbf{Deformation}}\\\cmidrule(r){2-2}\cmidrule(r){3-4}
&DSC$_{\pm std}\uparrow$&DSC$_{\pm std}\uparrow$ &$|J_{\phi}|\leq0$$\%\downarrow$  \\
\midrule
\cellcolor[gray]{0.9}Only $\mathcal{L}_{Seg}$&\cellcolor[gray]{0.9}84.3$_{\pm9.6}$&\cellcolor[gray]{0.9}-&\cellcolor[gray]{0.9}-\\
\cellcolor[gray]{0.9}No deformation&\cellcolor[gray]{0.9}-&\cellcolor[gray]{0.9}62.1$_{\pm8.7}$&\cellcolor[gray]{0.9}-\\
\cdashline{1-4}[0.8pt/2pt]
GVS only&89.3$_{\pm4.1}$&51.3$_{\pm10.3}$&40.8$_{\pm1.2}$\\
GSS only&\textbf{90.0$_{\pm3.4}$}&\textbf{84.1$_{\pm13.0}$}&\textbf{10.2$_{\pm2.2}$}\\
\bottomrule
\end{tabular}}

\label{tab:similarity}
\end{table}
The GSS has a very effective measurement ability to guide the learning of deformation (correspondence). As shown in Tab.\ref{tab:similarity}, we perform the learning of correspondence with the GVS or with the GSS, and the losses for the continuity and bijection are removed to avoid their potential influence. The ``GVS only" has 89.3\% DSC on the segmentation which is 5.0\% DSC higher than the ``Only $\mathcal{L}_{Seg}$", demonstrating the improvement from the correspondence learning. However, it only has 51.3\% DSC on the deformation which is 10.8\% lower than the images without deformation. This is because the appearance variation makes the GVS measurement unreliable, so it will train the network to learn the correspondence for those similar-but-different semantic regions (although it is still better than the direct correspondence discovery like \cite{li2021dense,o2020unsupervised,wang2022densecl,wang2022exploring}), finally the deformation will have very large folds (40.8\% if it has no smoothness loss) limiting its performance. The ``GSS only" achieves a significant 32.8\% DSC improvement and 30.6\% $|J_{\phi}|\leq0$ reduction on the deformation compared with the ``GVS only" owing to its measurement of the features which are robustness for the appearance variation. Therefore, better correspondence learning further promotes segmentation learning, achieving 90.0\% DSC which is 0.7\% higher.

\subsubsection{Non-homeomorphic medical images}\label{sec:non}
Our basic hypothesis, homeomorphism prior, limits our GEMINI learning only to be trained between the medical images meeting homeomorphism. However, in the real world, there are a large number of medical images that are unsatisfied with this prior, such as the images with lesions, whose features are unable to be paired only via deformation, hindering the learning of GEMINI. Fortunately, the homeomorphic medical images are easy to collect in real world. The physical examination or disease screening will accumulate a large number of medical images without lesions every year \cite{luo2016big}, providing a potential to collect a big dataset meeting the hypothesis of our framework. This provides data support for our model and hypothesis to have sufficient clinical significance and scope of application.

\subsubsection{Advancements of our preliminary work}\label{sec:adv}
Our preliminary efforts \cite{He_2023_CVPR} first presented the GVSL in 3D medical image SSP, learning the inter-image similarity for powerful representation and efficiently transferring to downstream tasks. This paper extends it (GVSL \cite{He_2023_CVPR}) substantially on self-supervised representation learning with the advancements on principle, method, and application.
\begin{enumerate}[leftmargin=*]
  \item We have proposed the GEMINI learning which is a novel paradigm for the large-scale FP\&N problem in the DCRL with detailed motivation in Sec.\ref{sec:intro}, and proposed a new principle concept, the homeomorphism prior, behind the GVSL in Sec.\ref{sec:hp}.
  \item We have conducted a more comprehensive review of the technological and theoretical research related to our task and provided a clear overview of the field in Sec.\ref{sec:related}.
  \item We have proposed a novel similarity measurement strategy, the GSS, enabling the learned representation in turn to promote the correspondence discovery during the learning process, and promoting learning efficiency for correspondence discovery in Sec.\ref{sec:methodology}.
  \item We have enlarged the application boundary of GVSL advancing the original model that only runs on 3D medical images to any dimension of medical images that satisfies homeomorphism prior.
  \item We have extended our method to more kinds of representation learning tasks, advancing the original model that was only used in pre-training to the variants with both the pre-training and few-shot semi-supervised learning in Sec.\ref{sec:task2} and Sec.\ref{sec:task1}.
  \item We have carried out more experiments for performance analysis and comparison, thus more completely demonstrating the power of our GEMINI learning in Sec.\ref{sec:discussion}.
\end{enumerate}

\subsubsection{Future works}
The future works of the proposed GEMINI and the GVSL are in three aspects:
\begin{enumerate}[leftmargin=*]
  \item As discussed in Sec.\ref{sec:non}, one of our important future works is to expand the learning of correspondence to some images without homeomorphic topology, like the images with lesions \cite{he2021meta}, to cope with the large-scale FP\&N problem in more images types.
  \item Further explore the homeomorphism mapping between images and non-images, like the medical images and the deformed grids for super-pixel segmentation \cite{yang2020superpixel}.
  \item Extend the pre-training to the datasets with multiple image categories, and evaluate the potential of the GVSL as foundation models \cite{liu2024imaging,10750441} for wider scenes.
  \item Design a lighter pre-training process to reduce the computing costs (Sec.C.5 of \emph{Appendix}).
\end{enumerate}

\section{Conclusion}
\label{sec:conclusion}
In this paper, we have advanced the homeomorphism prior in the open problem of large-scale FP\&N pairs in the medical image DCRL, and proposed the \textbf{GE}o\textbf{M}etric v\textbf{I}sual de\textbf{N}se s\textbf{I}milarity (\textbf{GEMINI}) Learning for a reliable dense correspondence discovery and learning. Based on our GEMINI, dense contrastive representation for medical images is learned, effectively reducing the data and annotation costs in medical image dense prediction tasks. Its unique properties of learning implicit negative pairs in our DHL and positive pairs in our GSS have bright powerful performance in few-shot semi-supervised medical image segmentation tasks and self-supervised medical image pre-training tasks. We believe that our GEMINI in DCRL will promote the research of efficient learning in medical image analysis, and coping with the large challenge in data collection and dense annotation. For intuition progress, the objects with the homeomorphic property are all able to construct a reliable point-to-point correspondence discovery via a homeomorphism mapping. This effectively promotes a new representation learning paradigm based on topological consistency, and will inspire future researchers for more powerful innovations.


\appendix

\begin{table*}[tb]
\centering
\caption{The fine-tuning evaluations demonstrate our great transferring ability on SS-MIP tasks which pre-trained on PPMI dataset. Our GEMINI-MIP achieves the best performance compared with 18 methods on two downstream tasks.}
\begin{tabular}{clcccccccccc}
\toprule
\multirow{2}{*}{\textbf{Type}}
&\multirow{2}{*}{\textbf{Pre-training}}
&\multicolumn{2}{c}{\textbf{T2: KiPA22} \emph{Inter-scene}}
&\multicolumn{2}{c}{\textbf{T3: CANDI} \emph{Inner-scene}}
&\textbf{AVG}
\\
\cmidrule(r){3-4}
\cmidrule(r){5-6}
\cmidrule(r){7-7}
&
&DSC$_{\pm std}\uparrow$
&AVD$_{\pm std}\downarrow$
&DSC$_{\pm std}\uparrow$
&AVD$_{\pm std}\downarrow$
&DSC$_{\pm std}\uparrow$
\\
\midrule
-
&Scratch (3D U-Net)
&72.4$_{\pm16.3}$
&6.11$_{\pm5.91}$
&84.0$_{\pm3.2}$
&0.52$_{\pm0.14}$
&78.2$_{\pm9.8}$
\\
\cdashline{1-7}[0.8pt/2pt]
Sup
&Med3D \cite{chen2019med3d}
&81.7$_{\pm12.0}$
&\color{blue}2.61$_{\pm2.77}$
&72.7$_{\pm19.0}$
&1.57$_{\pm2.56}$
&77.2$_{\pm15.5}$
\\
GRL
&Denosing \cite{vincent2010stacked}
&70.0$_{\pm15.4}$
&7.60$_{\pm5.03}$
&\cellcolor[gray]{0.9}83.7$_{\pm3.3}$
&\cellcolor[gray]{0.9}1.71$_{\pm0.20}$
&76.9$_{\pm9.4}$
\\
&In-painting \cite{pathak2016context}
&69.7$_{\pm17.1}$
&7.57$_{\pm5.93}$
&\cellcolor[gray]{0.9}88.5$_{\pm3.1}$
&\cellcolor[gray]{0.9}0.32$_{\pm0.11}$
&79.1$_{\pm10.1}$
\\
&Models Genesis \cite{zhou2019models}
&75.8$_{\pm13.7}$
&4.64$_{\pm4.49}$
&\cellcolor[gray]{0.9}88.7$_{\pm3.1}$
&\cellcolor[gray]{0.9}0.31$_{\pm0.10}$
&82.3$_{\pm8.4}$
\\
&Rotation \cite{komodakis2018unsupervised}
&77.4$_{\pm14.3}$
&4.82$_{\pm6.29}$
&\cellcolor[gray]{0.9}89.4$_{\pm2.6}$
&\cellcolor[gray]{0.9}0.28$_{\pm0.08}$
&83.4$_{\pm8.5}$
\\
CRL
&SimSiam \cite{Chen2021CVPR}
&83.8$_{\pm11.9}$
&3.69$_{\pm7.47}$
&\cellcolor[gray]{0.9}87.3$_{\pm3.1}$
&\cellcolor[gray]{0.9}0.36$_{\pm0.10}$
&85.6$_{\pm7.5}$
\\
&BYOL \cite{grill2020bootstrap}
&83.6$_{\pm11.2}$
&2.78$_{\pm5.42}$
&\cellcolor[gray]{0.9}89.7$_{\pm2.4}$
&\cellcolor[gray]{0.9}\color{blue}0.27$_{\pm0.08}$
&\color{blue}86.7$_{\pm6.8}$
\\
&SimCLR \cite{chen2020simple}
&78.9$_{\pm13.9}$
&4.49$_{\pm5.15}$
&\cellcolor[gray]{0.9}89.2$_{\pm3.0}$
&\cellcolor[gray]{0.9}0.30$_{\pm0.14}$
&84.1$_{\pm8.5}$
\\
&MoCov2 \cite{chen2020improved}
&78.0$_{\pm15.3}$
&4.42$_{\pm5.67}$
&\cellcolor[gray]{0.9}89.7$_{\pm2.4}$
&\cellcolor[gray]{0.9}0.28$_{\pm0.11}$
&83.9$_{\pm8.9}$
\\
&DeepCluster \cite{caron2018deep}
&79.7$_{\pm13.7}$
&4.28$_{\pm5.76}$
&\cellcolor[gray]{0.9}89.8$_{\pm2.4}$
&\cellcolor[gray]{0.9}\color{blue}0.27$_{\pm0.08}$
&84.8$_{\pm8.1}$
\\
DCRL
&VADeR \cite{o2020unsupervised}
&72.1$_{\pm13.8}$
&6.56$_{\pm5.89}$
&\cellcolor[gray]{0.9}87.4$_{\pm3.6}$
&\cellcolor[gray]{0.9}0.35$_{\pm0.11}$
&79.8$_{\pm8.7}$
\\
&DenseCL \cite{wang2022densecl}
&74.0$_{\pm15.8}$
&6.42$_{\pm8.21}$
&\cellcolor[gray]{0.9}87.7$_{\pm3.8}$
&\cellcolor[gray]{0.9}0.34$_{\pm0.13}$
&80.9$_{\pm9.8}$
\\
&SetSim \cite{wang2022exploring}
&73.5$_{\pm15.9}$
&6.34$_{\pm6.68}$
&\cellcolor[gray]{0.9}88.4$_{\pm3.1}$
&\cellcolor[gray]{0.9}0.32$_{\pm0.10}$
&81.0$_{\pm9.5}$
\\
&DSC-PM \cite{li2021dense}
&79.0$_{\pm14.6}$
&4.90$_{\pm6.05}$
&\cellcolor[gray]{0.9}88.5$_{\pm3.4}$
&\cellcolor[gray]{0.9}0.32$_{\pm0.13}$
&83.8$_{\pm9.0}$
\\
&PixPro \cite{xie2021propagate}
&80.0$_{\pm14.4}$
&4.60$_{\pm6.25}$
&\cellcolor[gray]{0.9}\color{blue}89.9$_{\pm2.4}$
&\cellcolor[gray]{0.9}\color{blue}0.27$_{\pm0.07}$
&85.0$_{\pm8.4}$
\\
&GLCL \cite{chaitanya2020contrastive}
&70.7$_{\pm16.9}$
&7.33$_{\pm7.05}$
&\cellcolor[gray]{0.9}87.4$_{\pm3.2}$
&\cellcolor[gray]{0.9}0.34$_{\pm0.09}$
&79.1$_{\pm10.1}$
\\
\cdashline{1-7}[0.8pt/2pt]
\textbf{DCRL}
&\textbf{GVSL-MIP (CVPR)}\cite{He_2023_CVPR}
&\color{blue}84.3$_{\pm10.3}$
&2.85$_{\pm5.12}$
&\cellcolor[gray]{0.9}89.1$_{\pm2.8}$
&\cellcolor[gray]{0.9}0.31$_{\pm0.11}$
&\color{blue}86.7$_{\pm6.6}$
\\
\textbf{(Ours)}
&\textbf{GEMINI-MIP}
&\color{red}\textbf{85.0$_{\pm10.2}$}
&\color{red}\textbf{2.55$_{\pm5.71}$}
&\cellcolor[gray]{0.9}\color{red}\textbf{90.0$_{\pm2.4}$}
&\cellcolor[gray]{0.9}\color{red}\textbf{0.26$_{\pm0.07}$}
&\color{red}\textbf{87.5$_{\pm6.3}$}
\\
\bottomrule
\end{tabular}
\label{supp:tab:metrics}
\end{table*}
\section*{A SS-MIP on more datasets}
\label{aupp:sec:task1}
\subsection*{A.1 Self-supervised pre-training on PPMI dataset}
We further evaluate the SS-MIP task on another pretext dataset for pre-training to demonstrate our representation ability. We extracted 837 3D brain T1 MR images with Parkinson’s disease from the PPMI database\footnote{PPMI database: \url{https://www.ppmi-info.org/}} as our pretext dataset. In our experiment, we extract the brain regions via HD-BET \cite{isensee2019automated}, crop and resize the images to $160\times160\times128$, and finally normalize them via the zero-score. Due to the consistency of the human brain regions, we randomly pair these brain images to pre-train the frameworks. Following the Experiment 2 (Sec.5) in our manuscript, we take the Task 2: KiPA22 dataset and Task 3: CANDA as the downstream tasks to evaluate the inter-scene and inner-scene transferring abilities. (Because the Task 1: SCR$_{25}$ dataset is 2D and the pre-trained models are 3D, we exclude this task in this experiment.) We utilize the same implementation and evaluation metrics as the Sec.5 in this experiment.

As shown in Tab.\ref{supp:tab:metrics}, it achieves similar observations as the SS-MIP experiment in Sec.5. For most of the methods, the pre-training on the PPMI dataset will bring better performance than random initialization (“Scratch”) both in the T2: KiPA22 and T3: CANDI tasks. Especially in the T3 (inner-scene), most of the pre-training methods achieve more than 4.0\% DSC improvement compared with the “Scratch”. Even though the other CRL and DCRL methods have FP\&N problems in this experiment, they are still able to learn the representation of some domain features and promote their final performance to the upper limit of the task 3 (near 90\%). When transferring the pre-trained models to the T2 (inter-scene), the SimSiam, BYOL, our GVSL-MIP, and our GEMINI-MIP all still have significant improvement (more than 10\% DSC). This is because these methods learn the consistency of features and avoid the FP\&N problems. The other methods’ performance improvement is obviously decreased owing to the FP or FN problem which interrupts their representation learning of high-level semantics and makes their representations deviate from reality. On both two tasks, our GEMINI-MIP achieves the highest performance showing our superiority.
\begin{table}[tb]
\centering
\caption{The gap coefficient $G^{i}$ quantifies the gap between “pre-trained on chest X-ray images \& fine-tuning on brain T1 MR images” (inter-scene) and “pre-trained on brain T1 MR images \& fine-tuning on brain T1 MR images” (inner-scene).}
\resizebox{\linewidth}{!}{
\begin{tabular}{ccccc}
\toprule
\textbf{Index}
&\multirow{3}{*}{\textbf{Method}}
&\textbf{Chest X-ray}
&\textbf{Brain T1 MR}
&\textbf{Gap}
\\
\cmidrule(r){1-1}
\cmidrule(r){3-3}
\cmidrule(r){4-4}
\cmidrule(r){5-5}
\multirow{2}{*}{$i$}
&
&2D U-Net
&3D U-Net
&\multirow{2}{*}{$G^{i}$}
\\
&
&\emph{Inter-scene}
&\emph{Inner-scene}
&
\\
\midrule
0
&Scratch
&65.0$_{\pm4.4}$
&84.0$_{\pm3.2}$
&1
\\
\cdashline{1-5}[0.8pt/2pt]
1
&BYOL
&70.5$_{\pm2.1}$
&\cellcolor[gray]{0.9}89.7$_{\pm2.4}$
&1.01
\\
2
&DeepCluster
&60.0$_{\pm2.2}$
&\cellcolor[gray]{0.9}89.8$_{\pm2.4}$
&1.57
\\
3
&Model Genesis
&88.1$_{\pm3.1}$
&\cellcolor[gray]{0.9}88.7$_{\pm3.1}$
&0.03
\\
4
&DenseCL
&76.8$_{\pm2.9}$
&\cellcolor[gray]{0.9}87.7$_{\pm3.8}$
&0.57
\\
\cdashline{1-5}[0.8pt/2pt]
\textbf{5}
&\textbf{Our GEMINI-MIP}
&\textbf{89.8$_{\pm2.6}$}
&\cellcolor[gray]{0.9}\textbf{90.0$_{\pm2.4}$}
&\textbf{0.01}
\\
\bottomrule
\end{tabular}
}
\label{supp:tab:gap}
\end{table}
\subsection*{A.2 Analysis of the gap between the inner-scene and inter-scene transferring}
As shown in Tab.\ref{supp:tab:gap}, the quantitative evaluation of the gap between the inner-scene and inter-scene transferring show our great transferring ability both inner scene and inter scene. Here, we formulate a gap coefficient $G$ to quantify this gap:
\begin{equation}\label{equ:gap}
G^{i}=\frac{S^{i}_{inner}-S^{i}_{inter}}{S^{0}_{inner}-S^{0}_{inter}},
\end{equation}
where the $i$ is the index of the method, $S$ is the score of the method (here we take the DSC). The $S_{inner}^{0}-S_{inter}^{0}$ is the gap of the “Scratch” between the two settings which means the difference caused by the initial situation, such as network structure and dimension. The $S_{inner}^{i}-S_{inter}^{i}$ is the gap of the $i_{th}$ method between the two settings. Therefore, the $\frac{S_{inner}^{i}-S_{inter}^{i}}{S_{inner}^{0}-S_{inter}^{0}}$ means the gap of the model in two settings excluding the gap caused by the initial network. If the $G^{i}$ is larger than 1, it means that the pre-trained model has weaker inter-scene transferring ability than inner-scene transferring. If it is smaller than 1, it means that the model has great inter-scene transferring ability.

Most self-supervised learning methods have large gap between inner- and inter-scene transferring, and our GEMINI has great universal representation for different scenes. The BYOL and DeepCluster are limited in the inter-scene transferring ($G>1$) because they only take the image-level contrast which will represent the high-level semantic features and this representation is very different between scenes. The DenseCL has 0.57 gap coefficient which is better than the BYOL and DenseCL. Because it takes dense contrastive learning which also represent low-level detail features and this representation is shared in different scene. Our GEMINI and the Model Genesis all have good inter-scene transferring ability with very low gap coefficient (0.01 and 0.03), showing their great universal representation ability and demonstrating their potential as an initialization for more scenes.

\section*{B Discussion of the research problem and method}
\subsection*{B.1 Discussion of FP\&N problem}
\begin{figure}[tb]
  \centering
  \includegraphics[width=\linewidth]{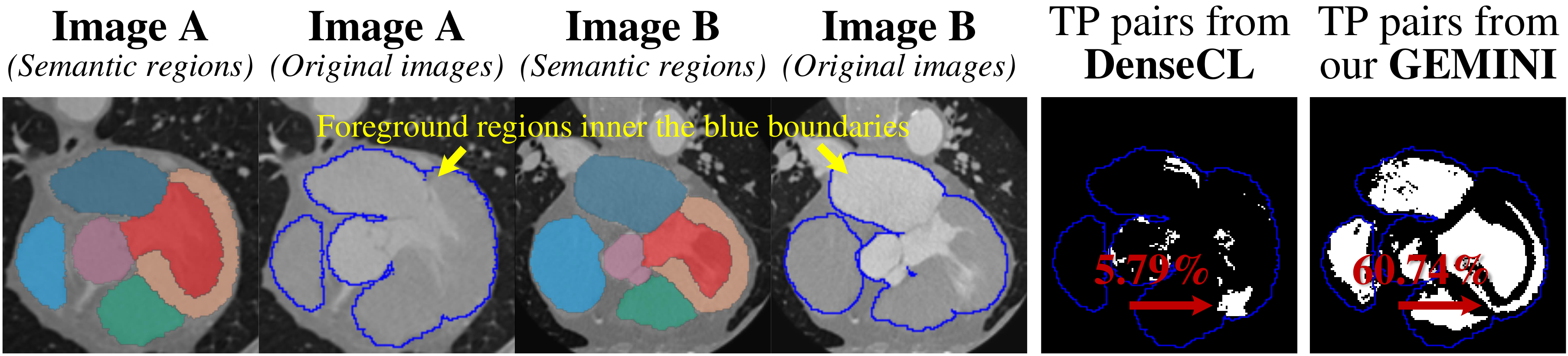}
  \caption{The evaluation of the large-scale FP problem. The true positive (TP) pairs constructed by the features’ similarity (used in DenseCL) only occupy the 5.79\% of the foreground region, and our GEMINI is able to bring 60.74\% TP pairs. }\label{supp:fig:fp}
\end{figure}
As analyzed in the Introduction section, medical images' semantic dependence property will make large-scale FP problem, and their semantic continuity and semantic overlap properties will make large-scale FN problem. In this section, we make an experiment to quantitatively count the percentage of FP and FN pairs in the pairing process.

For FP pairs, we utilize two cardiac CT images (image A and B), and extract their pixel-wise features via a random initialized 3D U-Net. Then, we utilize the pixel-wise feature similarity measurement method in the DenseCL \cite{wang2022densecl} to extract the positive pairs. Because the semantics of the background region are unclear, we count the accuracy of the feature pairs in the foreground regions. As shown in Fig.A, only 5.79\% of the positive pairs in the foreground region are accurate. Therefore, if we directly pair the features only according to their similarity, most of the contrasts (94.21\%) for positive pairs are inaccurate in the medical images and will interrupt the whole contrastive learning process. This is because medical images have very weak contrast due to their special imaging way, making the directly extracted features lack discrimination. Therefore, it makes the “Semantic dependence” one of the inherent properties in medical images constructing large-scale FP pairs.

For FN pairs, we further evaluate the percentage of FN pairs caused by the semantic continuity and semantic overlap properties, and the results show large potential limitations in the DCRL. a) For the PN pairs caused by the “Semantic continuity”, we follow the SimCLR \cite{chen2020simple} which pairs the negative features for each feature. We pair the features in different positions of image A’s foreground regions as negative pairs. The result shows that 17.79\% of the negative pairs are FN pairs which have the same semantics. Although the existing DCRL methods utilize attention \cite{wang2022exploring} or clustering \cite{li2021dense} to avoid directly dividing adjacent pixel-wise features as negative pairs, the FN caused by “semantic continuity” is still an open and challenging problem. b) For the FN pairs caused by the “Semantic overlap”, we follow the DenseCL \cite{wang2022densecl} which pairs the current features and the memory bank features as the negative pairs. We make the features of image B in the foreground as the memory bank features and the features of image A in the foreground as the current features. Then, we pair the current and memory bank features as negative pairs and calculate the accuracy. Finally, 17.53\% of the negative pairs are FN pairs which have the same semantics. The “Semantic overlap” property of the medical images makes it inevitable that there will be numerous consistent semantic regions between medical images. Therefore, it will produce 17.53\% FN pairs in the training process making the model learn in an unreliable direction.

\begin{figure}[tb]
  \centering
  \includegraphics[width=\linewidth]{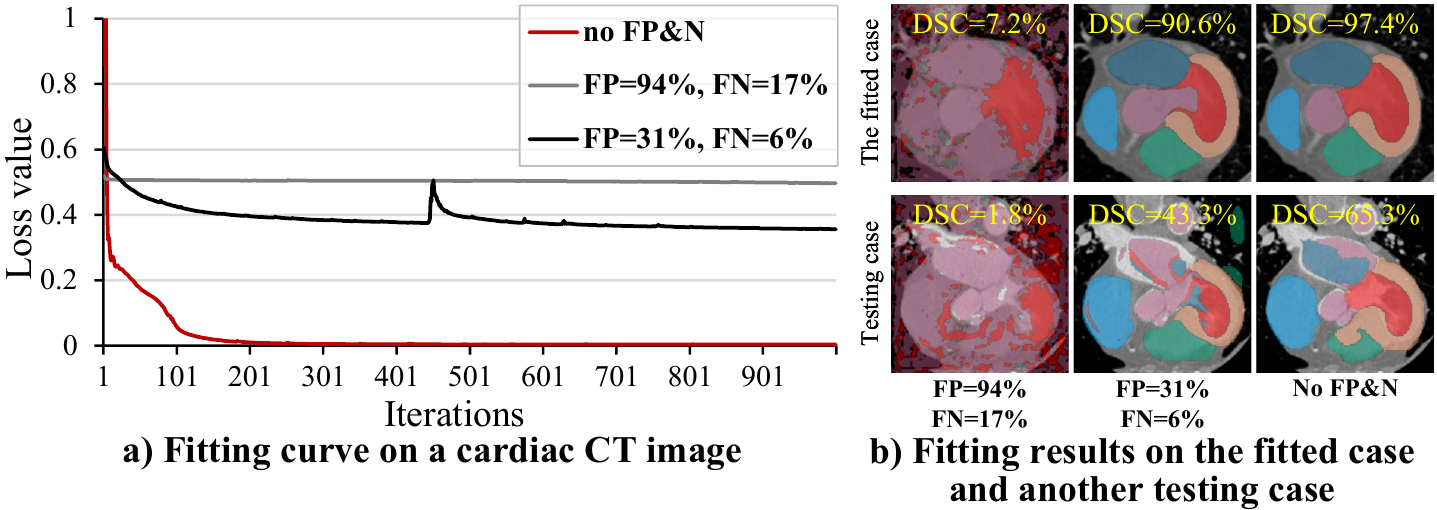}
  \caption{The FP and FN pairs have a serious impact on learning. a) The fitting process with FP and FN pairs on a cardiac CT image. b) The models' learned segmentation ability on the fitted case and their generalization ability on another testing case.}\label{supp:fig:fitting}
\end{figure}

According to the above probability of FP and FN pairs, we simulated the number of these FP and FN pairs in a supervised heart segmentation learning task. Specifically, we train a U-Net on the cardiac structures segmentation task with a cardiac CT image (Image A in Fig.\ref{supp:fig:fp}) to evaluate the fitting ability of the model with or without FP\&N pairs. a) In the non-FP\&N pairs setting, we utilize the contrastive segmentation learning like Wang et al. \cite{wang2019panet}. b) In the FP\&N pairs setting, we randomly generate FP (94\%) and FN (17\%) pairs in the contrastive segmentation learning. c) We further reduce the probabilities of FP and FN pairs to one-third of original (31\% and 6\%) to give an ablation of the false pairs’ degree. We take 1000 iterations, and draw the loss values of the learning process on a line chart to visualize the fitting process. We also evaluate the segmentation of the fitted case and another testing case (Image B in Fig.\ref{supp:fig:fp}) to evaluate the model learned representation with false pairs.

As shown in Fig.\ref{supp:fig:fitting}, the FP and FN pairs have a serious impact on learning. Without the FP\&N pairs, the model is able to be fitted to the target cardiac images, and learn the representation ability of the semantic regions. However, when learning with large-scale FP\&N pairs (94\%, 17\%), the model is unable to be fitted to the targets owing to the interference of the noisy optimization targets. When reducing the FP\&N degree to one-third, the model is able to be gradually fitted to the target image and has a certain generalization, but its performance is weaker than the ``no FP\&N" situation. Therefore, we can draw the following two conclusions in DCRL: a) the large-scale FP\&N problem will make the model unable to learn representation; b) alleviating the FP\&N degree, the model will be able to learn the representation ability of data with generalization ability. Therefore, our GEMINI embeds the homeomorphism prior to the DCRL for the large-scale FP\&N problem, enhancing the learning of true feature pairs. Although it is challenging to remove FP\&N pairs without annotation, reducing the FP\&N degree via our GEMINI is still able to guides the model to learn a generalizable representation.

\subsection*{B.2 Discussion of the novelty in GEMINI}
The proposed GEMINI is a novel dense contrastive representation learning paradigm in medical image analysis. Not only in the innovations, i.e., our DHL and GSS, it also achieved great novelty in principle.

\emph{In principle}, our GEMINI has advanced the theoretical foundation of homeomorphism for the dense contrastive representation learning, providing a principle inspiration to the community. It modeled the human consistent anatomy in medical images based on the principle of topologie \cite{hubbard1991differential}, proposed a new principal concept, homeomorphism prior, and formulated it in the DCRL task as a new paradigm. Therefore, the community will be further inspired by our principle of homeomorphism and make new scientific and technological progress in other tasks and fields.

\emph{In methodology}, our work has proposed a novel dense contrastive representation learning framework that enables the contrast of feature pairs under the condition of human inherent topology, thus promoting the DCRL in medical images. It modeled the consistency of human inherent topology (i.e., homeomorphism prior) as a learning for deformable mapping to overcome the reliability issue in DCRL’s feature correspondence process, giving one potential answer to the long-standing question of “how to achieve a reliable dense feature correspondence for unlabeled data?” Based on the modeling, the proposed DHL and GSS bring soft learning of feature pairs and reliable learning of positive pairs, promoting the contrast of features in DCRL. Finally, our work has achieved a new ability to learn reliable semi-supervised medical image segmentation and pre-training models.

\section*{C More framework analysis and experiment discussion}
\subsection*{C.1 Discussion of the receptive field $r$ in the Deformer network}
\begin{figure}[tb]
  \centering
  \includegraphics[width=\linewidth]{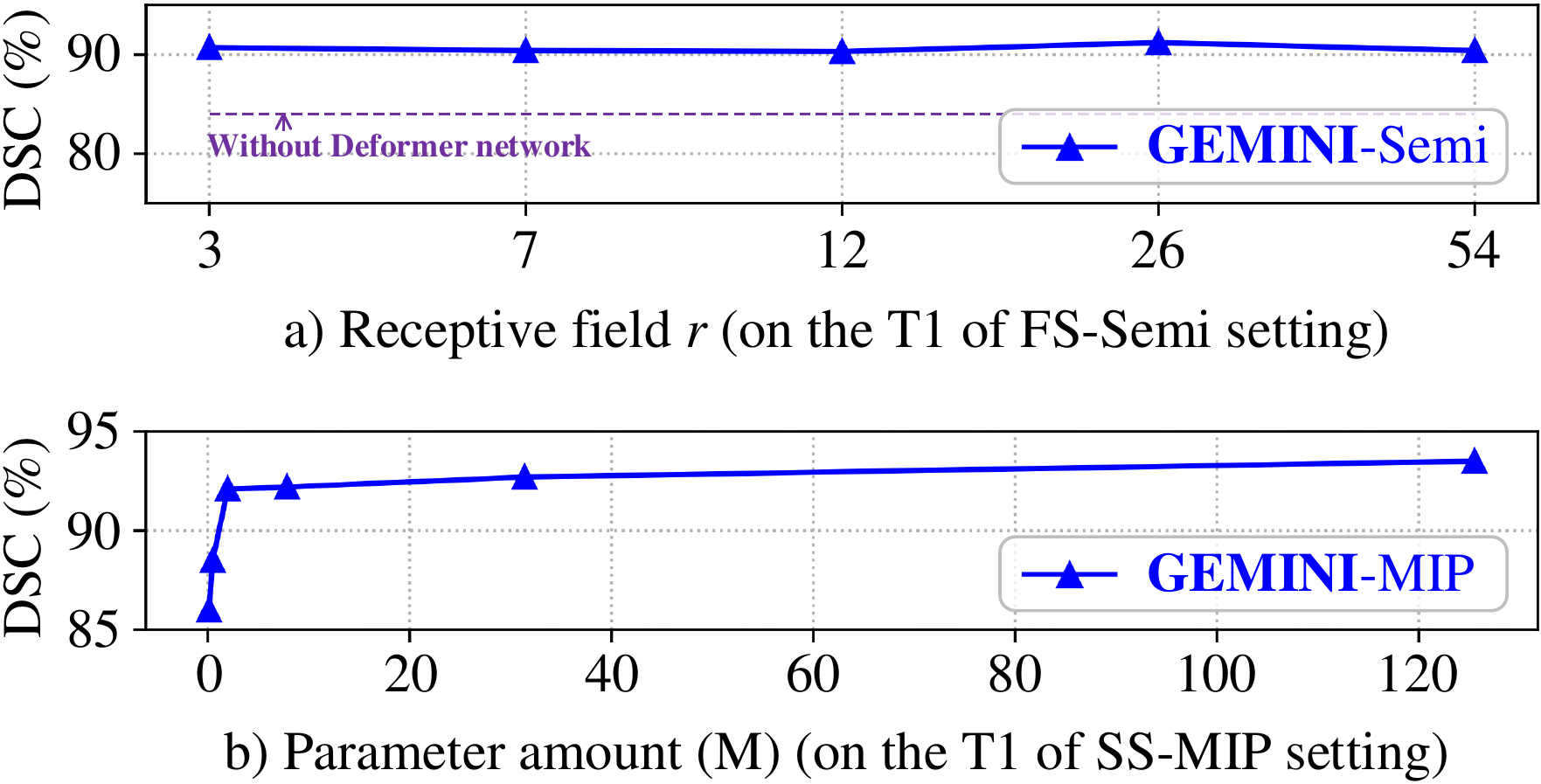}
  \caption{The ablation study of the receptive field size $r$ and the network parameter amounts. a) The segmentation performance on the T1 of FS-Semi setting with the increasing of the receptive field size $r$. b) The fine-tuning performance with the enlarging of the parameter amount (million, $M$) in the pre-trained networks.}\label{supp:fig:rece}
\end{figure}

The performance is robust for the receptive field $r$. As shown in Fig.\ref{supp:fig:rece} a), we enlarge the receptive field $r$ via adding the depth and down-sampling stages of the “Deformer” network and evaluate the model's performance on T1 of FS-Semi setting. With the enlarging of the receptive field, the models’ performance is stalely around 90\% DSC. Because the backbone network and “Deformer” network together constitute a whole network to learn the feature representation, and the features from the backbone network have been extracted from a large receptive field. Therefore, even the receptive field of the “Deformer” network is small, the final DVF is still calculated from a large receptive field. The layers inner the backbone is still optimized by the gradient with a big reception, so that our GEMINI keeps stable performance with the enlarging of $r$. Owing to the soft learning of feature pairs in our DHL, once added this module, the framework achieves more than 5\% DSC improvement.

\subsection*{C.2 Discussion of the parameter amount}
As shown in Fig.\ref{supp:fig:rece} b), we have evaluated our GEMINI on different settings of model parameters. We pre-trained our GEMINI-MIP on the ChestX-ray8 dataset for the networks with 0.12$M$, 0.49$M$, 1.97$M$, 7.85$M$, 31.39$M$, and 125.52$M$ ($M$ is million) parameters, and fine-tuned them on the T1: SCR25 task. With the enlarging of the network, the model performance is improving quickly. This is because the network capacity increases with the enlarging of the networks so that it will be able to learn the representation of more features in the pre-training process. When the parameter amount is 1.97M, the speed of performance improving is reduced, illustrating that the increase of network capacity has approached the upper bound of this task. Therefore, when the network is further enlarged to 125.52$M$ (more than 50 times compared with 1.97$M$), the performance is only improved 1.4\% DSC. 

\subsection*{C.3 Discussion of the feature distribution}
\begin{figure}[tb]
  \centering
  \includegraphics[width=\linewidth]{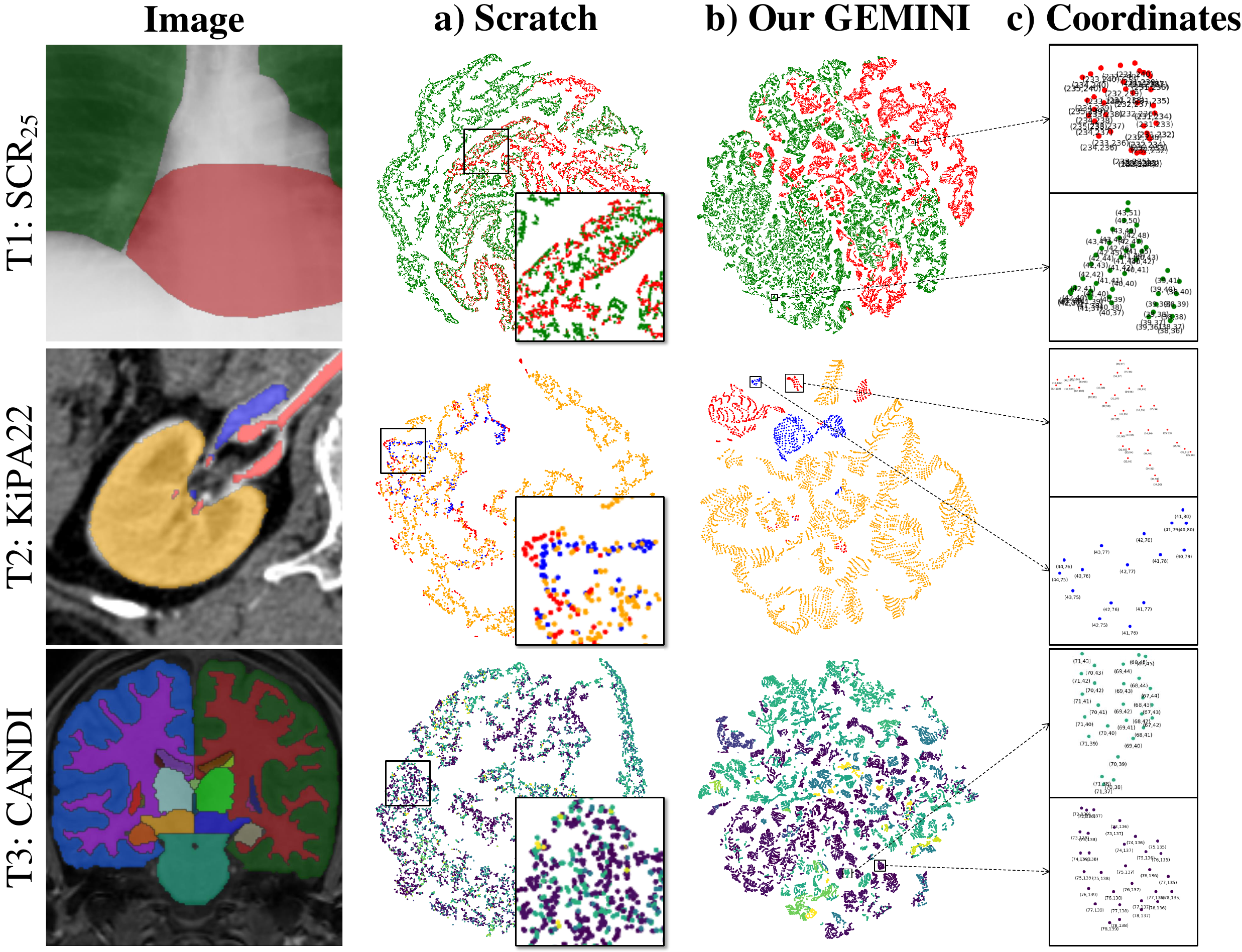}
  \caption{The t-SNE visualization of the learned pixel representations. We provide the coordinates of pixels in a zoomed view, indicating their spatial relationship.}\label{supp:fig:distri}
\end{figure}

As shown in Fig.\ref{supp:fig:distri}, we visualize the learned representation by our framework to demonstrate its effectiveness in distinguishing different semantic regions. In the three tasks of the SS-MIP experiment, we randomly select the slices or patches from the test datasets and extract their pixel-wise features via the backbone network initialized from scratch (a) and our GEMINI-MIP (b). Then, these features are zoomed by t-SNE \cite{van2008visualizing} to two dimensions. As demonstrated in the enlarged region, the features from the ``Scratch" model is mixed owing to its initial weak representation. The pixel-wise features from our framework are clustered into several meaningful groups. Most of the pixels in each group are spatially close and in different groups are also spatially separated (indicated by their coordinates (c)) in the original image. Because our GEMINI discovers the correspondence of pixel-wise features based on the homeomorphism of human body and learns the representation according to the consistent context topology, the same semantic features which are spatially close will be clustered.

\subsection*{C.4 Discussion of the cross-architecture compatibility}
\begin{table*}[tb]
\centering
\caption{The FS-Semi evaluations on U-Net \cite{ronneberger2015u}, TransUNet \cite{chen2021transunet}, and SwinUNet \cite{cao2022swin} demonstrate the cross-architecture compatibility of our GEMINI. The ``-" means that the setting is unable to be implemented.}
{
\begin{tabular}{clccccccccccccccc}
  \toprule
  \multirow{2}{*}{\textbf{Type}}
  &\multirow{2}{*}{\textbf{Method}}
  &\multicolumn{2}{c}{\textbf{T1: 3D cardiac structures}}
  &\multicolumn{2}{c}{\textbf{T2: 3D brain tissues}}
  &\multicolumn{2}{c}{\textbf{T3: 2D chest structures}}
  &\textbf{AVG}\\ \cmidrule(r){3-4}\cmidrule(r){5-6}\cmidrule(r){7-8}\cmidrule(r){9-9}
  &
  &DSC$_{\pm std}\uparrow$
  &AVD$_{\pm std}\downarrow$
  &DSC$_{\pm std}\uparrow$
  &AVD$_{\pm std}\downarrow$
  &DSC$_{\pm std}\uparrow$
  &AVD$_{\pm std}\downarrow$
  &DSC$_{\pm std}\uparrow$
  \\
  \midrule
  \textbf{Five}
  &U-Net \cite{ronneberger2015u}
  &84.3$_{\pm9.6}$
  &2.43$_{\pm2.14}$
  &69.5$_{\pm8.8}$
  &1.59$_{\pm0.84}$
  &83.4$_{\pm6.9}$
  &10.34$_{\pm4.80}$
  &79.1$_{\pm8.4}$
  \\
  \textbf{(Lower)}
  & TransUNet \cite{chen2021transunet}
  & 74.5$_{\pm8.3}$
  & 4.41$_{\pm1.39}$
  & 67.4$_{\pm5.4}$
  & 2.02$_{\pm0.46}$
  & 76.5$_{\pm8.2}$
  & 16.59$_{\pm6.53}$
  & 72.8$_{\pm7.3}$
  \\
  & SwinUNet \cite{cao2022swin}
  & 40.8$_{\pm8.0}$
  & 11.59$_{\pm1.32}$
  & 67.8$_{\pm5.3}$
  & 4.04$_{\pm0.39}$
  & 63.9$_{\pm11.5}$
  & 14.26$_{\pm8.91}$
  & 57.5$_{\pm8.3}$
  \\
  \cdashline{1-9}[0.8pt/2pt]
  \textbf{Full}
  &U-Net \cite{ronneberger2015u}
  &-
  &-
  &88.7$_{\pm1.2}$
  &0.31$_{\pm0.04}$
  &96.1$_{\pm1.4}$
  &2.28$_{\pm1.00}$
  &-
  \\
   \textbf{(Upper)}
  & TransUNet \cite{chen2021transunet}
  & -
  & -
  & 85.7$_{\pm1.2}$
  & 0.43$_{\pm0.05}$
  & 95.2$_{\pm2.1}$
  & 2.78$_{\pm1.35}$
  & -
  \\
  & SwinUNet \cite{cao2022swin}
  & -
  & -
  & 82.8$_{\pm2.7}$
  & 0.54$_{\pm0.15}$
  & 95.3$_{\pm1.2}$
  & 2.17$_{\pm0.65}$
  & -
  \\
  \cdashline{1-9}[0.8pt/2pt]
  \textbf{Semi}
  &\textbf{GEMINI+U-Net}
  &91.2$_{\pm3.6}$
  &0.97$_{\pm0.56}$
  &87.3$_{\pm1.0}$
  &0.35$_{\pm0.03}$
  &87.7$_{\pm5.2}$
  &7.14$_{\pm3.63}$
  &88.7$_{\pm3.3}$
  \\
   \textbf{(Ours)}
  &\textbf{GEMINI+TransUNet}
  &90.8$_{\pm3.4}$
  &0.94$_{\pm0.51}$
  &84.4$_{\pm1.3}$
  &0.45$_{\pm0.05}$
  &88.4$_{\pm5.7}$
  &8.63$_{\pm4.68}$
  &87.9$_{\pm3.5}$
  \\
  &\textbf{GEMINI+SwinUNet}
  &88.6$_{\pm4.2}$
  &1.28$_{\pm0.64}$
  &79.9$_{\pm5.0}$
  &0.62$_{\pm0.20}$
  &86.2$_{\pm7.8}$
  &6.34$_{\pm4.34}$
  &84.9$_{\pm5.7}$
  \\
  \bottomrule
\end{tabular}
}
\label{tab:arch}
\end{table*}
As shown in the Tab.\ref{tab:arch}, we perform the TransUNet \cite{chen2021transunet} (CNN+transformer), SwinUNet \cite{cao2022swin} (Transformer), and U-Net \cite{ronneberger2015u} (CNN) on the FS-Semi tasks with three datasets, and our GEMINI demonstrates a great compatibility across these model architectures. There are two observations: \textbf{a)} Our GEMINI has achieved significant improvement on both TransUNet \cite{chen2021transunet}, SwinUNet \cite{cao2022swin}, and U-Net \cite{ronneberger2015u}. Compared with the lower bound of the architectures that are trained only with five labeled images, our GEMINI has improved them more than 9\% DSC on AVG owing to our learning of the  homeomorphism mapping between medical images. Especially, for the 3D brain tissues segmentation, GEMINI achieves a similar performance compared with the FULL setting (83 labels) only with 5 labels in all architectures, demonstrating our great potential in reducing of annotation costs. \textbf{b)} Our GEMINI has great architecture compatibility across CNN-based (U-Net [3]), transformer-based (SwinUNet [2]), and CNN-transformer-based (TransUNet) networks. For U-Net and TransUNet that utilizes CNN to encode and decode features, our GEMINI has similar significant improvement that achieves 88.7\% and 87.9\% on AVG DSC. SwinUNet takes patch-embedding and four-times down sampling at the beginning, and utilizes the shifted window to learn global features. Therefore, it is challenging to represent fine-grained dense features and makes the whole network easy to overfit to the global features when the amount of training cases is small. As a result, SwinUNet has very poor performance on “FIVE” setting. When adding GEMINI, it learns the inter-image consistency for unlabeled images and effectively reduces the over-fitting, thus also achieving more than 20\% DSC improvement.

\subsection*{C.5 Discussion of the computing costs}
\begin{table}[tb]
\centering
\caption{Owing to the additional deformer networks, our GEMINI has relatively higher computing costs in the pre-training stage, but it has same computing costs in the fine-tuning for downstream tasks as other methods and achieves much higher performance.}
\resizebox{\linewidth}{!}
{
\begin{tabular}{clccccccccccccccc}
  \toprule
  \multirow{2}{*}{\textbf{Type}}
  &\multirow{2}{*}{\textbf{Method}}
  &\textbf{Pre-training}
  &\textbf{Downstream}
  &\textbf{T1: SCR$_{25}$}
  \\
  \cmidrule(r){3-3}\cmidrule(r){4-4}\cmidrule(r){5-5}
  &
  & FLOPs
  & FLOPs
  & DSC$_{\pm std}$
  \\
  \midrule
  1$\times$Encoder
  &Rotation \cite{komodakis2018unsupervised}
  &5.99G
  &20.15G
  &80.5$_{\pm7.7}$
  \\
  2$\times$Encoder
  &BYOL \cite{grill2020bootstrap}
  &11.98G
  &20.15G
  &89.4$_{\pm4.9}$
  \\
  1$\times$Encoder-decoder
  &Model Genesis \cite{zhou2019models}
  &19.74G
  &20.15G
  &86.1$_{\pm4.6}$
  \\
  2$\times$Encoder-decoder
  &VADeR \cite{o2020unsupervised}
  &39.67G
  &20.15G
  &85.2$_{\pm5.1}$
  \\
  \cdashline{1-5}[0.8pt/2pt]
  2$\times$Encoder-decoder
  &\textbf{Our GEMINI}
  &\textbf{52.59G}
  &20.15G
  &\textbf{92.1$_{\pm2.8}$}
  \\
  \bottomrule
\end{tabular}
}
\label{tab:computing}
\end{table}
As shown in Tab.\ref{tab:computing}, we compare the methods’ number of floating-point operations (FLOPs) in four architecture types in pre-training stage, i.e., ``1$\times$Encoder" that only runs an encoder in the pre-training, here, we take the Rotation method \cite{komodakis2018unsupervised}; ``2$\times$Encoder" that runs two encoders for contrastive representation learning in the pre-training, here we take the BYOL \cite{grill2020bootstrap}; ``1$\times$Encoder-decoder" that runs an encoder-decoder network, like the U-Net, in the pre-training, here we take the Model Genesis \cite{zhou2019models}; ``2$\times$Encoder-decoder" that runs two encoder-decoder networks for dense contrastive representation learning in the pre-training, here we take the VADeR \cite{o2020unsupervised}. Our GEMINI is also a ``2$\times$Encoder-decoder" method. In the pre-training stage, all methods take U-Net (for Encoder-decoder) or the encoder part of the U-Net (for Encoder) as their backbone. In the downstream adaptation stage, all methods’ pre-trained parameters are used to initialize the U-Net to learn segmentation task (T1: SCR$_{25}$) and the part without pre-training is initialized randomly. All methods utilize same input sizes with [300$\times$300] in pre-training stage and [512$\times$512] in downstream stage. Owing to two additional deformer networks to learning the homeomorphism mapping, our GEMINI has the highest FLOPs in the pre-training stage, but it greatly contributes to the pre-training performance. In the downstream stages, owing to all methods take the U-Net with same parameter amount, our GEMINI has same FLOPs as other methods. As a results, our GEMINI has very significant performance improvement on the SCR$_{25}$ task owing our reliable learning for positive and negative feature pairs. The large-scale FP\&N problem in VADeR makes it has worse performance than the BYOL even it has larger FLOPs in pre-training.

The additional the one-time cost of our GEMINI in the pre-training stage bring obtain better representation, effectively reducing the long-time computing costs in the downstream tasks. Because once the pre-training is completed, stronger representation will accelerate the convergence speed of the model on downstream tasks, thus reducing the long-time computing cost in the training of numerous downstream tasks. As illustrated in the Fig.12 of our paper, compared with the BYOL \cite{grill2020bootstrap}, DenseCL \cite{wang2022densecl}, Model Genesis \cite{zhou2019models}, our GEMINI achieved better performance with fewer iterations, illustrating its potential in reducing the computing costs in downstream tasks.

\subsection*{C.6 Discussion of the second-best models}
Compared with the second-best methods (BRBS \cite{he2022learning} in Tab.2 and our CVPR version, GVSL \cite{He_2023_CVPR}, in Tab.3), we can find these methods also fused the homeomorphism prior into their framework, and their great performance demonstrates the great potential of this prior knowledge in medical images. In our Experiment 1: FS-Semi (Sec.4), the BRBS is a “learning registration to learn segmentation” method whose registration part is based on the homeomorphism prior. Therefore, its powerful performance in the T1: 3D cardiac structures and T2: 3D brain tissues illustrate the advantages. However, the BRBS’s visual similarity make it unable to generalize to the chest X-ray (T3: 2D chest structures) that has relatively low contrast as illustrated in Fig.8. Our GEMINI utilize the semantic similarity based on features and achieves significant improvement on this task, demonstrating our superiority. In our Experiment 2: SS-MIP (Sec.4), our CVPR version, GVSL, benefits from our homeomorphism prior, achieving second-best performance on the T2: KiPA22 and T3: CANDI. However, it also utilizes the visual similarity which is limited on the low-contrast images, i.e., the chest x-ray images in the pre-training dataset. Therefore, its pre-trained representation for chest x-ray is relatively weaker and limits its performance in the inner-scene transferring. Our GSS improves the measurement of the correspondence degree, and drive the representation learning for low-contrast targets during pre-training. Therefore, our GEMINI has significantly improved the GVSL’s performance on the T1: SCR$_{25}$ task.

\subsection*{C.7 Discussion of the reliability}
\begin{table*}[tb]
\centering
\caption{The evaluation of our GEMINI's reliability on the tasks in Experiment 1. The \emph{Cor} is the Pearson correlation coefficient \cite{cohen2009pearson}, and the \emph{p} is the p-value.}
{
\begin{tabular}{lccccccccccccccc}
  \toprule
  \diagbox{Evaluations}{Tasks}
  &\multicolumn{2}{c}{\textbf{T1: 3D cardiac structures}}
  &\multicolumn{2}{c}{\textbf{T2: 3D brain tissues}}
  &\multicolumn{2}{c}{\textbf{T3: 2D chest structures}}
  &\multicolumn{2}{c}{\textbf{AVG}}
  \\
  \midrule
  \multirow{2}{*}{\textbf{a) Reliability across samples}}
  &\textbf{DSC} $\uparrow$
  &\textbf{std} $\downarrow$
  &\textbf{DSC} $\uparrow$
  &\textbf{std} $\downarrow$
  &\textbf{DSC} $\uparrow$
  &\textbf{std} $\downarrow$
  &\textbf{DSC} $\uparrow$
  &\textbf{std} $\downarrow$
  \\
  &91.2
  &3.6
  &87.3
  &1.0
  &87.7
  &5.2
  &88.7
  &3.3
  \\
  \midrule
  \multirow{2}{*}{\textbf{b) Reliability across training}}
  &\textbf{\emph{Cor}} $\uparrow$
  &\textbf{\emph{p}} $\downarrow$
  &\textbf{\emph{Cor}} $\uparrow$
  &\textbf{\emph{p}} $\downarrow$
  &\textbf{\emph{Cor}} $\uparrow$
  &\textbf{\emph{p}} $\downarrow$
  &\textbf{\emph{Cor}} $\uparrow$
  &\textbf{\emph{p}} $\downarrow$
  \\
  &0.989
  & $<$0.001
  &0.999
  & $<$0.001
  &0.968
  & $<$0.001
  &0.985
  & $<$0.001
  \\
  \bottomrule
\end{tabular}
}
\label{tab:reliability}
\end{table*}
As shown in Tab.\ref{tab:reliability}, in the three tasks of our Experiment 1, we calculated the standard deviations (std) and the inter-training Pearson correlation coefficients (Cor) \cite{cohen2009pearson}. The results indicate that our GEMINI demonstrates strong reliability across different tested samples and training initializations. \emph{a) Reliability across samples:} We evaluated the DSC and std of the performance across the tested samples. Our GEMINI-Semi achieved an average of 88.7\% DSC with a 3.3 std, indicating high performance with robustness across diverse samples, which supports its reliability in real-world applications. \emph{b) Reliability across training:} We conducted a test-retest reliability analysis \cite{guttman1945basis} and reported the Cor for the performance when our GEMINI-Semi was trained twice from different initialization states. The Cors for all three tasks exceeded 0.95 demonstrating very high consistency between the two training sessions. Additionally, all p-values were below 0.001, indicating significant consistency. Thus, our model shows excellent reliability across different initialization states, which supports its reliability in model implementation.

\section*{D More details in experiments}

\subsection*{D.1 Details of the training diagram}
\begin{figure}[tb]
  \centering
  \includegraphics[width=0.9\linewidth]{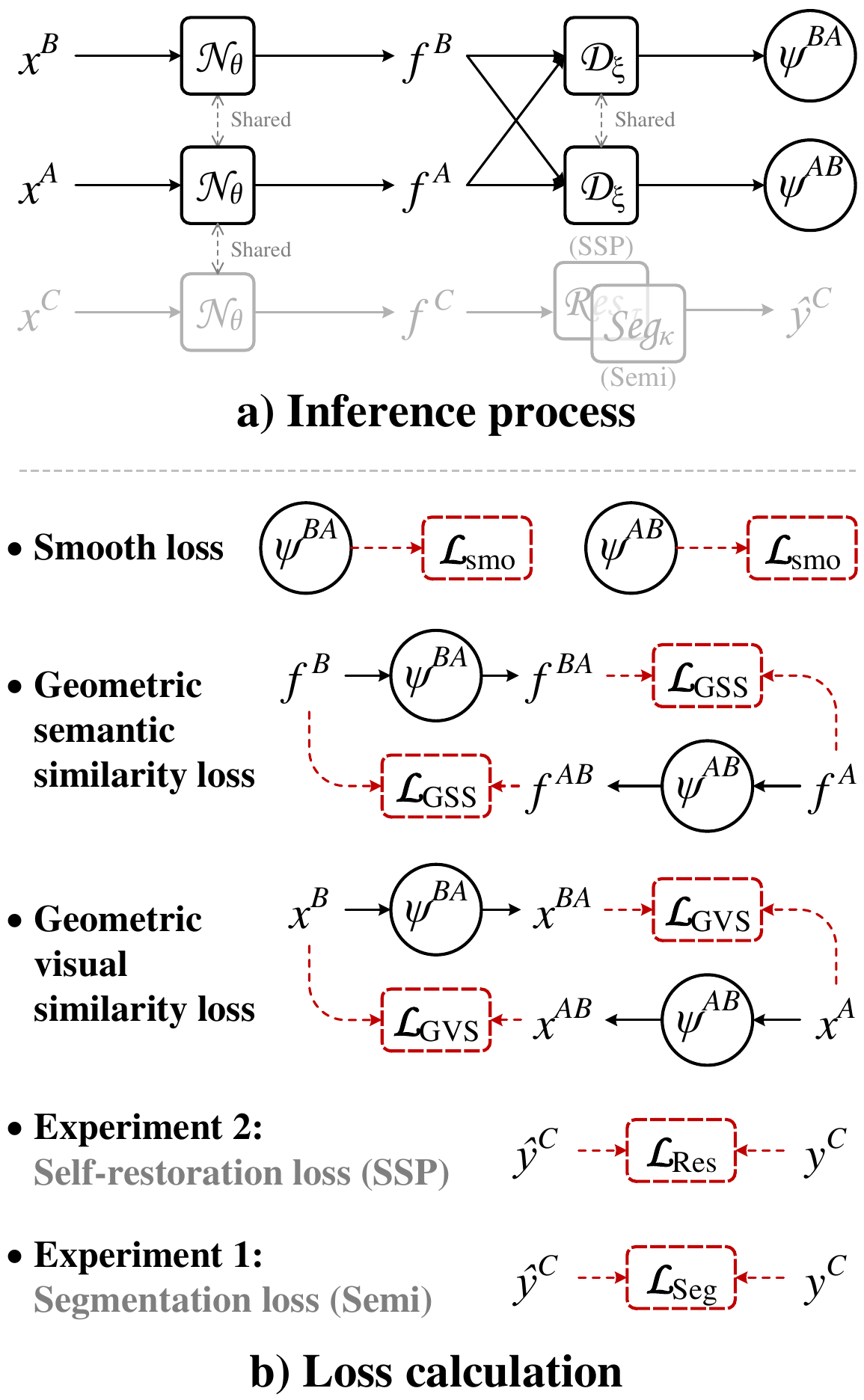}
  \caption{The overall training diagram of our GEMINI. a) The inference process of the whole framework. The gray path in the last line is the additional learning part in the variants of our GEMINI in self-supervised pre-training (GEMINI-MIP) and semi-supervised segmentation (GEMINI-Semi). b) The loss calculation to optimize the whole framework.}\label{supp:fig:diagram}
\end{figure}
\begin{figure*}[!]
  \centering
  \includegraphics[width=\linewidth]{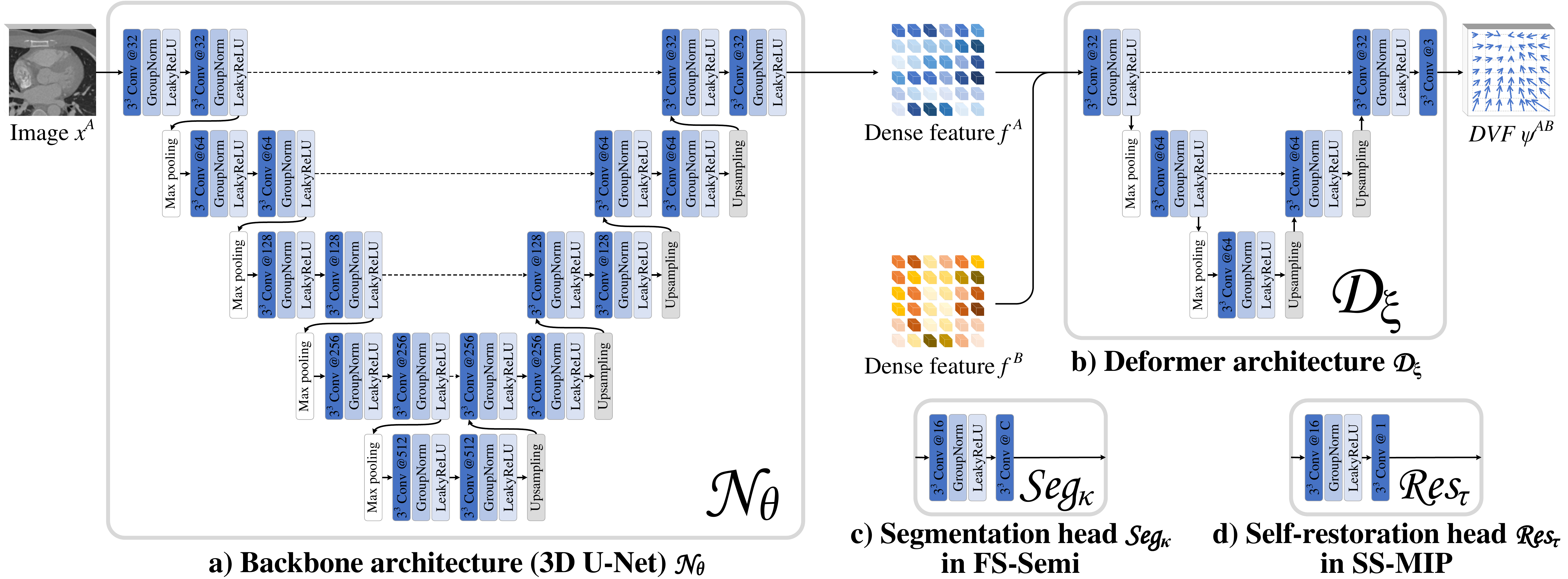}
  \caption{The detailed architecture of our GEMINI. a) The backbone architecture utilizes the 3D U-Net in 3D image tasks and 2D U-Net in 2D image tasks. b) The deformer network architecture utilized a lightweight U-Net. c-d) The segmentation head in the variant of GEMINI-Semi and the self-restoration head in the variant of GEMINI-MIP.}\label{supp:fig:architecture}
\end{figure*}
As shown in Fig.\ref{supp:fig:diagram}, the training diagram introduces the details of our GEMINI's variants in SSP and Semi experiments. In the forward inference, as described in the ``Methodology" section of our paper, two images $x^{A}, x^{B}$ are put into two shared-weight backbones $\mathcal{N}_{\theta}$ separately to extract the features $f^A, f^B$. The features are further put into two shared-weight deformers together to predict two DVFs $\psi^{AB}, \psi^{BA}$ that are bidirectional. For the variant of GEMINI-Semi, a labeled image $x^C$ is put into the shared-weight backbone $\mathcal{N}_{\theta}$, and then put into an additional segmentation head $Seg_{\kappa}$ to predict the segmentation results $\hat{y}^{C}_{Seg}$. For the variant of GEMINI-MIP, an appearance transformed image $x^C$ (described in Sec.5.1.1) is put into the shared-weight backbones $\mathcal{N}_{\theta}$, and then put into an additional self-restoration head $Res_{\tau}$ to predict the restored image $\hat{y}^{C}_{Res}$. In the loss calculation, the smooth loss (Equ.3) is calculated on the DVFs $\psi^{AB}, \psi^{BA}$ to learn the continuity of the deformable mapping, the GVS loss ($\mathcal{L}_{GVS}$, Equ.6) and GSS loss ($\mathcal{L}_{GSS}$, Equ.7) are calculated on the deformed images $x^{AB}=\psi^{AB}(x^A), x^{BA}=\psi^{BA}(x^B)$ and deformed features $f^{AB}=\psi^{AB}(f^A), f^{BA}=\psi^{BA}(f^B)$ (described in Sec.3.3) to learn the correspondence. For the variant of GEMINI-Semi, a segmentation loss $\mathcal{L}_{Seg}$ is calculated on segmentation result $\hat{y}^{C}_{Seg}$ and the groundtruth $y^{C}_{Seg}$. For the variant of GEMINI-MIP, a self-restoration loss $\mathcal{L}_{Res}$ is calculated on the restored image $\hat{y}^{C}_{Res}$ and the original image $y^{C}_{Res}$. The learning of self-restoration in SSP is a fundamental task for a warm-up of our GSS, due to the initial weak representation in the pretext task.

\subsection*{D.2 Details of the architectures and implementation}
As shown in Fig.\ref{supp:fig:architecture}, we utilize the U-Net \cite{ronneberger2015u} architecture (3D U-Net for 3D images and 2D U-Net for 2D images) as our backbone architecture for great basic dense representation in our experiment. In the encoding path, it takes max pooling layers to reduce the feature maps' resolution and in the decoding path, it takes up-sampling layers (bilinear for 2D images and trilinear for 3D images) to restore the features' resolution. Skip connections are used to transmit features from the encoding path to the decoding path in each resolution stage. There are five resolution stages in the network and each stage utilizes Conv-GN-LeckyReLU\footnote{Conv is a convolution layer and GN is a group normalization layer \cite{wu2018group}.} modules to extract features. The deformer network also takes a lightweight U-Net architecture with very shallow depth to estimate the DVF. It only has three resolution stages and each stage has half of the Conv-GN-LeckyReLU module amount compared with the backbone. Both the segmentation head and self-restoration head take one Conv-GN-LeckyReLU module to project the input features and follow a convolution layer to predict the targets. The detailed hyper-parameters inner these architectures are marked in Fig.\ref{supp:fig:architecture}.

\bibliographystyle{IEEEtran}
\bibliography{mybib}

\begin{thebibliography}{100}
\providecommand{\url}[1]{#1}
\csname url@samestyle\endcsname
\providecommand{\newblock}{\relax}
\providecommand{\bibinfo}[2]{#2}
\providecommand{\BIBentrySTDinterwordspacing}{\spaceskip=0pt\relax}
\providecommand{\BIBentryALTinterwordstretchfactor}{4}
\providecommand{\BIBentryALTinterwordspacing}{\spaceskip=\fontdimen2\font plus
\BIBentryALTinterwordstretchfactor\fontdimen3\font minus \fontdimen4\font\relax}
\providecommand{\BIBforeignlanguage}[2]{{%
\expandafter\ifx\csname l@#1\endcsname\relax
\typeout{** WARNING: IEEEtran.bst: No hyphenation pattern has been}%
\typeout{** loaded for the language `#1'. Using the pattern for}%
\typeout{** the default language instead.}%
\else
\language=\csname l@#1\endcsname
\fi
#2}}
\providecommand{\BIBdecl}{\relax}
\BIBdecl

\bibitem{li2021dense}
X.~Li, Y.~Zhou, Y.~Zhang, A.~Zhang, W.~Wang, N.~Jiang, H.~Wu, and W.~Wang, ``Dense semantic contrast for self-supervised visual representation learning,'' in \emph{Proceedings of the 29th ACM International Conference on Multimedia}, 2021, pp. 1368--1376.

\bibitem{o2020unsupervised}
P.~O. O~Pinheiro, A.~Almahairi, R.~Benmalek, F.~Golemo, and A.~C. Courville, ``Unsupervised learning of dense visual representations,'' \emph{Advances in Neural Information Processing Systems}, vol.~33, pp. 4489--4500, 2020.

\bibitem{wang2022densecl}
X.~Wang, R.~Zhang, C.~Shen, and T.~Kong, ``Densecl: A simple framework for self-supervised dense visual pre-training,'' \emph{Visual Informatics}, 2022.

\bibitem{wang2022exploring}
Z.~Wang, Q.~Li, G.~Zhang, P.~Wan, W.~Zheng, N.~Wang, M.~Gong, and T.~Liu, ``Exploring set similarity for dense self-supervised representation learning,'' in \emph{Proceedings of the IEEE/CVF Conference on Computer Vision and Pattern Recognition}, 2022, pp. 16\,590--16\,599.

\bibitem{xie2021propagate}
Z.~Xie, Y.~Lin, Z.~Zhang, Y.~Cao, S.~Lin, and H.~Hu, ``Propagate yourself: Exploring pixel-level consistency for unsupervised visual representation learning,'' in \emph{Proceedings of the IEEE/CVF Conference on Computer Vision and Pattern Recognition}, 2021, pp. 16\,684--16\,693.

\bibitem{bengio2013representation}
Y.~Bengio, A.~Courville, and P.~Vincent, ``Representation learning: A review and new perspectives,'' \emph{IEEE transactions on pattern analysis and machine intelligence}, vol.~35, no.~8, pp. 1798--1828, 2013.

\bibitem{you2022simcvd}
C.~You, Y.~Zhou, R.~Zhao, L.~Staib, and J.~S. Duncan, ``Simcvd: Simple contrastive voxel-wise representation distillation for semi-supervised medical image segmentation,'' \emph{IEEE Transactions on Medical Imaging}, vol.~41, no.~9, pp. 2228--2237, 2022.

\bibitem{he2022learning}
Y.~He, R.~Ge, X.~Qi, Y.~Chen, J.~Wu, J.-L. Coatrieux, G.~Yang, and S.~Li, ``Learning better registration to learn better few-shot medical image segmentation: Authenticity, diversity, and robustness,'' \emph{IEEE Transactions on Neural Networks and Learning Systems}, 2022.

\bibitem{he2021meta}
Y.~He, G.~Yang, J.~Yang, R.~Ge, Y.~Kong, X.~Zhu, S.~Zhang, P.~Shao, H.~Shu, J.-L. Dillenseger \emph{et~al.}, ``Meta grayscale adaptive network for 3d integrated renal structures segmentation,'' \emph{Medical Image Analysis}, vol.~71, p. 102055, 2021.

\bibitem{you2022class}
C.~You, R.~Zhao, F.~Liu, S.~Dong, S.~Chinchali, U.~Topcu, L.~Staib, and J.~Duncan, ``Class-aware adversarial transformers for medical image segmentation,'' \emph{Advances in Neural Information Processing Systems}, vol.~35, pp. 29\,582--29\,596, 2022.

\bibitem{piccialli2021survey}
F.~Piccialli, V.~Di~Somma, F.~Giampaolo, S.~Cuomo, and G.~Fortino, ``A survey on deep learning in medicine: Why, how and when?'' \emph{Information Fusion}, vol.~66, pp. 111--137, 2021.

\bibitem{cheplygina2019not}
V.~Cheplygina, M.~de~Bruijne, and J.~P. Pluim, ``Not-so-supervised: a survey of semi-supervised, multi-instance, and transfer learning in medical image analysis,'' \emph{Medical image analysis}, vol.~54, pp. 280--296, 2019.

\bibitem{milbich2021visual}
T.~Milbich, ``Visual similarity and representation learning,'' Ph.D. dissertation, 2021.

\bibitem{zhang2022attributable}
B.~Zhang, W.~Zheng, J.~Zhou, and J.~Lu, ``Attributable visual similarity learning,'' in \emph{Proceedings of the IEEE/CVF Conference on Computer Vision and Pattern Recognition}, 2022, pp. 7532--7541.

\bibitem{roth2020pads}
K.~Roth, T.~Milbich, and B.~Ommer, ``Pads: Policy-adapted sampling for visual similarity learning,'' in \emph{Proceedings of the IEEE/CVF Conference on Computer Vision and Pattern Recognition}, 2020, pp. 6568--6577.

\bibitem{you2022momentum}
C.~You, R.~Zhao, L.~H. Staib, and J.~S. Duncan, ``Momentum contrastive voxel-wise representation learning for semi-supervised volumetric medical image segmentation,'' in \emph{International Conference on Medical Image Computing and Computer-Assisted Intervention}.\hskip 1em plus 0.5em minus 0.4em\relax Springer, 2022, pp. 639--652.

\bibitem{litjens2017survey}
G.~Litjens, T.~Kooi, B.~E. Bejnordi, A.~A.~A. Setio, F.~Ciompi, M.~Ghafoorian, J.~A. Van Der~Laak, B.~Van~Ginneken, and C.~I. S{\'a}nchez, ``A survey on deep learning in medical image analysis,'' \emph{Medical image analysis}, vol.~42, pp. 60--88, 2017.

\bibitem{chuang2020debiased}
C.-Y. Chuang, J.~Robinson, Y.-C. Lin, A.~Torralba, and S.~Jegelka, ``Debiased contrastive learning,'' \emph{Advances in neural information processing systems}, vol.~33, pp. 8765--8775, 2020.

\bibitem{chuang2022robust}
C.-Y. Chuang, R.~D. Hjelm, X.~Wang, V.~Vineet, N.~Joshi, A.~Torralba, S.~Jegelka, and Y.~Song, ``Robust contrastive learning against noisy views,'' in \emph{Proceedings of the IEEE/CVF Conference on Computer Vision and Pattern Recognition}, 2022, pp. 16\,670--16\,681.

\bibitem{zhou2019high}
S.~Zhou, D.~Nie, E.~Adeli, J.~Yin, J.~Lian, and D.~Shen, ``High-resolution encoder--decoder networks for low-contrast medical image segmentation,'' \emph{IEEE Transactions on Image Processing}, vol.~29, pp. 461--475, 2019.

\bibitem{he2020momentum}
K.~He, H.~Fan, Y.~Wu, S.~Xie, and R.~Girshick, ``Momentum contrast for unsupervised visual representation learning,'' in \emph{Proceedings of the IEEE/CVF conference on computer vision and pattern recognition}, 2020, pp. 9729--9738.

\bibitem{chen2020simple}
T.~Chen, S.~Kornblith, M.~Norouzi, and G.~Hinton, ``A simple framework for contrastive learning of visual representations,'' in \emph{International conference on machine learning}.\hskip 1em plus 0.5em minus 0.4em\relax PMLR, 2020, pp. 1597--1607.

\bibitem{chen2020improved}
X.~Chen, H.~Fan, R.~Girshick, and K.~He, ``Improved baselines with momentum contrastive learning,'' \emph{arXiv preprint arXiv:2003.04297}, 2020.

\bibitem{grill2020bootstrap}
J.-B. Grill, F.~Strub, F.~Altch{\'e}, C.~Tallec, P.~Richemond, E.~Buchatskaya, C.~Doersch, B.~Pires, Z.~Guo, M.~Azar \emph{et~al.}, ``Bootstrap your own latent: A new approach to self-supervised learning,'' in \emph{Neural Information Processing Systems}, 2020.

\bibitem{He2020CVPR}
K.~He, H.~Fan, Y.~Wu, S.~Xie, and R.~Girshick, ``Momentum contrast for unsupervised visual representation learning,'' in \emph{Proceedings of the IEEE/CVF Conference on Computer Vision and Pattern Recognition (CVPR)}, June 2020.

\bibitem{alexandroff2013topologie}
P.~Alexandroff and H.~Hopf, \emph{Topologie I: Erster Band. Grundbegriffe der Mengentheoretischen Topologie Topologie der Komplexe{\textperiodcentered} Topologische Invarianzs{\"a}tze und Anschliessende Begriffsbildungen{\textperiodcentered} Verschlingungen im n-Dimensionalen Euklidischen Raum Stetige Abbildungen von Polyedern}.\hskip 1em plus 0.5em minus 0.4em\relax Springer-Verlag, 2013.

\bibitem{i1996new}
J.~D. i~Ferrer, ``A new privacy homomorphism and applications,'' \emph{Information Processing Letters}, vol.~60, no.~5, pp. 277--282, 1996.

\bibitem{heimann2009statistical}
T.~Heimann and H.-P. Meinzer, ``Statistical shape models for 3d medical image segmentation: a review,'' \emph{Medical image analysis}, vol.~13, no.~4, pp. 543--563, 2009.

\bibitem{bazin2008homeomorphic}
P.-L. Bazin and D.~L. Pham, ``Homeomorphic brain image segmentation with topological and statistical atlases,'' \emph{Medical image analysis}, vol.~12, no.~5, pp. 616--625, 2008.

\bibitem{miller2001group}
M.~I. Miller and L.~Younes, ``Group actions, homeomorphisms, and matching: A general framework,'' \emph{International Journal of Computer Vision}, vol.~41, no.~1, pp. 61--84, 2001.

\bibitem{hubbard1991differential}
J.~H. Hubbard and B.~H. West, \emph{Differential Equations: A Dynamical Systems Approach: A Dynamical Systems Approach. Part II: Higher Dimensional Systems}.\hskip 1em plus 0.5em minus 0.4em\relax Springer Science \& Business Media, 1991.

\bibitem{netter2014atlas}
F.~H. Netter, \emph{Atlas of human anatomy, Professional Edition E-Book: including NetterReference. com Access with full downloadable image Bank}.\hskip 1em plus 0.5em minus 0.4em\relax Elsevier health sciences, 2014.

\bibitem{He_2023_CVPR}
Y.~He, G.~Yang, R.~Ge, Y.~Chen, J.-L. Coatrieux, B.~Wang, and S.~Li, ``Geometric visual similarity learning in 3d medical image self-supervised pre-training,'' in \emph{Proceedings of the IEEE/CVF Conference on Computer Vision and Pattern Recognition (CVPR)}, June 2023.

\bibitem{7987758}
S.~Darkner, A.~Pai, M.~G. Liptrot, and J.~Sporring, ``Collocation for diffeomorphic deformations in medical image registration,'' \emph{IEEE Transactions on Pattern Analysis and Machine Intelligence}, vol.~40, no.~7, pp. 1570--1583, 2018.

\bibitem{ba2018un}
G.~Balakrishnan, A.~Zhao, M.~R. Sabuncu, J.~Guttag, and A.~V. Dalca, ``An unsupervised learning model for deformable medical image registration,'' in \emph{Proceedings of the IEEE conference on computer vision and pattern recognition}, 2018, pp. 9252--9260.

\bibitem{haskins2020deep}
G.~Haskins, U.~Kruger, and P.~Yan, ``Deep learning in medical image registration: a survey,'' \emph{Machine Vision and Applications}, vol.~31, no.~1, pp. 1--18, 2020.

\bibitem{dalca2019unsupervised}
A.~V. Dalca, G.~Balakrishnan, J.~Guttag, and M.~R. Sabuncu, ``Unsupervised learning of probabilistic diffeomorphic registration for images and surfaces,'' \emph{Medical image analysis}, vol.~57, pp. 226--236, 2019.

\bibitem{nehaniv2002correspondence}
C.~L. Nehaniv, K.~Dautenhahn \emph{et~al.}, ``The correspondence problem,'' \emph{Imitation in animals and artifacts}, vol.~41, 2002.

\bibitem{brass2005imitation}
M.~Brass and C.~Heyes, ``Imitation: is cognitive neuroscience solving the correspondence problem?'' \emph{Trends in cognitive sciences}, vol.~9, no.~10, pp. 489--495, 2005.

\bibitem{scholkopf2005object}
B.~Sch{\"o}lkopf, F.~Steinke, and V.~Blanz, ``Object correspondence as a machine learning problem,'' in \emph{Proceedings of the 22nd international conference on Machine learning}, 2005, pp. 776--783.

\bibitem{he2021few}
Y.~He, T.~Li, R.~Ge, J.~Yang, Y.~Kong, J.~Zhu, H.~Shu, G.~Yang, and S.~Li, ``Few-shot learning for deformable medical image registration with perception-correspondence decoupling and reverse teaching,'' \emph{IEEE Journal of Biomedical and Health Informatics}, 2021.

\bibitem{lake2015human}
B.~M. Lake, R.~Salakhutdinov, and J.~B. Tenenbaum, ``Human-level concept learning through probabilistic program induction,'' \emph{Science}, vol. 350, no. 6266, pp. 1332--1338, 2015.

\bibitem{jumper2021highly}
J.~Jumper, R.~Evans, A.~Pritzel, T.~Green, M.~Figurnov, O.~Ronneberger, K.~Tunyasuvunakool, R.~Bates, A.~{\v{Z}}{\'\i}dek, A.~Potapenko \emph{et~al.}, ``Highly accurate protein structure prediction with alphafold,'' \emph{Nature}, vol. 596, no. 7873, pp. 583--589, 2021.

\bibitem{wu2016google}
Y.~Wu, M.~Schuster, Z.~Chen, Q.~V. Le, M.~Norouzi, W.~Macherey, M.~Krikun, Y.~Cao, Q.~Gao, K.~Macherey \emph{et~al.}, ``Google's neural machine translation system: Bridging the gap between human and machine translation,'' \emph{arXiv preprint arXiv:1609.08144}, 2016.

\bibitem{chen2021unsupervised}
H.~Chen, J.~Li, R.~Wang, Y.~Huang, F.~Meng, D.~Meng, Q.~Peng, and L.~Wang, ``Unsupervised learning of local discriminative representation for medical images,'' in \emph{Information Processing in Medical Imaging: 27th International Conference, IPMI 2021, Virtual Event, June 28--June 30, 2021, Proceedings 27}.\hskip 1em plus 0.5em minus 0.4em\relax Springer, 2021, pp. 373--385.

\bibitem{gao2022unsupervised}
Z.~Gao, C.~Jia, Y.~Li, X.~Zhang, B.~Hong, J.~Wu, T.~Gong, C.~Wang, D.~Meng, Y.~Zheng \emph{et~al.}, ``Unsupervised representation learning for tissue segmentation in histopathological images: From global to local contrast,'' \emph{IEEE Transactions on Medical Imaging}, vol.~41, no.~12, pp. 3611--3623, 2022.

\bibitem{de2000mahalanobis}
R.~De~Maesschalck, D.~Jouan-Rimbaud, and D.~L. Massart, ``The mahalanobis distance,'' \emph{Chemometrics and intelligent laboratory systems}, vol.~50, no.~1, pp. 1--18, 2000.

\bibitem{wang2005euclidean}
L.~Wang, Y.~Zhang, and J.~Feng, ``On the euclidean distance of images,'' \emph{IEEE transactions on pattern analysis and machine intelligence}, vol.~27, no.~8, pp. 1334--1339, 2005.

\bibitem{chaitanya2020contrastive}
K.~Chaitanya, E.~Erdil, N.~Karani, and E.~Konukoglu, ``Contrastive learning of global and local features for medical image segmentation with limited annotations,'' \emph{Advances in neural information processing systems}, vol.~33, pp. 12\,546--12\,558, 2020.

\bibitem{Chen2021CVPR}
X.~Chen and K.~He, ``Exploring simple siamese representation learning,'' in \emph{Proceedings of the IEEE/CVF Conference on Computer Vision and Pattern Recognition (CVPR)}, June 2021, pp. 15\,750--15\,758.

\bibitem{caron2018deep}
M.~Caron, P.~Bojanowski, A.~Joulin, and M.~Douze, ``Deep clustering for unsupervised learning of visual features,'' in \emph{European Conference on Computer Vision}, 2018.

\bibitem{jing2021understanding}
L.~Jing, P.~Vincent, Y.~LeCun, and Y.~Tian, ``Understanding dimensional collapse in contrastive self-supervised learning,'' in \emph{International Conference on Learning Representations}, 2021.

\bibitem{huynh2022boosting}
T.~Huynh, S.~Kornblith, M.~R. Walter, M.~Maire, and M.~Khademi, ``Boosting contrastive self-supervised learning with false negative cancellation,'' in \emph{Proceedings of the IEEE/CVF Winter Conference on Applications of Computer Vision}, 2022, pp. 2785--2795.

\bibitem{6226425}
H.~Wang, J.~W. Suh, S.~R. Das, J.~B. Pluta, C.~Craige, and P.~A. Yushkevich, ``Multi-atlas segmentation with joint label fusion,'' \emph{IEEE Transactions on Pattern Analysis and Machine Intelligence}, vol.~35, no.~3, pp. 611--623, 2013.

\bibitem{8014481}
L.~M. Koch, M.~Rajchl, W.~Bai, C.~F. Baumgartner, T.~Tong, J.~Passerat-Palmbach, P.~Aljabar, and D.~Rueckert, ``Multi-atlas segmentation using partially annotated data: Methods and annotation strategies,'' \emph{IEEE Transactions on Pattern Analysis and Machine Intelligence}, vol.~40, no.~7, pp. 1683--1696, 2018.

\bibitem{wang2020lt}
S.~{Wang}, S.~{Cao}, D.~{Wei}, R.~{Wang}, K.~{Ma}, L.~{Wang}, D.~{Meng}, and Y.~{Zheng}, ``Lt-net: Label transfer by learning reversible voxel-wise correspondence for one-shot medical image segmentation,'' in \emph{2020 IEEE/CVF Conference on Computer Vision and Pattern Recognition (CVPR)}, 2020, pp. 9162--9171.

\bibitem{BalakrishnanVoxelMorph(u)}
G.~Balakrishnan, A.~Zhao, M.~R. Sabuncu, J.~Guttag, and A.~V. Dalca, ``Voxelmorph: A learning framework for deformable medical image registration,'' \emph{IEEE Transactions on Medical Imaging}, vol.~38, pp. 1788--1800, 2019.

\bibitem{lecun2015deep}
Y.~LeCun, Y.~Bengio, and G.~Hinton, ``Deep learning,'' \emph{nature}, vol. 521, no. 7553, pp. 436--444, 2015.

\bibitem{shen2017deep}
D.~Shen, G.~Wu, and H.-I. Suk, ``Deep learning in medical image analysis,'' \emph{Annual review of biomedical engineering}, vol.~19, p. 221, 2017.

\bibitem{he2020deep}
Y.~He, T.~Li, G.~Yang, Y.~Kong, Y.~Chen, H.~Shu, J.-L. Coatrieux, J.-L. Dillenseger, and S.~Li, ``Deep complementary joint model for complex scene registration and few-shot segmentation on medical images,'' in \emph{European Conference on Computer Vision}.\hskip 1em plus 0.5em minus 0.4em\relax Springer, 2020, pp. 770--786.

\bibitem{ding2021modeling}
Y.~{Ding}, X.~{Yu}, and Y.~{Yang}, ``Modeling the probabilistic distribution of unlabeled data for one-shot medical image segmentation,'' in \emph{AAAI}, 2021, pp. 1246--1254.

\bibitem{zhao2019data}
A.~Zhao, G.~Balakrishnan, F.~Durand, J.~V. Guttag, and A.~V. Dalca, ``Data augmentation using learned transformations for one-shot medical image segmentation,'' in \emph{Proceedings of the IEEE conference on computer vision and pattern recognition}, 2019, pp. 8543--8553.

\bibitem{xu2019deepatlas}
Z.~Xu and M.~Niethammer, ``Deepatlas: Joint semi-supervised learning of image registration and segmentation,'' in \emph{International Conference on Medical Image Computing and Computer-Assisted Intervention}.\hskip 1em plus 0.5em minus 0.4em\relax Springer, 2019, pp. 420--429.

\bibitem{kong1989digital}
T.~Y. Kong and A.~Rosenfeld, ``Digital topology: Introduction and survey,'' \emph{Computer Vision, Graphics, and Image Processing}, vol.~48, no.~3, pp. 357--393, 1989.

\bibitem{bottou2010large}
L.~Bottou, ``Large-scale machine learning with stochastic gradient descent,'' in \emph{Proceedings of COMPSTAT'2010}.\hskip 1em plus 0.5em minus 0.4em\relax Springer, 2010, pp. 177--186.

\bibitem{kingma2014adam}
D.~P. Kingma and J.~Ba, ``Adam: A method for stochastic optimization,'' \emph{arXiv preprint arXiv:1412.6980}, 2014.

\bibitem{saad2009coalitional}
W.~Saad, Z.~Han, M.~Debbah, A.~Hjorungnes, and T.~Basar, ``Coalitional game theory for communication networks,'' \emph{Ieee signal processing magazine}, vol.~26, no.~5, pp. 77--97, 2009.

\bibitem{zhou2019models}
Z.~Zhou, V.~Sodha, J.~Pang, M.~B. Gotway, and J.~Liang, ``Models genesis,'' vol.~67.\hskip 1em plus 0.5em minus 0.4em\relax Elsevier, 2021, p. 101840.

\bibitem{pathak2016context}
D.~Pathak, P.~Krahenbuhl, J.~Donahue, T.~Darrell, and A.~A. Efros, ``Context encoders: Feature learning by inpainting,'' in \emph{Proceedings of the IEEE conference on computer vision and pattern recognition}, 2016, pp. 2536--2544.

\bibitem{ma2021loss}
J.~Ma, J.~Chen, M.~Ng, R.~Huang, Y.~Li, C.~Li, X.~Yang, and A.~L. Martel, ``Loss odyssey in medical image segmentation,'' \emph{Medical Image Analysis}, vol.~71, p. 102035, 2021.

\bibitem{gharleghi2022automated}
R.~Gharleghi, D.~Adikari, K.~Ellenberger, S.-Y. Ooi, C.~Ellis, C.-M. Chen, R.~Gao, Y.~He, R.~Hussain, C.-Y. Lee \emph{et~al.}, ``Automated segmentation of normal and diseased coronary arteries-the asoca challenge,'' \emph{Computerized Medical Imaging and Graphics}, p. 102049, 2022.

\bibitem{schaap2009standardized}
M.~Schaap, C.~T. Metz, T.~van Walsum, A.~G. van~der Giessen, A.~C. Weustink, N.~R. Mollet, C.~Bauer, H.~Bogunovi{\'c}, C.~Castro, X.~Deng \emph{et~al.}, ``Standardized evaluation methodology and reference database for evaluating coronary artery centerline extraction algorithms,'' \emph{Medical image analysis}, vol.~13, no.~5, pp. 701--714, 2009.

\bibitem{zhuang2019evaluation}
X.~Zhuang, L.~Li, C.~Payer, D.~{\v{S}}tern, M.~Urschler, M.~P. Heinrich, J.~Oster, C.~Wang, {\"O}.~Smedby, C.~Bian \emph{et~al.}, ``Evaluation of algorithms for multi-modality whole heart segmentation: an open-access grand challenge,'' \emph{Medical image analysis}, vol.~58, p. 101537, 2019.

\bibitem{kennedy2012candishare}
D.~N. Kennedy, C.~Haselgrove, S.~M. Hodge, P.~S. Rane, N.~Makris, and J.~A. Frazier, ``Candishare: A resource for pediatric neuroimaging data,'' \emph{Neuroinformatics}, vol.~10, no.~3, p. 319, 2012.

\bibitem{van2006segmentation}
B.~Van~Ginneken, M.~B. Stegmann, and M.~Loog, ``Segmentation of anatomical structures in chest radiographs using supervised methods: a comparative study on a public database,'' \emph{Medical image analysis}, vol.~10, no.~1, pp. 19--40, 2006.

\bibitem{wang2017chestx}
X.~Wang, Y.~Peng, L.~Lu, Z.~Lu, M.~Bagheri, and R.~M. Summers, ``Chestx-ray8: Hospital-scale chest x-ray database and benchmarks on weakly-supervised classification and localization of common thorax diseases,'' in \emph{Proceedings of the IEEE conference on computer vision and pattern recognition}, 2017, pp. 2097--2106.

\bibitem{you2024mine}
C.~You, W.~Dai, F.~Liu, Y.~Min, N.~C. Dvornek, X.~Li, D.~A. Clifton, L.~Staib, and J.~S. Duncan, ``Mine your own anatomy: Revisiting medical image segmentation with extremely limited labels,'' \emph{IEEE Transactions on Pattern Analysis and Machine Intelligence}, 2024.

\bibitem{you2024rethinking}
C.~You, W.~Dai, Y.~Min, F.~Liu, D.~Clifton, S.~K. Zhou, L.~Staib, and J.~Duncan, ``Rethinking semi-supervised medical image segmentation: A variance-reduction perspective,'' \emph{Advances in neural information processing systems}, vol.~36, 2024.

\bibitem{shiraishi2000development}
J.~Shiraishi, S.~Katsuragawa, J.~Ikezoe, T.~Matsumoto, T.~Kobayashi, K.-i. Komatsu, M.~Matsui, H.~Fujita, Y.~Kodera, and K.~Doi, ``Development of a digital image database for chest radiographs with and without a lung nodule: receiver operating characteristic analysis of radiologists' detection of pulmonary nodules,'' \emph{American Journal of Roentgenology}, vol. 174, no.~1, pp. 71--74, 2000.

\bibitem{ronneberger2015u}
O.~Ronneberger, P.~Fischer, and T.~Brox, ``U-net: Convolutional networks for biomedical image segmentation,'' in \emph{International Conference on Medical image computing and computer-assisted intervention}.\hskip 1em plus 0.5em minus 0.4em\relax Springer, 2015, pp. 234--241.

\bibitem{yu2019uncertainty}
L.~{Yu}, S.~{Wang}, X.~{Li}, C.-W. {Fu}, and P.-A. {Heng}, ``Uncertainty-aware self-ensembling model for semi-supervised 3d left atrium segmentation,'' in \emph{Lecture Notes in Computer Science (including subseries Lecture Notes in Artificial Intelligence and Lecture Notes in Bioinformatics)}, 2019, pp. 605--613.

\bibitem{chen2019multi}
S.~{Chen}, G.~{Bortsova}, A.~G.-U. {Juárez}, G.~van {Tulder}, and M.~de~{Bruijne}, ``Multi-task attention-based semi-supervised learning for medical image segmentation,'' in \emph{International Conference on Medical Image Computing and Computer-Assisted Intervention}, 2019, pp. 457--465.

\bibitem{he2020dense}
Y.~He, G.~Yang, J.~Yang, Y.~Chen, Y.~Kong, J.~Wu, L.~Tang, X.~Zhu, J.-L. Dillenseger, P.~Shao \emph{et~al.}, ``Dense biased networks with deep priori anatomy and hard region adaptation: Semi-supervised learning for fine renal artery segmentation,'' \emph{Medical image analysis}, vol.~63, p. 101722, 2020.

\bibitem{chen2021semi}
X.~{Chen}, Y.~{Yuan}, G.~{Zeng}, and J.~{Wang}, ``Semi-supervised semantic segmentation with cross pseudo supervision,'' in \emph{Proceedings of the IEEE/CVF Conference on Computer Vision and Pattern Recognition}, 2021, pp. 2613--2622.

\bibitem{wu2018group}
Y.~Wu and K.~He, ``Group normalization,'' in \emph{Proceedings of the European conference on computer vision (ECCV)}, 2018, pp. 3--19.

\bibitem{paszke2019pytorch}
A.~Paszke, S.~Gross, F.~Massa, A.~Lerer, J.~Bradbury, G.~Chanan, T.~Killeen, Z.~Lin, N.~Gimelshein, L.~Antiga \emph{et~al.}, ``Pytorch: An imperative style, high-performance deep learning library,'' \emph{Advances in neural information processing systems}, vol.~32, 2019.

\bibitem{taha2015metrics}
A.~A. Taha and A.~Hanbury, ``Metrics for evaluating 3d medical image segmentation: analysis, selection, and tool,'' \emph{BMC medical imaging}, vol.~15, no.~1, pp. 1--28, 2015.

\bibitem{deng2009imagenet}
J.~Deng, W.~Dong, R.~Socher, L.-J. Li, K.~Li, and L.~Fei-Fei, ``Imagenet: A large-scale hierarchical image database,'' in \emph{2009 IEEE conference on computer vision and pattern recognition}.\hskip 1em plus 0.5em minus 0.4em\relax Ieee, 2009, pp. 248--255.

\bibitem{vincent2010stacked}
P.~Vincent, H.~Larochelle, I.~Lajoie, Y.~Bengio, P.-A. Manzagol, and L.~Bottou, ``Stacked denoising autoencoders: Learning useful representations in a deep network with a local denoising criterion.'' \emph{Journal of machine learning research}, vol.~11, no.~12, 2010.

\bibitem{komodakis2018unsupervised}
N.~Komodakis and S.~Gidaris, ``Unsupervised representation learning by predicting image rotations,'' in \emph{International Conference on Learning Representations (ICLR)}, 2018.

\bibitem{luo2016big}
J.~Luo, M.~Wu, D.~Gopukumar, and Y.~Zhao, ``Big data application in biomedical research and health care: a literature review,'' \emph{Biomedical informatics insights}, vol.~8, pp. BII--S31\,559, 2016.

\bibitem{yang2020superpixel}
F.~Yang, Q.~Sun, H.~Jin, and Z.~Zhou, ``Superpixel segmentation with fully convolutional networks,'' in \emph{Proceedings of the IEEE/CVF conference on computer vision and pattern recognition}, 2020, pp. 13\,964--13\,973.

\bibitem{liu2024imaging}
Y.~Liu, R.~Ge, Y.~He, Z.~Wu, C.~You, S.~Li, and Y.~Chen, ``Imaging foundation model for universal enhancement of non-ideal measurement ct,'' \emph{arXiv preprint arXiv:2410.01591}, 2024.

\bibitem{10750441}
Y.~He, F.~Huang, X.~Jiang, Y.~Nie, M.~Wang, J.~Wang, and H.~Chen, ``Foundation model for advancing healthcare: Challenges, opportunities and future directions,'' \emph{IEEE Reviews in Biomedical Engineering}, pp. 1--20, 2024.

\bibitem{chen2019med3d}
S.~Chen, K.~Ma, and Y.~Zheng, ``Med3d: Transfer learning for 3d medical image analysis,'' \emph{arXiv preprint arXiv:1904.00625}, 2019.

\bibitem{isensee2019automated}
F.~Isensee, M.~Schell, I.~Pflueger, G.~Brugnara, D.~Bonekamp, U.~Neuberger, A.~Wick, H.-P. Schlemmer, S.~Heiland, W.~Wick \emph{et~al.}, ``Automated brain extraction of multisequence mri using artificial neural networks,'' \emph{Human brain mapping}, vol.~40, no.~17, pp. 4952--4964, 2019.

\bibitem{wang2019panet}
K.~Wang, J.~H. Liew, Y.~Zou, D.~Zhou, and J.~Feng, ``Panet: Few-shot image semantic segmentation with prototype alignment,'' in \emph{proceedings of the IEEE/CVF international conference on computer vision}, 2019, pp. 9197--9206.

\bibitem{van2008visualizing}
L.~Van~der Maaten and G.~Hinton, ``Visualizing data using t-sne.'' \emph{Journal of machine learning research}, vol.~9, no.~11, 2008.

\bibitem{chen2021transunet}
J.~Chen, Y.~Lu, Q.~Yu, X.~Luo, E.~Adeli, Y.~Wang, L.~Lu, A.~L. Yuille, and Y.~Zhou, ``Transunet: Transformers make strong encoders for medical image segmentation,'' \emph{arXiv preprint arXiv:2102.04306}, 2021.

\bibitem{cao2022swin}
H.~Cao, Y.~Wang, J.~Chen, D.~Jiang, X.~Zhang, Q.~Tian, and M.~Wang, ``Swin-unet: Unet-like pure transformer for medical image segmentation,'' in \emph{European conference on computer vision}.\hskip 1em plus 0.5em minus 0.4em\relax Springer, 2022, pp. 205--218.

\bibitem{cohen2009pearson}
I.~Cohen, Y.~Huang, J.~Chen, J.~Benesty, J.~Benesty, J.~Chen, Y.~Huang, and I.~Cohen, ``Pearson correlation coefficient,'' \emph{Noise reduction in speech processing}, pp. 1--4, 2009.

\bibitem{guttman1945basis}
L.~Guttman, ``A basis for analyzing test-retest reliability,'' \emph{Psychometrika}, vol.~10, no.~4, pp. 255--282, 1945.

\end{thebibliography}

\end{document}